\documentclass{article}
\usepackage{style/foci}
\usepackage{tgpagella}   
\usepackage{tgheros}     

\usepackage[utf8]{inputenc}
\usepackage[T1]{fontenc}
\usepackage{microtype}

\usepackage{amsmath}
\usepackage{amssymb}
\usepackage{amsfonts}
\usepackage{amsthm}
\usepackage{mathtools}
\usepackage{mathrsfs}
\usepackage{physics}
\usepackage{braket}
\usepackage{slashed}
\usepackage{nicefrac}
\usepackage{textcomp}
\usepackage{dsfont}
\usepackage{bbm}
\usepackage{bm}

\usepackage{graphicx}
\usepackage{subcaption}
\usepackage[export]{adjustbox}
\usepackage{float}
\usepackage{booktabs}
\usepackage{dcolumn}
\newcolumntype{d}[1]{D{.}{.}{#1}}
\usepackage{bigstrut, tabularx, multirow, makecell, diagbox}
\usepackage{colortbl}
\usepackage{tabularray}
\UseTblrLibrary{booktabs}
\usepackage{threeparttable}
\usepackage{tablefootnote}
\usepackage{fontawesome5}

\usepackage{placeins}
\usepackage{caption}
\usepackage{footnote}
\usepackage{enumitem}
\usepackage{multicol}
\usepackage{xspace}
\usepackage{titletoc}
\usepackage{titlesec}
\usepackage[bottom]{footmisc}
\usepackage{setspace}

\usepackage{wrapfig}
\usepackage{tikz}
\usepackage{quantikz}
\usepackage{dashbox}
\usepackage{mdframed}
\usepackage{marvosym}
\usepackage{pifont}
\usepackage{CJK}
\usepackage{url}

\usepackage[table,x11names]{xcolor}
\usepackage[most]{tcolorbox}
\tcbuselibrary{breakable}
\usetikzlibrary{decorations.pathreplacing, fit}

\definecolor{primaryblue}{HTML}{0066CC}
\definecolor{accentcyan}{HTML}{00D4AA}
\definecolor{warmorange}{HTML}{FF6B35}
\definecolor{deepgray}{HTML}{2C3E50}
\definecolor{lightgray}{HTML}{F8F9FA}
\definecolor{gradientstart}{HTML}{667eea}
\definecolor{gradientend}{HTML}{764ba2}

\definecolor{citecolor}{HTML}{0071bc}
\definecolor{citeblue}{RGB}{0, 113, 188}
\definecolor{linkcolor}{HTML}{9A4D92}
\definecolor{firebrick}{rgb}{0.698,0.133,0.133}

\definecolor{paleviolet}{HTML}{E1EEFC}
\definecolor{CarolinaUltraLight}{HTML}{E7F4FC}
\definecolor{lightgrey}{RGB}{247, 247, 247}
\definecolor{shadecolor}{HTML}{EFEFEF}
\definecolor{lightyellow}{rgb}{1.0, 0.95, 0.7}
\definecolor{lightblue}{rgb}{0.90, 0.95, 1.0}
\definecolor{light-gray}{gray}{0.95}

\definecolor{darkgrey}{rgb}{0.5, 0.5, 0.5}
\definecolor{darkgreen}{rgb}{0, 0.5, 0}
\definecolor{mydarkblue}{rgb}{0,0.08,0.45}
\definecolor{mydarkblue2}{rgb}{0.133, 0.133, 0.698}
\definecolor{echodrk}{HTML}{0099cc}
\definecolor{mymauve}{rgb}{0.58,0,0.82}
\definecolor{midnightblue}{rgb}{0.1,0.1,0.44}
\definecolor{oxfordblue}{rgb}{0.0,0.13,0.28}
\definecolor{prussianblue}{rgb}{0.0,0.19,0.33}
\definecolor{coolteal}{rgb}{0, 0.45, 0.45}
\definecolor{olive}{rgb}{0.1, 0.3, 0}
\definecolor{mypurple}{rgb}{0.5,0,0.5}
\definecolor{almond}{rgb}{0.94, 0.87, 0.8}

\definecolor{blue_ampEncoding}{HTML}{DAE8FC}
\definecolor{green_encoder}{HTML}{D5E8D4}
\definecolor{purple_decoder}{HTML}{E1D5E7}
\definecolor{yellow_measure}{HTML}{FFF2CC}
\definecolor{gray_block}{HTML}{F5F5F5}
\definecolor{pink_dru}{HTML}{FAD9D5}
\definecolor{orange_v}{HTML}{FAD7AC}

\definecolor{colorA}{rgb}{1,0,0}
\definecolor{colorB}{rgb}{0,0.3,1}
\definecolor{colorC}{rgb}{0.9,0.8,0.2}
\definecolor{colorD}{rgb}{0,0.65,0}
\definecolor{lesslightgray}{rgb}{0.5,0.5,0.5}
\definecolor{fundamental}{RGB}{55, 110, 111}
\definecolor{Gred}{RGB}{219, 50, 54}
\definecolor{ToCgreen}{RGB}{0, 128, 0}
\definecolor{Sepia}{RGB}{112, 66, 20}
\definecolor{Dblue}{rgb}{0,0.08,0.45}
\definecolor{Blue}{rgb}{0, 0, 0.8}
\definecolor{blue}{rgb}{0,0,1}
\definecolor{UNCblue!10}{rgb}{0.84,0.91,0.98}
\definecolor{RowAlt}{rgb}{0.98,0.98,0.99}

\definecolor{CarolinaBlue}{HTML}{7BAFD4}        
\definecolor{CarolinaLightBlue}{HTML}{B3D4E5}   
\definecolor{CarolinaUltraLight}{HTML}{E8F4F8}  
\definecolor{CarolinaText}{HTML}{1C2B33}        

\usepackage[pagebackref=true,breaklinks=true,colorlinks,hyperfootnotes=false]{hyperref}
\hypersetup{
  colorlinks,
  citecolor=citeblue,
  linkcolor=firebrick,
  urlcolor=firebrick
}
\usepackage[nameinlink,capitalize,noabbrev]{cleveref}

\titlespacing\section{0pt}{4pt plus 4pt minus 2pt}{-2pt plus 2pt minus 2pt}
\titlespacing\subsection{0pt}{2pt plus 4pt minus 2pt}{-2pt plus 2pt minus 2pt}
\titlespacing\subsubsection{0pt}{2pt plus 4pt minus 2pt}{-2pt plus 2pt minus 2pt}

\makeatletter
\def\th@remark{%
  \thm@headfont{\bfseries}%
  \normalfont 
  \thm@preskip\topsep \divide\thm@preskip\tw@
  \thm@postskip\thm@preskip
}
\makeatother

\theoremstyle{definition}

\tcbset{focithm/.style={
  enhanced, breakable, frame hidden,
  colback=black!5,
  width=\linewidth,
  enlarge left by=0pt, enlarge right by=0pt,
  boxsep=6pt, boxrule=0pt,
  left=4pt, right=4pt, top=1pt, bottom=1pt,
  arc=8pt,
  before skip=\topsep, after skip=\topsep
}}

\tcolorboxenvironment{theorem}{focithm}

\tcolorboxenvironment{lemma}{focithm}

\tcolorboxenvironment{corollary}{focithm}

\tcolorboxenvironment{proposition}{focithm}

\tcolorboxenvironment{definition}{focithm}

\tcolorboxenvironment{assumption}{focithm}

\tcolorboxenvironment{claim}{focithm}

\tcolorboxenvironment{problem}{focithm}

\tcolorboxenvironment{question}{focithm}

\newtcolorbox{titleblock}{
  enhanced,
  frame hidden,
  colback=CarolinaUltraLight,
  colframe=CarolinaUltraLight,
  boxrule=0pt,
  arc=10pt,
  left=14pt,
  right=14pt,
  top=14pt,
  bottom=14pt,
  width=\linewidth,
  before skip=12pt plus 4pt,
  after skip=12pt plus 4pt,
  grow to left by=1.5pt,
  grow to right by=1.5pt,
  before upper={
    \setlength{\parindent}{0cm}
    \setlength{\parskip}{0.5cm}
  }
}

\crefname{theorem}{Theorem}{Theorems}
\crefname{proposition}{Proposition}{Propositions}
\crefname{lemma}{Lemma}{Lemmas}
\crefname{corollary}{Corollary}{Corollaries}
\crefname{definition}{Definition}{Definitions}
\crefname{assumption}{Assumption}{Assumptions}
\crefname{remark}{Remark}{Remarks}
\crefname{problem}{Problem}{Problems}
\crefname{property}{Property}{property}
\crefname{question}{Question}{Questions}

\numberwithin{equation}{section}
\numberwithin{theorem}{section}
\numberwithin{proposition}{section}
\numberwithin{definition}{section}
\numberwithin{lemma}{section}
\numberwithin{assumption}{section}
\numberwithin{remark}{section}

\def\1{\bm{1}}

\makeatletter
\let\save@mathaccent\mathaccent
\newcommand*\if@single[3]{%
    \setbox0\hbox{${\mathaccent"0362{#1}}^H$}%
    \setbox2\hbox{${\mathaccent"0362{\kern0pt#1}}^H$}%
    \ifdim\ht0=\ht2 #3\else #2\fi
}
\newcommand*\rel@kern[1]{\kern#1\dimexpr\macc@kerna}
\newcommand*\widebar[1]{\@ifnextchar^{{\wide@bar{#1}{0}}}{\wide@bar{#1}{1}}}
\newcommand*\wide@bar[2]{\if@single{#1}{\wide@bar@{#1}{#2}{1}}{\wide@bar@{#1}{#2}{2}}}
\newcommand*\wide@bar@[3]{%
    \begingroup
    \def\mathaccent##1##2{%
        \let\mathaccent\save@mathaccent
        \if#32 \let\macc@nucleus\first@char \fi
        \setbox\z@\hbox{$\macc@style{\macc@nucleus}_{}$}%
        \setbox\tw@\hbox{$\macc@style{\macc@nucleus}{}_{}$}%
        \dimen@\wd\tw@
        \advance\dimen@-\wd\z@
        \divide\dimen@ 3
        \@tempdima\wd\tw@
        \advance\@tempdima-\scriptspace
        \divide\@tempdima 10
        \advance\dimen@-\@tempdima
        \ifdim\dimen@>\z@ \dimen@0pt\fi
        \rel@kern{0.6}\kern-\dimen@
        \if#31
        \overline{\rel@kern{-0.6}\kern\dimen@\macc@nucleus\rel@kern{0.4}\kern\dimen@}%
        \advance\dimen@0.4\dimexpr\macc@kerna
        \let\final@kern#2%
        \ifdim\dimen@<\z@ \let\final@kern1\fi
        \if\final@kern1 \kern-\dimen@\fi
        \else
        \overline{\rel@kern{-0.6}\kern\dimen@#1}%
        \fi
    }%
    \macc@depth\@ne
    \let\math@bgroup\@empty \let\math@egroup\macc@set@skewchar
    \mathsurround\z@ \frozen@everymath{\mathgroup\macc@group\relax}%
    \macc@set@skewchar\relax
    \let\mathaccentV\macc@nested@a
    \if#31
    \macc@nested@a\relax111{#1}%
    \else
    \def\gobble@till@marker##1\endmarker{}%
    \futurelet\first@char\gobble@till@marker#1\endmarker
    \ifcat\noexpand\first@char A\else
    \def\first@char{}%
    \fi
    \macc@nested@a\relax111{\first@char}%
    \fi
    \endgroup
    }
\makeatother
\let\bar\widebar

\def\va{{\bm{a}}} \def\vb{{\bm{b}}} \def\vc{{\bm{c}}} 
\def\ve{{\bm{e}}}   
   
\def\vm{{\bm{m}}}  \def\vo{{\bm{o}}} \def\vp{{\bm{p}}}
\def\vq{{\bm{q}}} \def\vr{{\bm{r}}}  
\def\vu{{\bm{u}}} \def\vv{{\bm{v}}} \def\vw{{\bm{w}}} \def\vx{{\bm{x}}}
 \def\vz{{\bm{z}}}

\DeclareMathAlphabet{\mathsfit}{\encodingdefault}{\sfdefault}{m}{sl}
\SetMathAlphabet{\mathsfit}{bold}{\encodingdefault}{\sfdefault}{bx}{n}

\newcommand{\R}{\mathbb{R}}

\newcommand{\E}{\mathbb{E}}

\let\tilde\widetilde
\let\hat\widehat

\renewcommand{\arraystretch}{1.15}
\makeatletter

\def\foci@authors{}
\def\foci@affil{}
\def\foci@email{}
\def\foci@project{}
\def\foci@date{\today}
\def\foci@authorsep{}

\newcommand{\fociauthor}[2]{%
  \g@addto@macro\foci@authors{\foci@authorsep\mbox{#1\textsuperscript{#2}}\gdef\foci@authorsep{,\ }}}
\newcommand{\fociaffil}[1]{\gdef\foci@affil{#1}}
\newcommand{\fociemail}[1]{\gdef\foci@email{#1}}
\newcommand{\fociproject}[1]{\gdef\foci@project{#1}}
\newcommand{\focidate}[1]{\gdef\foci@date{#1}}
\newcommand{\fociauthorlist}{\gdef\foci@authorsep{}\foci@authors}
\newcommand{\focicorr}{\textrm{\Letter}}

\def\foci@blfootnote{\gdef\@thefnmark{}\@footnotetext}
\newcommand{\focifootnotes}{%
  \foci@blfootnote{\textsuperscript{*}Equal contribution.\quad
    \textsuperscript{\Letter}Corresponding author: \texttt{\foci@email}}}

\newlength{\foci@mw}
\newcommand{\focidrawmark}[3]{%
  \setlength{\foci@mw}{#2}%
  \begin{scope}[shift={#3}, x={\foci@mw/564.03}, y={-\foci@mw/564.03}]
    \foreach \cx/\cy in {282.02/66.03, 282.02/222.08, 215.77/348.8, 348.73/348.8,
                         414.88/474.93, 149.16/474.93, 67.06/222.08, 496.99/222.08}
      \fill[#1] (\cx,\cy) circle[radius=42.04];
  \end{scope}}
\newcommand{\focimark}[2][black]{%
  \setlength{\foci@mw}{#2}%
  \begin{tikzpicture}[x={\foci@mw/564.03}, y={-\foci@mw/564.03}]
    \useasboundingbox (0,0) rectangle (564.03,540.96);
    \focidrawmark{#1}{#2}{(0,0)}
  \end{tikzpicture}}

\fancypagestyle{focititlepage}{\fancyhf{}\renewcommand{\headrulewidth}{0pt}}

\newcommand{\foci@headerline}{%
  \noindent
  \begin{minipage}[c]{0.5\linewidth}
    \includegraphics[height=0.58cm]{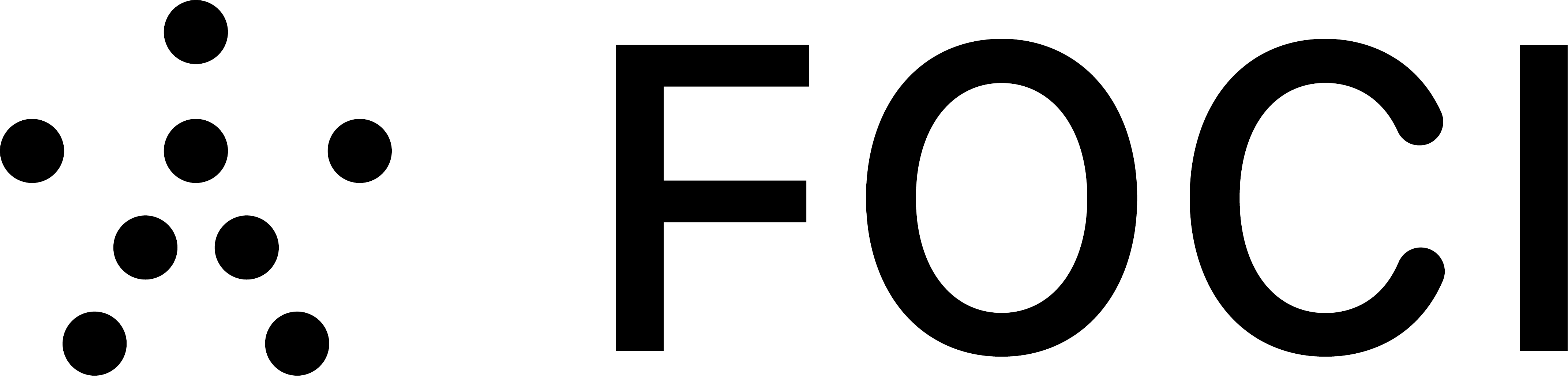}
  \end{minipage}%
  \begin{minipage}[c]{0.5\linewidth}
    \raggedleft\small\itshape\color{black!55}\foci@date
  \end{minipage}\par}

\newcommand{\foci@contactline}{%
  {\footnotesize\color{black!65}%
   \textbf{Email:}~\texttt{\foci@email}\qquad
   \textbf{Project:}~\url{\foci@project}\par}}

\newcommand{\foci@hairline}{\noindent\textcolor{black}{\rule{\linewidth}{0.5pt}}\par}

\newcommand{\focicoverCard}{%
  \thispagestyle{focititlepage}%
  \foci@headerline
  \vskip -0.5em
  \foci@hairline
  \vskip 0.9em
  \begin{tcolorbox}[enhanced, frame hidden, boxrule=0pt, arc=12pt,
      colback=black!4, width=\linewidth,
      left=18pt, right=18pt, top=18pt, bottom=16pt,
      grow to left by=2pt, grow to right by=2pt]
    \noindent
    \begin{minipage}[t]{\linewidth}
      {\setstretch{1.15}\LARGE\bfseries\raggedright On the Divergence of Accuracy and Mechanism Consistency in Time Series World Models
\unskip\par}
      \vskip 0.55em
      {\normalsize\fociauthorlist\par}
      \vskip 0.2em
      {\small\color{black!70}\foci@affil\par}
    \end{minipage}\par
    \vskip 0.9em
    {\small\setstretch{1.08}A time series world model (TSWM) takes the observed history of a controlled system and a
plan of future actions and exogenous inputs, and predicts how the state will respond.
Current approaches build such models as forecasters with the actions as covariates,
trained and evaluated on prediction error under the executed plan. Yet a world model is used to compare plans that were never executed, so its
response to a changed plan matters as much as its error, and this practice never tests
it. Two questions are therefore open: which design choices matter, and whether an
accurate forecaster responds to a changed plan the way the real system does. We address
both with a formalization of TSWMs and a benchmark. The formalization separates state,
actions and exogenous inputs, distinguishes continuous, mode and event actions, and
introduces mechanism consistency, a metric built on declared mechanisms, relations
between an action and a state variable whose direction is known, such as a vasopressor
raising blood pressure: it checks whether shifting an action moves the forecast in the
declared direction. The benchmark consolidates eight public datasets with real actions
from engineered infrastructure and clinical care, and varies the prediction space, the
plan fusion and the plan encoding across seven backbones and five seeds. First, a frozen
latent prediction space lowers MAE by $9.9\%$ over observation space and gated fusion at
the output by $12.7\%$ over concatenation at the input, both on average and on all eight
datasets, whereas encoding the plan over time changes the average MAE by at most $2.2\%$.
Second, prediction error and
mechanism consistency diverge: the configuration with the lowest error is at or below
chance in consistency on four of the five datasets with declared mechanisms, and no
design choice avoids this. Finally, directional supervision, a loss term that penalizes
the wrong-signed part of the response to a shifted action, raises consistency
significantly on penalized mechanisms with no change in MAE. Together they give
TSWMs a recipe: a frozen latent space and output-side fusion for accuracy, and a
training objective for mechanism consistency.
\unskip\par}
    \vskip 0.9em
    \foci@contactline
  \end{tcolorbox}
  \vskip 0.4em}

\newcommand{\focicover}{\focicoverCard}

\makeatother

\usepackage{needspace}
\hypersetup{hypertexnames=false}

\newcommand{\McNprior}{twenty-one}
\newcommand{\McNpriorAll}{twenty-two}
\newcommand{\McNpriorDecl}{twenty-four}

\newcommand{\McCfgMin}{0.46}
\newcommand{\McCfgMax}{0.59}

\newcommand{\McWwMin}{0.03}
\newcommand{\McWwMax}{0.21}

\newcommand{\McSelfcheck}{3.6\times 10^{-6}}
\newcommand{\McSelfcheckDuet}{1.1\times 10^{-3}}
\newcommand{\McPlJoint}{21$ of $506}
\newcommand{\McPlRho}{-0.86}
\newcommand{\McGhJoint}{49$ of $518}

\newcommand{\McDeadCells}{85}
\newcommand{\McDeadTotal}{3150}

\makeatletter
\renewcommand{\focifootnotes}{%
  \foci@blfootnote{\textsuperscript{\Letter}Corresponding author: \texttt{\foci@email}}}
\makeatother

\begin{document}

\fociauthor{Haochen Zhang}{1}
\fociauthor{Jiaheng Guo}{1}
\fociauthor{Zhen Xu}{1}
\fociauthor{Zachary Plotkin}{2}
\fociauthor{Nicholas Konz}{1}
\fociauthor{Zhen Tan}{3}
\fociauthor{Tianlong Chen}{2\,\focicorr}

\fociaffil{\textsuperscript{1}University of North Carolina at Chapel Hill\quad
  \textsuperscript{2}Foci Labs\quad
  \textsuperscript{3}Stevens Institute of Technology}
\fociemail{tianlong@cs.unc.edu}
\fociproject{https://github.com/HaochenZhang717/TSWM-Official-Repo}

\pagestyle{fancy}
\fancyhf{}
\renewcommand{\headrulewidth}{0.4pt}
\chead{\small\scshape On the Divergence of Accuracy and Mechanism Consistency in Time Series World Models
\unskip}
\cfoot{\small\thepage}

\focicover
\focifootnotes


\section{Introduction}
\label{sec:intro}

\begin{wrapfigure}{r}{0.44\columnwidth}
    \centering
    \vspace{-4mm}
    \includegraphics[width=\linewidth]{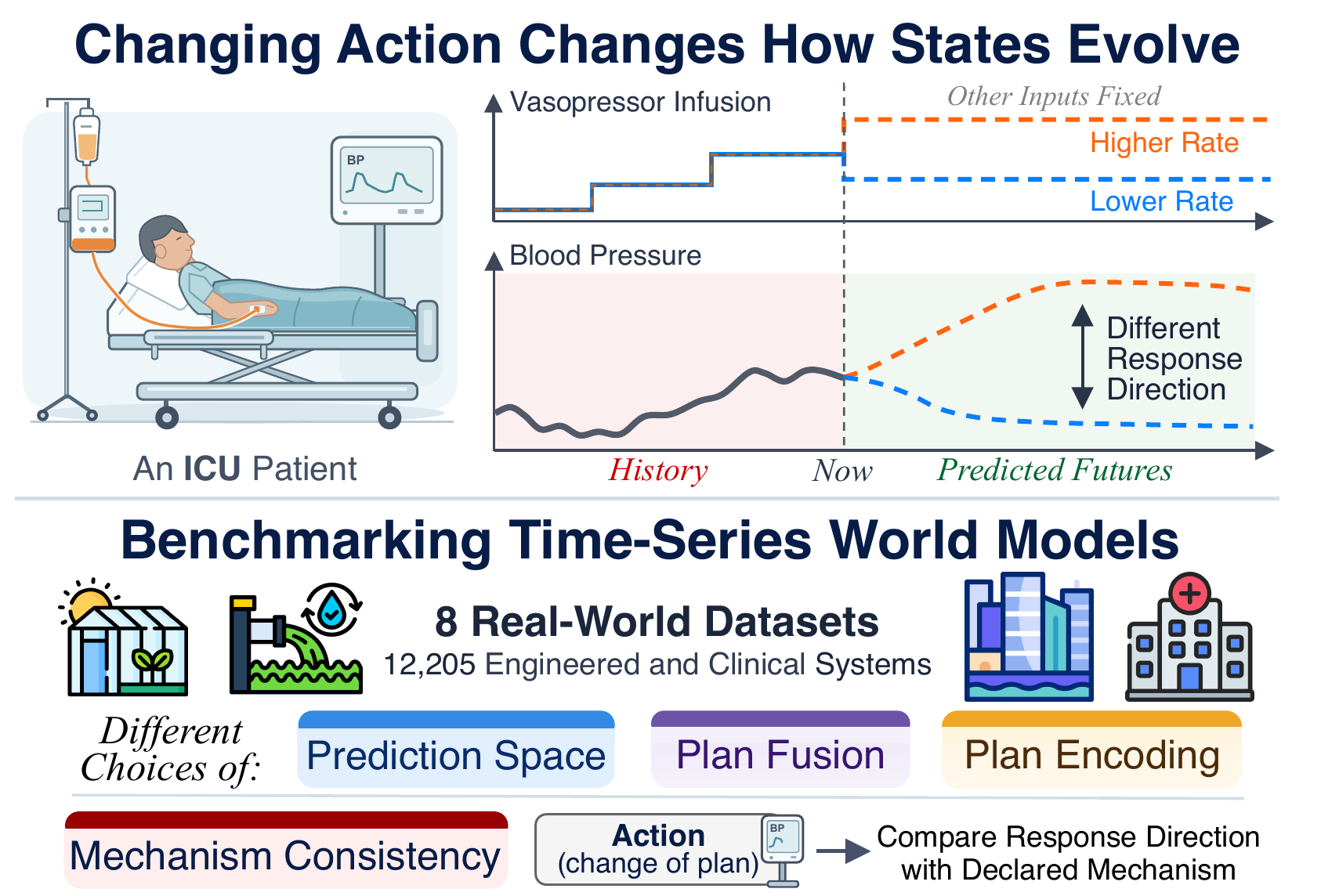}
    \vspace{-7mm}
    \caption{Two plans from the same history. We sweep three design choices and score mechanism consistency.}
    \label{fig:intro}
\end{wrapfigure}
Many operational and clinical decisions depend on how a multivariate time series will
evolve under different actions and exogenous inputs. For
example, in an intensive-care unit, vital signs such as arterial blood pressure are
held in an acceptable range by actions such as raising or lowering the infusion rate
of a vasopressor, and are moved as well by uncontrollable exogenous inputs. A model that takes the
observed history and a plan of future actions and exogenous inputs and predicts the response of the target is a world model in the
sense used in reinforcement learning and video generation
\citep{ha2018world,hafner2025dreamerv3,nvidia2025cosmos}; we
call it a time series world model (TSWM). Current practice builds it as a
forecasting model with actions and exogenous inputs as covariates
\citep{das2023tide,lim2021tft,olivares2023nbeatsx,arango2025chronosx,wang2024timexer},
and recent work has begun to call such models world models for time series
\citep{ding2025clarity,wang2026medicalwm,qi2026mobiwm}. Each of these works proposes
one design and compares it against few alternatives inside its own
architecture, so the design choices of a TSWM have not been investigated
systematically. By contrast, prior work on control problems and video generation has
examined world models along two such choices: the space in which the state is
predicted, observational \citep{nvidia2025cosmos} or a pretrained feature space
\citep{assran2025vjepa2,zhou2025dinowm}, and the way the action enters the model, by
concatenation \citep{hafner2019planet,zhou2025dinowm}, as interleaved tokens
\citep{assran2025vjepa2}, or as an additive embedding \citep{bruce2024genie}. Neither has been tested on time series with real actions.

In applications from multiple domains including retail demand, building control, and clinical prediction,
models trained for prediction accuracy alone have been found to respond incorrectly
to a change of action
\citep{crasson2024causal,kunz2023zalando,dinatale2022pcnn,keogh2024interventions}. A TSWM therefore needs a second property besides prediction accuracy, which we call mechanism consistency. A mechanism is a relation between an action and a
state variable whose direction is known, and a mechanism-consistent model responds to a
change of that action in that direction: a higher vasopressor infusion rate raises blood pressure,
so raising the rate in the input should raise the predicted pressure. This
property lets a world model answer what would happen under a plan that has not been
taken, so a mechanism-consistent TSWM can compare candidate plans before one is
carried out. However, existing benchmarks do not measure this. Forecasting benchmarks rarely
contain action-conditional prediction \citep{godahewa2021monash,aksu2024gifteval},
and the benchmarks that score the reaction to an intervention usually test models on the
dynamics of a simulator rather than of a real system
\citep{hafner2019planet,mu2026epibench,thumm2026dotime}.

Two desiderata are therefore unmet: a benchmark on time series with real actions that
compares the design choices of a TSWM systematically, and a formalization of the
TSWM with a metric of mechanism consistency, so that the comparison evaluates the
reaction to a change of action next to prediction error. We formalize the TSWM problem
and build a benchmark for it. The formalization assigns each channel a role, the target
as the state, what the controller sets as the actions, and everything else that moves
the state as an exogenous input; distinguishes three forms of action, a continuous
level such as a dose, a discrete mode such as heating or cooling, and a one-off event
such as a meal; and adds mechanism consistency as a metric next to prediction error
(Section~\ref{sec:setup}).

For dataset construction, we select eight public datasets with real
actions, four from engineered infrastructure, namely horticulture, building climate,
district heating, and water treatment, and four from clinical care, namely anesthesia,
two glucose-monitoring cohorts, and intensive care. Together they cover $12{,}205$
systems, sampling intervals from $2$ seconds to $30$ minutes, and $41$ action
channels of all three forms (Section~\ref{sec:datasets}). On these datasets, the
benchmark sweeps the three design components of a TSWM one at a time with the other
two fixed: the prediction space, the observation space or the latent space of a frozen
autoencoder (AE), a frozen variational autoencoder (VAE) or a joint-embedding predictive
architecture (JEPA) \citep{assran2023ijepa} trained with the forecaster; the plan
fusion, at the backbone input by concatenation or at its output by adapters such as
FiLM-0 \citep{perez2018film} and Gate \citep{dauphin2017glu}; and the plan encoding
over time, with or without a memory of the preceding steps. Each option is run with seven forecasting backbones and five
seeds (Section~\ref{sec:architecture-benchmark}). Our contributions are as
follows.
\begin{itemize}[leftmargin=*,nosep]
\item We formalize the TSWM problem and build a benchmark for it from eight public
datasets with real actions. Sweeping the three design choices shows that the frozen
AE latent space is the best prediction space while JEPA is worse than the observation
space, that FiLM-0 and Gate at the output are the best fusion methods and beat fusion at
the input, and that a memory of the preceding steps in the plan encoding gives no
consistent gain.
\item We find a contradiction between prediction accuracy and mechanism consistency:
the configuration selected for prediction accuracy is at or below chance in mechanism
consistency on four of the five datasets with declared mechanisms, and no design choice
escapes this.
\item We propose directional supervision, a soft penalty on the sign of the response to
a changed action, added to the training loss without changing the network, which raises mechanism
consistency above $0.99$ on engineered and clinical mechanisms alike without loss of
prediction accuracy (Section~\ref{sec:method}).
\end{itemize}

\section{Related work}
\label{sec:related}

\textbf{Forecasting with exogenous inputs and actions.}
Covariate-aware forecasters differ in where the covariates enter the model and how they
are combined with the target. Concatenation feeds them to the backbone input \citep{das2023tide};
gating learns how much of them to pass \citep{lim2021tft,han2026unica,qiu2026dag};
addition keeps them in a separate path summed with the target
\citep{olivares2023nbeatsx,arango2025chronosx,zhou2025crosslinear}; modulation
scales the target representation instead \citep{qin2025cora}; and cross-attention
lets target tokens query covariate tokens \citep{wang2024timexer,yamaguchi2026citras}, as
it does for treatment sequences \citep{melnychuk2022causal,bica2020crn}. Foundation models
that take covariates natively \citep{ansari2025chronos2,hoo2025tabpfnts} predict in
observation space and are outside our scope. How the effect of an input is spread over time is usually fixed by convention in
each field: a carryover kernel in marketing \citep{jin2017mmm}, a learned decay in clinical
series \citep{che2018grud}. Each paper compares its design against two or three
alternatives inside its own architecture, and none holds the backbone fixed while comparing
the families. We hold the backbone fixed and split the plan path into a plan
encoder and a fusion adapter that can be swapped independently.

\textbf{World models in other domains.}
A world model learns the state transition kernel $p(x_{t+1}\mid x_{\le t}, a_{\le t})$ and is judged by rolling it
forward. RL rolls it out in a latent space
\citep{ha2018world,hafner2019planet,hafner2025dreamerv3},
vision makes it a controllable video generator
\citep{yang2024unisim,nvidia2025cosmos} or a predictor in a
pretrained feature space \citep{assran2025vjepa2,zhou2025dinowm}, and medicine has begun
to use the name for treatment-conditioned disease trajectories
\citep{ding2025clarity,wang2026medicalwm}. Two choices recur: whether the encoder is
shaped by the dynamics, and whether the action enters by concatenation, interleaved
tokens, or an additive embedding. Time series forecasting has only recently moved into a
learned latent space \citep{yang2026latenttsf}, and its benchmarks do not score the
response to a change of action. We study both choices on time series with real actions
and add that score next to the forecast error.

\textbf{Constraining learned dynamics with prior knowledge.}
Prior knowledge enters a learned dynamics model as the residual of a governing equation in
the loss \citep{raissi2019pinn,karniadakis2021piml}, as a monotonicity constraint built
into the architecture \citep{kitouni2023monotonic,runje2023monotonic}, or
as a loss on the ranking of treatment effects in a hybrid mechanistic--neural model
\citep{zou2024hybrid}. The treatment-effect models above instead recover the effect from
observational data by correcting for confounding, and \citet{crasson2024causal} show that
forecast accuracy and causal accuracy can come apart. We impose the sign instead of learning it, because
our actions come from a controller that reacts to the state, and the observational sign is
often the reverse of the mechanism. Our penalty needs only the sign of a relation; it is a
term in the loss, leaves the network unchanged, and is checked by a
finite-difference intervention on the same held-out windows, reported alongside the
forecast error.


\section{Time Series World Model}
\label{sec:tswm}

\subsection{Problem Formalization}
\label{sec:setup}

We consider a controllable system observed at a fixed interval $\Delta t$ as a
discrete-time multivariate series whose channels play one of three roles. The action
$\vu_t\in\mathcal{U}$ collects the channels a controller sets directly, the only degrees
of freedom in the system. The exogenous input $\ve_t\in\mathcal{E}\subset\R^{d_e}$
collects channels that influence the system but that no controller sets. The state
$\vx_t\in\mathcal{X}\subset\R^{d_x}$ collects the regulated quantities, each scaled to
$[-1,1]$, and is what the model predicts. For instance, in a greenhouse, heating-pipe temperature and vent
aperture are actions, outdoor temperature and irradiance are exogenous inputs, and indoor
temperature and CO$_2$ concentration are the state.

An action has three components, continuous, mode and event,
$\vu_t=\big(\vu^{\mathrm{c}}_t,\;\vu^{\mathrm{m}}_t,\;u^{\mathrm{e}}_t\big)$. Subscripts
give the time step and, where present, the channel; bold denotes all channels of a
component; $[n]:=\{1,\dots,n\}$.
\textbf{(i)} The continuous component holds $d_c$ channels the controller may set
anywhere in a bounded range, e.g.\ an infusion rate or a temperature setpoint. A channel
whose declared range starts at an off position, e.g.\ a closed vent or a stopped infusion,
is scaled to $[0,1]$ so that zero is exactly off, and every other channel to $[-1,1]$;
thus $\vu^{\mathrm{c}}_t\in[-1,1]^{d_c}$.
\textbf{(ii)} The mode component holds $d_m$ channels of operating conditions the
controller selects, e.g.\ a heat pump in heating or in cooling. Channel $j\in[d_m]$
takes values in a finite unordered set $\mathcal{M}_j$ of size $K_j$, which we index by
$[K_j]$, so $\vu^{\mathrm{m}}_t\in\prod_{j=1}^{d_m}[K_j]$.
\textbf{(iii)} The event component holds interventions that occur at an instant, e.g.\ a
drug push. With $\mathcal{A}$ the finite set of event types in a dataset, the value at a
step is the subset of $\mathcal{A}$ administered then, empty at most steps; it is a single
channel $u^{\mathrm{e}}_t\in\mathcal{V}$, where $\mathcal{V}\subseteq2^{\mathcal{A}}$
consists of $\varnothing$ and the subsets observed in the training set.
The action space is the Cartesian product
$\mathcal{U}=[-1,1]^{d_{c}}\times\prod_{j=1}^{d_{m}}[K_j]\times\mathcal{V}$.

Fix a context length $L$ and a horizon $H$, and write $\vx_{i:j}$ for the block of state
values from step $i$ to step $j$ inclusive, likewise for $\ve$ and $\vu$. A window at
time $t$ is $W_t=(h,a,y)$ with history $h=\vx_{t-L+1:t}$, plan
$a=(\ve_{t-L+1:t+H},\vu_{t-L+1:t+H})$ and outcome $y=\vx_{t+1:t+H}$. Given training
windows $\{(h_n,a_n,y_n)\}_{n=1}^{N}$ drawn from an unknown joint distribution $P$, the
aim is to learn a map
$f_\theta:\mathcal{X}^{L}\times(\mathcal{E}\times\mathcal{U})^{L+H}\to\mathcal{X}^{H}$
from the history and the plan to the outcome that minimizes the risk
$\E_{(h,a,y)\sim P}\big[\|f_{\theta}(h,a)-y\|_2^{2}\big]$.

We evaluate a model on two metrics. Prediction error is measured on held-out
data and reported as MAE in observation space. Mechanism consistency asks
whether $f_\theta$ moves the state the way a known mechanism dictates once the plan is
changed. A mechanism pairs a continuous action channel $i\in[d_c]$ with a state channel
$c\in[d_x]$ under a sign $s_{i\to c}\in\{-1,+1\}$ declared from domain knowledge, $+1$
if raising channel $i$ drives channel $c$ up and $-1$ if it drives it down; opening a
greenhouse vent lowers the air temperature, so $s_{i\to c}=-1$. To test the model, we
shift that one action channel, which yields a new plan, and read off the direction in
which the forecast moves. Let $F_i$ be the empirical distribution function of
$u^{\mathrm{c}}_{t,i}$ over all held-out windows and horizon steps. For
$\delta\in(-1,1)\setminus\{0\}$, the quantile shift $\tau_{i,\delta}$ sets
$(\tau_{i,\delta}a)^{\mathrm{c}}_{t+k,i}
=F_{i}^{-1}\big(\min\{\max\{F_{i}(u^{\mathrm{c}}_{t+k,i})+\delta,0\},1\}\big)$
at every horizon step $k\in[H]$ and leaves every other entry of $h$ and $a$ unchanged.
Writing $[\,\cdot\,]_{k,c}$ for the
entry of a forecast at horizon step $k$ and state channel $c$, the predicted effect is
$\Delta^{i\to c}_{\theta}(h,a;\delta)=\tfrac{1}{H}\sum_{k=1}^{H}[f_{\theta}(h,\tau_{i,\delta}a)-f_{\theta}(h,a)]_{k,c}$
and mechanism consistency is the rate at which its sign agrees with the mechanism,
$\mathrm{MC}_{\theta}(i\!\to\!c;\delta)=\Pr_{(h,a)}[\operatorname{sign}\Delta^{i\to c}_{\theta}(h,a;\delta)=\operatorname{sign}(\delta)\,s_{i\to c}]$,
the probability running over the held-out windows on which the shift changes the plan,
$\tau_{i,\delta}a\neq a$, and the forecast moves,
$|\Delta^{i\to c}_{\theta}|>10^{-3}\sigma_c$ with $\sigma_c$ the median absolute
deviation of state channel $c$ over the held-out outcomes; other windows are excluded from
the count. A mechanism declared under a mode
condition, e.g.\ only while a heat pump is heating, is scored over the windows that
spend at least half of their steps in that mode. $\mathrm{MC}_\theta=1$ is perfect
agreement with the mechanism and $1/2$ is chance.

\subsection{Model Design Space}
\label{sec:design-space}

\begin{figure}[t]
    \centering
    \includegraphics[width=\textwidth]{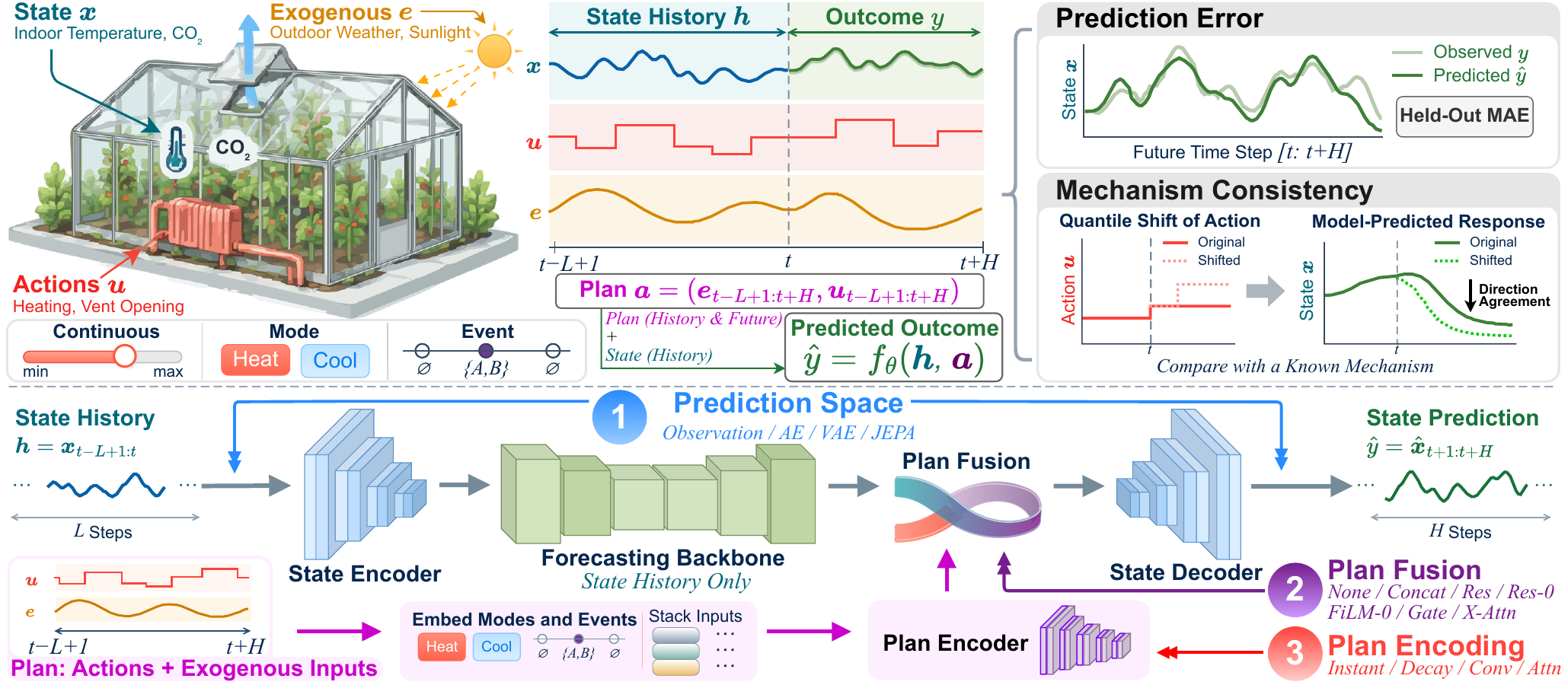}
    \vspace{-3mm}
    \caption{Overview of the formalization and the model. Top: the channel roles, the
    window of history, plan and outcome, and the two metrics (Section~\ref{sec:setup}).
    Bottom: the model and its three design choices (Section~\ref{sec:design-space});
    the concatenation baseline conditions the backbone instead
    (Appendix~\ref{app:design-space}).}
    \label{fig:benchmark-architecture}
    \vspace{-3mm}
\end{figure}

Figure~\ref{fig:benchmark-architecture} shows the model: a state encoder maps the
state history into a prediction space, a backbone forecasts there from the
history alone, and a fusion module adjusts that plan-blind forecast with the
encoded plan, before a decoder returns to the state channels. The colored blocks are the three design choices.
In the notation of Section~\ref{sec:setup}, write $\va_s\in\R^{C}$ for step $s$ of the
plan as the model receives it, the exogenous input $\ve_s$ and the continuous actions
$\vu^{\mathrm{c}}_s$ stacked with the embeddings of the mode indices
$\vu^{\mathrm{m}}_s$ and of the event value $u^{\mathrm{e}}_s$. The plan encoder maps
$\va_{t-L+1:t+H}$ causally to $\tilde\va_{t-L+1:t+H}$ of the same shape. The state
encoder maps every state $\vx_s$ to a latent $\vz_s$ (the identity in observation
space), and the backbone maps the latent history $\vz_{t-L+1:t}$ to the plan-blind
forecast $\hat\vz_{t+1:t+H}$. The fusion method maps $\hat\vz_{t+k}$ and $\tilde\va$
to the adjusted forecast $\vz'_{t+k}$, which the decoder maps to the predicted state
$\hat\vx_{t+k}$. Tables~\ref{tab:fusion-formulas} and~\ref{tab:encoding-formulas}
in Appendix~\ref{app:design-space} give the formula of every option.
\textbf{(1) Prediction space}: the observation space, the frozen latent space of an
autoencoder (\textsc{AE}) or variational autoencoder (\textsc{VAE}) pretrained
on the state channels, or a \textsc{JEPA} trained jointly with the forecaster.
\textbf{(2) Plan fusion}: \textsc{None} discards the plan, \textsc{Concat}
feeds it into the backbone, and \textsc{Res}, \textsc{Res-0}, \textsc{FiLM-0},
\textsc{Gate} and \textsc{X-attn} keep the backbone plan-blind and adjust its
forecast at each horizon step by a residual, a zero-initialized residual,
feature-wise scaling and shifting, a gated residual, or cross-attention over
the plan. \textbf{(3) Plan encoding}: a causal transform of the plan before
fusion, either the identity (\textsc{Instant}) or one that lets each step depend
on the preceding steps through a learned exponential moving average
(\textsc{Decay}), dilated convolutions (\textsc{Conv}) or self-attention
(\textsc{Attn}), all three initialized as the identity map.

\begingroup
\raggedbottom
\section{Benchmarking the Design Choices}
\label{sec:architecture-benchmark}

\subsection{Datasets}
\label{sec:datasets}

\begin{table}[t]
\caption{The eight datasets, four engineered and four clinical. Systems are counted train/held-out; windows are held-out windows
at $L\!=\!256$, $H\!=\!16$, stride $80$. The action columns are the three factors of
$\mathcal{U}$: $d_c$ continuous channels, $d_m$ mode channels with their cardinalities
$K_j$, and the number of event types $|\mathcal{A}|$.}
\label{tab:datasets}
\centering
\small
\setlength{\tabcolsep}{3pt}
\begin{tabular}{llcccccccc}
\toprule
& & & \multicolumn{2}{c}{scale} & \multicolumn{2}{c}{series dim.} & \multicolumn{3}{c}{actions} \\
\cmidrule(lr){4-5}\cmidrule(lr){6-7}\cmidrule(lr){8-10}
Dataset & Domain & $\Delta t$ & systems & windows & state & exog. & cont. & mode ($K$) & event \\
\midrule
Greenhouse    & horticulture      & 5\,min  & 5/1             & 163      & 4 & 8  & 7  & --        & -- \\
PleiaData     & building HVAC     & 10\,min & 50/12           & 7{,}548  & 2 & 8  & 1  & 2 (2,6)   & -- \\
PREDIST       & district heating  & 10\,min & 46/12           & 6{,}109  & 5 & 2  & 2  & 3 (2,3,2) & -- \\
Wastewater    & water treatment   & 2\,min  & 1 plant         & 1{,}311  & 3 & 2  & 3  & 2 (2,2)   & -- \\
\midrule
VitalDB       & anesthesia       & 2\,s    & 2{,}754/689     & 40{,}193 & 4 & -- & 2  & --        & -- \\
CGMacros      & metabolic (CGM)   & 1\,min  & 36/9            & 1{,}736  & 1 & -- & -- & --        & 4 \\
Shanghai      & metabolic (CGM)   & 15\,min & 90/22           & 214      & 1 & -- & -- & --        & 31 \\
MIMIC-Cardio  & critical care     & 30\,min & 6{,}782/1{,}696 & 8{,}370  & 6 & -- & 16 & --        & 2 \\
\midrule
\multicolumn{3}{l}{\emph{eight datasets}} & \emph{12{,}205} & \emph{65{,}644} & \emph{26} & \emph{20} & \emph{31} & \emph{7} & \emph{37} \\
\bottomrule
\end{tabular}

\end{table}

Table~\ref{tab:datasets} summarizes the eight public datasets. We selected them
because they record actions as defined in Section~\ref{sec:setup}. The four
engineered datasets come from horticulture, building HVAC, district heating and
wastewater treatment; the four clinical datasets come from anesthesia, two
continuous-glucose-monitoring studies and critical care. The engineered datasets have
exogenous inputs, and three of them also have mode actions; the clinical datasets have
neither, but they are the only datasets with event actions. In total there are $31$
continuous action channels, $7$ mode channels and $3$ event channels covering $37$
event types, so each of the three action components is present in at least two
datasets. Sampling intervals range from 2\,s to 30\,min, the number of systems from $1$
to $8{,}478$, and the state dimension is at most $6$ in every dataset. Each channel is
min--max scaled to the range given in Section~\ref{sec:setup}. The scaling bounds are
fixed per dataset and shared by all methods; they are taken from
instrument or physiological limits where such limits exist and from percentiles of the
data otherwise. Mode and event channels are represented as integer indices into an
embedding table, one index per condition in $\mathcal{M}_j$ or per set in
$\mathcal{V}$, plus one reserved index for missing codes and for sets not seen in
training. Appendix~\ref{app:datasets} gives the channel roles, scaling ranges and split
of each dataset.

\begin{table}[t]
\caption{The \McNprior{} declared mechanisms. A plus sign means that raising
the action raises the state, a minus sign that it lowers it; the PleiaData mechanisms hold in
the named mode. BIS, bispectral index; MAP, mean arterial pressure; FiO$_2$, inspired
oxygen fraction.}
\label{tab:declared-mechanisms}
\centering
\small
\setlength{\tabcolsep}{2pt}

\begin{minipage}[t]{0.47\textwidth}
\vspace{0pt}
\centering
\begin{tabular*}{\linewidth}{@{\extracolsep{\fill}}lllc@{}}
\toprule
Dataset & Action & State & Sign \\
\midrule
\multirow{4}{*}{Greenhouse}
& heating-pipe temp. & air temp. & $+$ \\
& leeward vent & air temp. & $-$ \\
& windward vent & air temp. & $-$ \\
& CO$_2$ dosing & CO$_2$ level & $+$ \\
\cmidrule(lr){1-4}
\multirow{4}{*}{VitalDB}
& propofol & BIS & $-$ \\
& propofol & MAP & $-$ \\
& remifentanil & heart rate & $-$ \\
& remifentanil & MAP & $-$ \\
\cmidrule(lr){1-4}
\multirow{2}{*}{PleiaData}
& setpoint (heating) & indoor temp. & $+$ \\
& setpoint (cooling) & indoor temp. & $-$ \\
\bottomrule
\end{tabular*}
\end{minipage}\hfill
\begin{minipage}[t]{0.52\textwidth}
\vspace{0pt}
\centering
\begin{tabular*}{\linewidth}{@{\extracolsep{\fill}}lllc@{}}
\toprule
Dataset & Action & State & Sign \\
\midrule
\multirow{3}{*}{PREDIST}
& valve position & flow & $+$ \\
& valve position & heat power & $+$ \\
& supply setpoint & supply temp. & $+$ \\
\cmidrule(lr){1-4}
\multirow{8}{*}{MIMIC-Cardio}
& norepinephrine & systolic pressure & $+$ \\
& norepinephrine & mean pressure & $+$ \\
& phenylephrine & systolic pressure & $+$ \\
& phenylephrine & mean pressure & $+$ \\
& vasopressin & mean pressure & $+$ \\
& dobutamine & heart rate & $+$ \\
& dopamine & heart rate & $+$ \\
& FiO$_2$ & oxygen saturation & $+$ \\
\bottomrule
\end{tabular*}
\end{minipage}
\end{table}

Computing mechanism consistency requires a set of declared mechanisms. Each
mechanism specifies a continuous action channel, a state channel that it affects, and
the sign of the effect. Table~\ref{tab:declared-mechanisms} lists the \McNprior{}
mechanisms used in this paper. All of them were fixed before any model was trained or
evaluated; in this section they are used only for evaluation, and in
Section~\ref{sec:method} a subset of them enters the training objective. For the
engineered datasets, the signs are taken from the documented operation of each
system: for example, a hotter heating pipe raises the greenhouse air temperature,
opening a vent lowers it, opening the control valve increases flow and heat power, and
the controlled temperature follows its setpoint. For the clinical datasets, the signs
are taken from the pharmacology of each drug: propofol deepens hypnosis and lowers
blood pressure, remifentanil lowers heart rate and blood pressure, vasopressors raise
blood pressure, inotropes raise heart rate, and a higher inspired oxygen fraction
raises oxygen saturation. The two PleiaData mechanisms depend on the operating mode:
raising the setpoint increases the indoor temperature in heating mode and decreases it
in cooling mode. Only five of the eight datasets are used for this metric. The two
glucose-monitoring datasets have only event actions, for which the quantile shift of
Section~\ref{sec:setup} is not defined, and the single mechanism of Wastewater is
excluded for a reason given in Appendix~\ref{app:mc-declared}.

\subsection{Experimental Setup}
\label{sec:benchmark-setup}

We run three experiments, each of which varies one of the design choices in
Section~\ref{sec:design-space} while fixing the other two. Experiment~1 compares the
prediction spaces with \textsc{Concat} fusion and \textsc{Instant} plan encoding.
Experiment~2 compares the fusion methods with the \textsc{AE} prediction space, which
Experiment~1 selects, and \textsc{Instant} plan encoding. Experiment~3 compares the
plan encodings with the \textsc{AE} prediction space and with each of \textsc{Gate}
and \textsc{FiLM-0}, the two best fusion methods in Experiment~2. Every configuration
is trained with seven backbones, TimeXer~\citep{wang2024timexer},
TiDE~\citep{das2023tide}, DUET~\citep{qiu2025duet},
PatchTST~\citep{nie2023patchtst}, TimeKAN~\citep{huang2025timekan},
CrossLinear~\citep{zhou2025crosslinear} and Amplifier~\citep{fei2025amplifier}, on all
eight datasets with five random seeds, under a single training protocol
(Table~\ref{tab:hyperparams}) and the fixed data split of Appendix~\ref{app:datasets}.
For each run we keep the checkpoint with the lowest validation loss and report its
validation MAE on the scaled state channels as the prediction error. The same
checkpoints are then evaluated for mechanism consistency, without any retraining,
using the $\mathrm{MC}$ metric of Section~\ref{sec:setup} with $|\delta|=0.1$ in both
directions on the \McNprior{} mechanisms in Table~\ref{tab:declared-mechanisms}.
Appendices~\ref{app:design-space} to~\ref{app:grids} give the formulas and
implementation of all options, hyperparameters, statistical tests, extended results,
an auxiliary-loss and rollout diagnostic, and the complete result grids.

\begin{table}[t]
    \caption{Validation MAE of the three sweeps. Mean $\pm$ std over the
    $35$ backbone--seed runs of one dataset and one option ($70$ for encoding, which
    pools \textsc{Gate} and \textsc{FiLM-0}); the std spans backbones. The fixed
    choices of each sweep are given in Section~\ref{sec:benchmark-setup}. Bold is the
    lowest mean of each block, Rank is the mean rank over dataset--backbone cells, and
    the shaded row has the best rank of its block. Lower is better.}
    \label{tab:sweeps}
    \centering
    \small
    \setlength{\tabcolsep}{2.5pt}
    \resizebox{\textwidth}{!}{
\begin{tabular}{llccccccccc}
\toprule
 & Option & Greenhouse & PleiaData & PREDIST & Wastewater & VitalDB & CGMacros & Shanghai & MIMIC-Cardio & Rank \\
\midrule
\multirow{4}{*}{\rotatebox[origin=c]{90}{Space}} & \textsc{Observation} & .0821$_{\pm.0269}$ & .0476$_{\pm.0306}$ & .1866$_{\pm.0099}$ & .1718$_{\pm.0116}$ & .0377$_{\pm.0090}$ & .0131$_{\pm.0054}$ & .1330$_{\pm.0132}$ & .0562$_{\pm.0029}$ & 2.52 \\
 & \cellcolor{gray!15}\textsc{AE} & \cellcolor{gray!15}\textbf{.0761}$_{\pm.0200}$ & \cellcolor{gray!15}\textbf{.0337}$_{\pm.0252}$ & \cellcolor{gray!15}\textbf{.1821}$_{\pm.0047}$ & \cellcolor{gray!15}.1695$_{\pm.0123}$ & \cellcolor{gray!15}\textbf{.0329}$_{\pm.0026}$ & \cellcolor{gray!15}.0114$_{\pm.0049}$ & \cellcolor{gray!15}.1249$_{\pm.0096}$ & \cellcolor{gray!15}.0552$_{\pm.0021}$ & \cellcolor{gray!15}\textbf{1.93} \\
 & \textsc{VAE} & .0956$_{\pm.0134}$ & .0339$_{\pm.0244}$ & .1836$_{\pm.0054}$ & \textbf{.1672}$_{\pm.0106}$ & .0336$_{\pm.0039}$ & \textbf{.0105}$_{\pm.0025}$ & \textbf{.1206}$_{\pm.0048}$ & \textbf{.0552}$_{\pm.0020}$ & 1.95 \\
 & \textsc{JEPA} & .1041$_{\pm.0271}$ & .0607$_{\pm.0359}$ & .2143$_{\pm.0327}$ & .1824$_{\pm.0266}$ & .0403$_{\pm.0112}$ & .0270$_{\pm.0233}$ & .1418$_{\pm.0208}$ & .0621$_{\pm.0095}$ & 3.61 \\
\midrule
\multirow{7}{*}{\rotatebox[origin=c]{90}{Fusion}} & \textsc{None} & .0751$_{\pm.0364}$ & .0292$_{\pm.0172}$ & .1858$_{\pm.0219}$ & .1928$_{\pm.0085}$ & .0396$_{\pm.0187}$ & .0174$_{\pm.0187}$ & .1257$_{\pm.0155}$ & .0608$_{\pm.0174}$ & 4.89 \\
 & \textsc{Concat} & .0761$_{\pm.0200}$ & .0337$_{\pm.0252}$ & .1821$_{\pm.0047}$ & .1695$_{\pm.0123}$ & .0329$_{\pm.0026}$ & .0114$_{\pm.0049}$ & .1249$_{\pm.0096}$ & .0552$_{\pm.0021}$ & 5.70 \\
 & \textsc{Res} & .0651$_{\pm.0041}$ & .0217$_{\pm.0112}$ & .1644$_{\pm.0019}$ & .1503$_{\pm.0065}$ & .0314$_{\pm.0008}$ & \textbf{.0086}$_{\pm.0003}$ & .1170$_{\pm.0030}$ & .0544$_{\pm.0007}$ & 3.43 \\
 & \textsc{Res-0} & .0652$_{\pm.0041}$ & .0216$_{\pm.0111}$ & .1642$_{\pm.0021}$ & .1509$_{\pm.0067}$ & .0315$_{\pm.0009}$ & .0087$_{\pm.0004}$ & .1166$_{\pm.0028}$ & .0544$_{\pm.0007}$ & 3.66 \\
 & \textsc{FiLM-0} & .0642$_{\pm.0040}$ & \textbf{.0203}$_{\pm.0106}$ & .1645$_{\pm.0023}$ & \textbf{.1489}$_{\pm.0053}$ & .0314$_{\pm.0009}$ & .0087$_{\pm.0004}$ & .1169$_{\pm.0028}$ & .0545$_{\pm.0007}$ & 2.91 \\
 & \cellcolor{gray!15}\textsc{Gate} & \cellcolor{gray!15}\textbf{.0632}$_{\pm.0037}$ & \cellcolor{gray!15}.0205$_{\pm.0104}$ & \cellcolor{gray!15}.1641$_{\pm.0018}$ & \cellcolor{gray!15}.1500$_{\pm.0064}$ & \cellcolor{gray!15}\textbf{.0313}$_{\pm.0009}$ & \cellcolor{gray!15}.0086$_{\pm.0004}$ & \cellcolor{gray!15}\textbf{.1165}$_{\pm.0032}$ & \cellcolor{gray!15}\textbf{.0543}$_{\pm.0006}$ & \cellcolor{gray!15}\textbf{2.52} \\
 & \textsc{X-attn} & .0764$_{\pm.0229}$ & .0203$_{\pm.0100}$ & \textbf{.1567}$_{\pm.0083}$ & .1617$_{\pm.0122}$ & .0323$_{\pm.0008}$ & .0107$_{\pm.0010}$ & .1236$_{\pm.0049}$ & .0548$_{\pm.0011}$ & 4.89 \\
\midrule
\multirow{4}{*}{\rotatebox[origin=c]{90}{Encoding}} & \textsc{Instant} & \textbf{.0637}$_{\pm.0038}$ & .0204$_{\pm.0104}$ & .1643$_{\pm.0021}$ & .1495$_{\pm.0059}$ & .0313$_{\pm.0009}$ & \textbf{.0087}$_{\pm.0004}$ & .1167$_{\pm.0030}$ & .0544$_{\pm.0007}$ & 2.79 \\
 & \textsc{Decay} & .0638$_{\pm.0041}$ & .0208$_{\pm.0104}$ & .1597$_{\pm.0021}$ & .1417$_{\pm.0089}$ & \textbf{.0313}$_{\pm.0009}$ & .0087$_{\pm.0005}$ & .1168$_{\pm.0031}$ & \textbf{.0544}$_{\pm.0007}$ & 2.40 \\
 & \cellcolor{gray!15}\textsc{Conv} & \cellcolor{gray!15}.0641$_{\pm.0037}$ & \cellcolor{gray!15}.0187$_{\pm.0086}$ & \cellcolor{gray!15}\textbf{.1570}$_{\pm.0018}$ & \cellcolor{gray!15}\textbf{.1388}$_{\pm.0061}$ & \cellcolor{gray!15}.0314$_{\pm.0009}$ & \cellcolor{gray!15}.0087$_{\pm.0004}$ & \cellcolor{gray!15}\textbf{.1163}$_{\pm.0030}$ & \cellcolor{gray!15}.0544$_{\pm.0006}$ & \cellcolor{gray!15}\textbf{2.06} \\
 & \textsc{Attn} & .0638$_{\pm.0039}$ & \textbf{.0172}$_{\pm.0068}$ & .1591$_{\pm.0029}$ & .1471$_{\pm.0063}$ & .0315$_{\pm.0007}$ & .0093$_{\pm.0005}$ & .1174$_{\pm.0033}$ & .0545$_{\pm.0006}$ & 2.74 \\
\bottomrule
\end{tabular}
}
\end{table}

\subsection{Results}
\label{sec:benchmark-results}

\label{sec:prediction-space-results}
\textbf{Prediction space.} The top block of Table~\ref{tab:sweeps} compares the four
prediction spaces, pooling the same backbones and seeds for each dataset. The two
frozen latent spaces perform best. \textsc{AE} has the lowest mean rank and beats
observation-space prediction on all eight datasets. \textsc{VAE} is close to
\textsc{AE}, within $3\%$ on average (Appendix~\ref{app:sweeps}).
The jointly trained \textsc{JEPA} is worse than observation-space prediction. We
therefore fix the prediction space to \textsc{AE} in the remaining experiments. A
likely reason is the frozen, pretrained latent space itself: \textsc{JEPA} also predicts
in a latent space, but its encoder is trained jointly, and it is worse than observation
space.

\label{sec:fusion-results}
\textbf{Plan fusion.} The middle block of Table~\ref{tab:sweeps} compares the seven
fusion methods under the same \textsc{AE} representation. The two modulation methods
work best: \textsc{Gate} and \textsc{FiLM-0} have the best mean ranks, both beat
\textsc{Concat} and \textsc{None} on all eight datasets, and they differ by less than
$1\%$ on average. The residual methods also do well, and cross-attention has no
consistent advantage over these simpler adapters. Adjusting the predicted
representation with the plan is therefore an effective way to bring in action and
exogenous information. We keep both \textsc{Gate} and \textsc{FiLM-0} for the encoding
experiment; detailed comparisons are in Appendix~\ref{app:sweeps}.

\label{sec:encoding-results}
\textbf{Plan encoding.} The bottom block of Table~\ref{tab:sweeps} tests whether
letting the encoding of each step depend on the preceding steps improves prediction.
\textsc{Conv} has the best mean rank and a small average improvement over
\textsc{Instant}, $2.2\%$, but the gain holds on only four of the eight datasets
(Appendix~\ref{app:sweeps}). \textsc{Decay}
and \textsc{Attn} give no consistent improvement either. A memory of the preceding
steps can therefore help on individual datasets, but we find no general benefit beyond
the history summaries that the fusion module already receives.

\label{sec:mechanism-consistency}
\textbf{Mechanism consistency.} Figure~\ref{fig:sec4-mechanism} reports the
$\mathrm{MC}$ metric of every design choice on the five datasets. No design choice does
well: averaged over the five datasets, all choices score between $\McCfgMin$ and
$\McCfgMax$, close to the chance level of $0.5$.
PREDIST is the easy case: its mechanisms are direct actuator effects, a valve on the
flow it throttles and a setpoint on the temperature the controller tracks, and the
training data already move in the declared direction, so fitting the training data is
enough to get the sign right, and every choice is consistent there. On the other four datasets all choices are near or below
chance, and on the two clinical datasets the models even learn the opposite sign. A
likely reason is that the drugs are given in response to the state: a vasopressor is
given when blood pressure is low and propofol when anesthesia is too shallow, so in the
training data a higher dose coincides with the condition it is meant to correct, and a
model that fits the training data may learn this association instead of the effect of
the drug. The choice with the lowest MAE on each dataset, boxed in
Figure~\ref{fig:sec4-mechanism}, is no more consistent than the others, so training on
forecasting error alone does not give a mechanism-consistent world model;
Section~\ref{sec:method} addresses this.

\begin{figure}[t]
    \centering
    \includegraphics[width=0.92\textwidth]{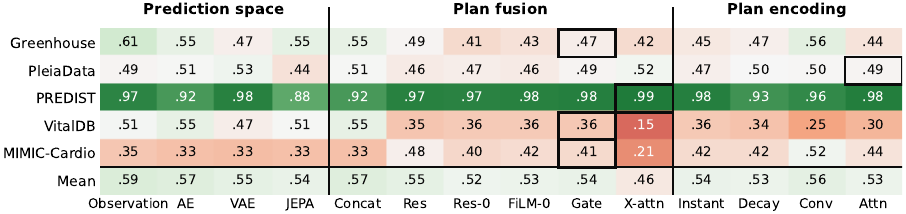}
    \caption{Mechanism consistency of the three sweeps. $\mathrm{MC}$ at
    $\delta=\pm0.1$, averaged over both directions, the declared mechanisms and the
    runs of Table~\ref{tab:sweeps}. White is chance, green agrees with the declared
    sign, red opposes it. Boxes mark the lowest-MAE choice of each row. \textsc{None}
    is omitted because its forecast does not move. Numbers are in
    Appendix~\ref{app:mechanism-consistency}.}
    \label{fig:sec4-mechanism}
\end{figure}

\par
\endgroup


\Needspace{10\baselineskip}
\section{Learning Mechanism-Consistent Responses}
\label{sec:method}

Section~\ref{sec:mechanism-consistency} shows that the forecasting objective
supervises $f_\theta(h,a)$ only under the plan that was carried out and provides no
target under an alternative plan. We add this supervision with a method we call
directional supervision. Given the same history, we compare
forecasts under the executed and the perturbed plan and penalize the part of
their difference that opposes a declared mechanism. We apply this objective
to the benchmark's \textsc{AE}--\textsc{Gate}--\textsc{Instant} configuration.

\subsection{Directional Supervision}
\label{sec:method-design}

\textbf{Objective.} The forecasting loss scores $f_\theta$ only under the plan
that was carried out. We add a second term that scores its response to a
change of plan. For a declared mechanism $(i,c,s)$ with $s\in\{-1,+1\}$, raising
action channel $i$ should move the forecast of state channel $c$ in direction
$s$; the term compares two forecasts of the same window that differ only in the
future values of channel $i$ and penalizes the part of their difference that
points against $s$. Let $E$ and $D$ be the frozen state encoder and decoder,
$B_\theta$ the backbone and $G_\theta$ the gated fusion module, with the
history-context arguments of $G_\theta$ omitted. The two forecasts are
\begin{equation}
    \hat z=B_\theta(E(h)),\qquad
    \hat y=D\!\left(G_\theta(\hat z,a)\right),\qquad
    \hat y^+=D\!\left(G_\theta(\hat z,\tau_{i,\delta}a)\right).
    \label{eq:paired-plans}
\end{equation}
The backbone runs once, since it never sees the plan; history, exogenous inputs
and the other action channels are identical in both branches. Gradients flow
through $D$ into $G_\theta$ and $B_\theta$, while $E$ and $D$ stay frozen.
Inference uses the single branch under the supplied plan.

\textbf{Perturbation.} The shifted plan is the quantile shift $\tau_{i,\delta}a$
of Section~\ref{sec:setup}, with the quantiles of channel $i$ estimated on the
training actions, so the shifted values stay within the range the channel
actually takes. For each batch we draw one $\delta\sim\mathcal{U}(0.05,0.30)$,
so the penalty covers a band of shift sizes. Training shifts only upward; downward shifts are
never penalized and are reserved for evaluation.

\textbf{Penalty.} For a mechanism $p=(i,c,s)$, let $d^p_{k,c}=[\hat y^+-\hat
y]_{k,c}$ be the decoded response at horizon step $k$, let
$\langle\cdot\rangle_p$ average over the windows of the batch and the horizon
steps, restricted to the steps in the required mode for a conditional
mechanism, and let $\mathcal{P}_b$ be the mechanisms with at least one eligible
step in the batch. We minimize
\begin{equation}
\mathcal{L}_{\mathrm{dir}}
=\frac{1}{|\mathcal{P}_b|}\sum_{p\in\mathcal{P}_b}
\frac{\left\langle\operatorname{relu}(-s\,d^p_{k,c})\right\rangle_p}
     {\operatorname{sg}\!\left(\left\langle|d^p_{k,c}|\right\rangle_p\right)+\varepsilon},
\qquad
\mathcal{L}=\mathcal{L}_{\mathrm{forecast}}+\rho\,\mathcal{L}_{\mathrm{dir}},
\label{eq:penalty}
\end{equation}
where $\mathcal{L}_{\mathrm{forecast}}$ is the unchanged MSE objective of the
benchmark, including any backbone-specific auxiliary loss, $\operatorname{sg}$
stops gradients, $\varepsilon=10^{-6}$, and $\mathcal{L}_{\mathrm{dir}}=0$ when
$\mathcal{P}_b$ is empty. The numerator is the response mass in the wrong
direction and the denominator the total response mass, so
$\mathcal{L}_{\mathrm{dir}}$ is roughly the fraction of the response with the
wrong sign: it is zero when the response has the declared sign everywhere, it is
dimensionless, so that one $\rho$ serves datasets whose effect sizes differ,
and the detached denominator lets the gradient act only on the wrong-signed
part. The penalty is applied after decoding, because the declared sign refers to the
state channels. We use $\rho=0.1$, the smallest value in $\{0.1,1,10\}$ that met
the consistency criterion of a Greenhouse pilot, fixed before the full grid;
early stopping uses the forecasting loss alone. The constraint is soft and
does not guarantee monotonicity.

\subsection{Results}
\label{sec:method-results}

\begin{figure}[t]
    \centering
    \includegraphics[width=\textwidth]{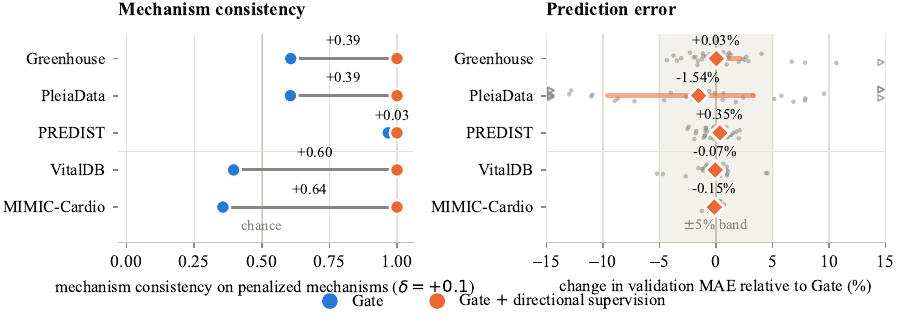}
    \vspace{-1mm}
    \caption{Directional supervision against the \textsc{Gate} control.
    Left: mechanism consistency ($\mathrm{MC}$ at $\delta=+0.1$) on the penalized
    mechanisms, mean over the $35$ backbone--seed runs of each dataset, with the
    paired gain.
    Right: change in validation MAE relative to \textsc{Gate} per run, with the
    median (diamond), a $90\%$ interval (bar) and the $\pm5\%$ band (shaded).}
    \label{fig:soft}
\end{figure}

We train on the six datasets with continuous actions, with the backbones, seeds, split
and protocol of Section~\ref{sec:benchmark-setup}, and pair each run with the
\textsc{Gate} run of Experiment~2 that shares its dataset, backbone and seed. The penalty
covers twelve mechanisms of Table~\ref{tab:declared-mechanisms}, three on Greenhouse,
one on PleiaData, two on PREDIST, two on VitalDB and four on MIMIC-Cardio; the remaining
nine are held out, and Wastewater, which has no penalized mechanism, serves as a null
control. Figure~\ref{fig:soft} reports $\mathrm{MC}$ at $\delta=+0.1$ on held-out
windows and validation MAE, with the paired change per run; Appendix~\ref{app:directional} gives every
number. $\mathrm{MC}$ on the penalized mechanisms rises on every dataset to a mean above
$0.99$ over the $35$ runs on each mechanism, including the windward vent, on which \textsc{Gate} scores $0.196$, and four
of the six clinical mechanisms, on which \textsc{Gate} is below chance. The downward
shift, which the penalty never sees, is as consistent as the upward one, so the learned
response holds on held-out windows and in the untrained direction. Prediction error is
unchanged: the median change in validation MAE, pooled over datasets, is $-0.05\%$, and
the per-dataset medians lie between $-1.54\%$ and $+0.35\%$, inside the $\pm5\%$ band;
MIMIC-Cardio shows a small improvement, and PleiaData has the widest spread across
runs. The gain is
confined to the supplied signs: the held-out drug mechanisms stay below chance except
dopamine, and three held-out mechanisms move away from their sign, the PleiaData
cooling-mode setpoint, vasopressin and remifentanil on mean pressure
(Appendix~\ref{app:directional}). Directional
supervision therefore enforces the signs it is given and only those: it does not recover
undeclared mechanisms and does not constrain effect sizes. Within that scope, it gives a
forecaster that keeps the accuracy of the best adapter and responds to a changed plan in
the declared direction, which no choice in the design space achieved.


\section{Conclusion}
\label{sec:conclusion}

We formalized the time series world model, a forecaster that takes the observed history
and a plan of actions and exogenous inputs and predicts the response of the state, and
asked two questions that current practice leaves open: which design choices matter, and
whether an accurate forecaster responds to a changed plan the way the real system does.
A benchmark of eight public datasets with real actions, swept one design choice at a
time, answers the first: a frozen latent prediction space and gated fusion at the output
lower the error across datasets and backbones, whereas a memory in the plan encoding
brings no consistent gain. Mechanism consistency, a metric that scores the direction of
the response to a shifted action, answers the second in the negative: the most accurate
configurations are at or below chance on most datasets, and no design choice escapes
this, because the sign in the training data is observational and an architecture does
not change the data. Directional supervision supplies the sign through the loss and
raises consistency to near its maximum at no cost in error, so a mechanism-consistent world
model is a property of the training objective rather than of the architecture. Future
work includes extending the supervision to event actions, clinical validation of the
declared pharmacological signs, and evaluation under closed-loop control.

\subsection*{AI use statement}

In this work, we used generative AI tools to check the manuscript for
grammatical and spelling errors, to refine language expression, to prepare the tables and figures from the stored
metric files, and to organize the code. We did not use generative AI tools to write
the manuscript, to write the experiment code, to generate or label data, to declare
the mechanism signs of Table~\ref{tab:declared-mechanisms}, or to produce or select
the experimental results. The authors are responsible for the final manuscript, code and other
research artifacts, including AI-assisted content.

\subsection*{Ethics statement}

This work studies action-conditioned forecasting on existing public datasets and
pretrained models; no new data were collected from human subjects. VitalDB and MIMIC-IV
are de-identified and used under their access agreements; MIMIC-IV requires credentialed access, so we release the
extraction code for MIMIC-Cardio rather than the data. The declared mechanisms are
simplified statements of direction, not clinical guidance, and a model that satisfies
mechanism consistency is not thereby validated for treatment decisions: the metric
checks the sign of a response and says nothing about its size or its safety.
Directional supervision enforces the signs it is given, so a wrongly declared sign
would be enforced as well; the signs should be reviewed by domain experts before the
method is used in a new setting. Any deployment on patient data requires evaluation
in the intended clinical setting and appropriate privacy protections.

\subsection*{Reproducibility statement}

The problem formalization, the three action forms and the mechanism-consistency metric
are defined in Section~\ref{sec:setup}, and the design space in
Section~\ref{sec:design-space}, with the formula of every fusion method and action
encoder in Tables~\ref{tab:fusion-formulas} and~\ref{tab:encoding-formulas}.
Section~\ref{sec:datasets} and Appendix~\ref{app:datasets} give the channel roles,
scaling ranges, split and windowing of each dataset, and
Table~\ref{tab:declared-mechanisms} the declared mechanisms.
Section~\ref{sec:benchmark-setup} states what each experiment fixes and varies;
Table~\ref{tab:hyperparams} lists every hyperparameter of the forecasting protocol, the
codecs, the fusion and encoding modules and the mechanism-consistency probe, and
Appendix~\ref{app:directional} those of the directional-supervision term; none is tuned
per dataset or backbone.
Appendix~\ref{app:protocol} gives the aggregation and the bootstrap intervals behind
the reported effects, Appendix~\ref{app:sweeps} the paired effects and their
interaction with the backbone, Appendix~\ref{app:grids} the complete validation grids of
Experiments~1--3, and Appendix~\ref{app:mechanism-consistency} the numbers behind
Figure~\ref{fig:sec4-mechanism}, the per-mechanism scores and the mechanisms that were
set aside. The objective of Section~\ref{sec:method} is given in full in
Equations~\ref{eq:paired-plans} and~\ref{eq:penalty}, and
Appendix~\ref{app:directional} gives its hyperparameters, the penalized set and the
score of every mechanism and dataset. The preprocessing code, the
MIMIC-Cardio extraction script, the declared mechanisms, the training and evaluation
code, the mechanism-consistency probe, the configuration of every run and the launch
scripts of every experiment are available at
\url{https://github.com/HaochenZhang717/TSWM-Official-Repo}.

% ...

\bibliographystyle{style/icml2025}
\bibliography{999_reference}

@article{agc3,
  title={3rd Autonomous Greenhouse Challenge: Time-series data on realized climate with annotated crop lettuce-images},
  author={Petropoulou, Anna and van Marrewijk, Bart and Hemming, Silke and de Zwart, Feije and Elings, Anne and Bijlaard, Monique},
  year={2023},
  publisher={Wageningen University \& Research}
}

@article{pleiadata,
  title={PLEIAData: consumption, HVAC, temperature, weather and motion sensor data for smart buildings applications},
  author={Ibarra, Antonio Mart{\'\i}nez and Gonz{\'a}lez-Vidal, Aurora and Skarmeta, Antonio},
  journal={Scientific Data},
  volume={10},
  number={1},
  pages={118},
  year={2023},
  publisher={Nature Publishing Group UK London}
}

@article{predist,
  title={Enabling predictive maintenance in district heating substations: A labelled dataset and fault detection evaluation framework based on service data},
  author={Roelofs, Cyriana MA and Bastidas, Edison Guevara and Hugo, Thomas and Faulstich, Stefan and Cadenbach, Anna},
  journal={Energy},
  pages={141178},
  year={2026},
  publisher={Elsevier}
}

@article{agtrup,
  title={Wastewater treatment plant data for nutrient removal system},
  author={Mohammadi, Esmaeel and Rani, Anju and Stokholm-Bjerregaard, Mikkel and Ortiz-Arroyo, Daniel and Durdevic, Petar},
  journal={arXiv preprint arXiv:2407.05346},
  year={2024}
}

@article{vitaldb,
  title={VitalDB, a high-fidelity multi-parameter vital signs database in surgical patients},
  author={Lee, Hyung-Chul and Park, Yoonsang and Yoon, Soo Bin and Yang, Seong Mi and Park, Dongnyeok and Jung, Chul-Woo},
  journal={Scientific Data},
  volume={9},
  number={1},
  pages={279},
  year={2022},
  publisher={Nature Publishing Group UK London}
}

@article{cgmacros,
  title={CGMacros: a pilot scientific dataset for personalized nutrition and diet monitoring},
  author={Das, Anurag and Kerr, David and Glantz, Namino and Bevier, Wendy and Santiago, Rony and Gutierrez-Osuna, Ricardo and Mortazavi, Bobak J},
  journal={Scientific Data},
  volume={12},
  number={1},
  pages={1557},
  year={2025},
  publisher={Nature Publishing Group UK London}
}

@article{shanghai,
  title={Chinese diabetes datasets for data-driven machine learning},
  author={Zhao, Qinpei and Zhu, Jinhao and Shen, Xuan and Lin, Chuwen and Zhang, Yinjia and Liang, Yuxiang and Cao, Baige and Li, Jiangfeng and Liu, Xiang and Rao, Weixiong and others},
  journal={Scientific Data},
  volume={10},
  number={1},
  pages={35},
  year={2023},
  publisher={Nature Publishing Group UK London}
}

@article{mimiciv,
  title={MIMIC-IV, a freely accessible electronic health record dataset},
  author={Johnson, Alistair EW and Bulgarelli, Lucas and Shen, Lu and Gayles, Alvin and Shammout, Ayad and Horng, Steven and Pollard, Tom J and Hao, Sicheng and Moody, Benjamin and Gow, Brian and others},
  journal={Scientific data},
  volume={10},
  number={1},
  pages={1},
  year={2023},
  publisher={Nature Publishing Group UK London}
}

@inproceedings{qiu2025duet,
  title={Duet: Dual clustering enhanced multivariate time series forecasting},
  author={Qiu, Xiangfei and Wu, Xingjian and Lin, Yan and Guo, Chenjuan and Hu, Jilin and Yang, Bin},
  booktitle={Proceedings of the 31st ACM SIGKDD conference on knowledge discovery and data mining v. 1},
  pages={1185--1196},
  year={2025}
}

@inproceedings{huang2025timekan,
  title={TimeKAN: KAN-based frequency decomposition learning architecture for long-term time series forecasting},
  author={Huang, Songtao and Zhao, Zhen and Li, Can and Bai, Lei},
  booktitle={International Conference on Learning Representations},
  volume={2025},
  pages={28514--28529},
  year={2025}
}

@article{nie2023patchtst,
  title={A time series is worth 64 words: Long-term forecasting with transformers},
  author={Nie, Yuqi and Nguyen, Nam H and Sinthong, Phanwadee and Kalagnanam, Jayant},
  journal={arXiv preprint arXiv:2211.14730},
  year={2022}
}

@inproceedings{fei2025amplifier,
  title={Amplifier: Bringing attention to neglected low-energy components in time series forecasting},
  author={Fei, Jingru and Yi, Kun and Fan, Wei and Zhang, Qi and Niu, Zhendong},
  booktitle={Proceedings of the AAAI conference on artificial intelligence},
  volume={39},
  number={11},
  pages={11645--11653},
  year={2025}
}

@inproceedings{mu2026epibench,
  title={Benchmarking Counterfactual Prediction in Epidemic Time Series with Time-Varying Interventions},
  author={Mu, Wenhao and Yan, Facundo and Mumssen, Anik and Eisenberg, Marisa and Rodr{\'\i}guez, Alexander},
  booktitle={Proceedings of the 32nd ACM SIGKDD Conference on Knowledge Discovery and Data Mining V. 2},
  pages={9593--9603},
  year={2026}
}

@article{thumm2026dotime,
  title={DoTime: A Synthetic Benchmark Generator for Interventional and Counterfactual Time Series},
  author={Thumm, Dennis and Anthony, Billy Tim and Chen, Ying},
  journal={arXiv preprint arXiv:2607.27263},
  year={2026}
}

@article{godahewa2021monash,
  title={Monash time series forecasting archive},
  author={Godahewa, Rakshitha and Bergmeir, Christoph and Webb, Geoffrey I and Hyndman, Rob J and Montero-Manso, Pablo},
  journal={arXiv preprint arXiv:2105.06643},
  year={2021}
}

@article{dinatale2022pcnn,
  title={Physically consistent neural networks for building thermal modeling: theory and analysis},
  author={Di Natale, Loris and Svetozarevic, Bratislav and Heer, Philipp and Jones, Colin N},
  journal={Applied Energy},
  volume={325},
  pages={119806},
  year={2022},
  publisher={Elsevier}
}

@incollection{kunz2023zalando,
  title={Deep Learning based Forecasting: a case study from the online fashion industry},
  author={Kunz, Manuel and Birr, Stefan and Raslan, Mones and Ma, Lei and Januschowski, Tim},
  booktitle={Forecasting with artificial intelligence: theory and applications},
  pages={279--311},
  year={2023},
  publisher={Springer}
}

@article{keogh2024interventions,
  title={Prediction under interventions: evaluation of counterfactual performance using longitudinal observational data},
  author={Keogh, Ruth H and Van Geloven, Nan},
  journal={Epidemiology (Cambridge, Mass.)},
  volume={35},
  number={3},
  pages={329-339},
  year={2024}
}

@inproceedings{bruce2024genie,
  title={Genie: Generative interactive environments},
  author={Bruce, Jake and Dennis, Michael D and Edwards, Ashley and Parker-Holder, Jack and Shi, Yuge and Hughes, Edward and Lai, Matthew and Mavalankar, Aditi and Steigerwald, Richie and Apps, Chris and others},
  booktitle={Forty-first international conference on machine learning},
  year={2024}
}

@inproceedings{perez2018film,
  title={Film: Visual reasoning with a general conditioning layer},
  author={Perez, Ethan and Strub, Florian and De Vries, Harm and Dumoulin, Vincent and Courville, Aaron},
  booktitle={Proceedings of the AAAI conference on artificial intelligence},
  volume={32},
  number={1},
  year={2018}
}

@inproceedings{dauphin2017glu,
  title={Language modeling with gated convolutional networks},
  author={Dauphin, Yann N and Fan, Angela and Auli, Michael and Grangier, David},
  booktitle={International conference on machine learning},
  pages={933--941},
  year={2017},
  organization={PMLR}
}

@inproceedings{assran2023ijepa,
  title={Self-supervised learning from images with a joint-embedding predictive architecture},
  author={Assran, Mahmoud and Duval, Quentin and Misra, Ishan and Bojanowski, Piotr and Vincent, Pascal and Rabbat, Michael and LeCun, Yann and Ballas, Nicolas},
  booktitle={2023 IEEE/CVF Conference on Computer Vision and Pattern Recognition (CVPR)},
  pages={15619--15629},
  year={2023},
  organization={IEEE}
}

@article{qi2026mobiwm,
  title={Beyond Static Forecasting: Unleashing the Power of World Models for Mobile Traffic Extrapolation},
  author={Qi, Xiaoqian and Chai, Haoye and Wang, Yue and Li, Yong},
  journal={arXiv preprint arXiv:2604.08199},
  year={2026}
}

@article{ansari2025chronos2,
  title={Chronos-2: From univariate to universal forecasting},
  author={Ansari, Abdul Fatir and Shchur, Oleksandr and K{\"u}ken, Jaris and Auer, Andreas and Han, Boran and Mercado, Pedro and Rangapuram, Syama Sundar and Shen, Huibin and Stella, Lorenzo and Zhang, Xiyuan and others},
  journal={arXiv preprint arXiv:2510.15821},
  year={2025}
}

@article{lim2021tft,
  title={Temporal fusion transformers for interpretable multi-horizon time series forecasting},
  author={Lim, Bryan and Ar{\i}k, Sercan {\"O} and Loeff, Nicolas and Pfister, Tomas},
  journal={International journal of forecasting},
  volume={37},
  number={4},
  pages={1748--1764},
  year={2021},
  publisher={Elsevier}
}

@article{das2023tide,
  title={Long-term forecasting with tide: Time-series dense encoder},
  author={Das, Abhimanyu and Kong, Weihao and Leach, Andrew and Mathur, Shaan and Sen, Rajat and Yu, Rose},
  journal={arXiv preprint arXiv:2304.08424},
  year={2023}
}

@article{olivares2023nbeatsx,
  title={Neural basis expansion analysis with exogenous variables: Forecasting electricity prices with NBEATSx},
  author={Olivares, Kin G and Challu, Cristian and Marcjasz, Grzegorz and Weron, Rafa{\l} and Dubrawski, Artur},
  journal={International Journal of Forecasting},
  volume={39},
  number={2},
  pages={884--900},
  year={2023},
  publisher={Elsevier}
}

@article{wang2024timexer,
  title={Timexer: Empowering transformers for time series forecasting with exogenous variables},
  author={Wang, Yuxuan and Wu, Haixu and Dong, Jiaxiang and Qin, Guo and Zhang, Haoran and Liu, Yong and Qiu, Yunzhong and Wang, Jianmin and Long, Mingsheng},
  journal={Advances in neural information processing systems},
  volume={37},
  pages={469--498},
  year={2024}
}

@article{arango2025chronosx,
  title={Chronosx: Adapting pretrained time series models with exogenous variables},
  author={Arango, Sebastian Pineda and Mercado, Pedro and Kapoor, Shubham and Ansari, Abdul Fatir and Stella, Lorenzo and Shen, Huibin and Senetaire, Hugo and Turkmen, Caner and Shchur, Oleksandr and Maddix, Danielle C and others},
  journal={arXiv preprint arXiv:2503.12107},
  year={2025}
}

@article{ha2018world,
  title={Recurrent world models facilitate policy evolution},
  author={Ha, David and Schmidhuber, J{\"u}rgen},
  journal={Advances in neural information processing systems},
  volume={31},
  year={2018}
}

@article{yang2024unisim,
  title={Learning interactive real-world simulators},
  author={Yang, Sherry and Du, Yilun and Ghasemipour, Kamyar and Tompson, Jonathan and Kaelbling, Leslie and Schuurmans, Dale and Abbeel, Pieter},
  journal={arXiv preprint arXiv:2310.06114},
  year={2023}
}

@article{ding2025clarity,
  title={CLARITY: Medical world model for guiding treatment decisions by modeling context-aware disease trajectories in latent space},
  author={Ding, Tianxingjian and Zou, Yuanhao and Chen, Chen and Shah, Mubarak and Tian, Yu},
  journal={arXiv preprint arXiv:2512.08029},
  year={2025}
}

@article{assran2025vjepa2,
  title={V-jepa 2: Self-supervised video models enable understanding, prediction and planning},
  author={Assran, Mido and Bardes, Adrien and Fan, David and Garrido, Quentin and Howes, Russell and Komeili, Mojtaba and Muckley, Matthew and Rizvi, Ammar and Roberts, Claire and Sinha, Koustuv and Zholus, Artem and others},
  journal={arXiv preprint arXiv:2506.09985},
  year={2025}
}

@article{karniadakis2021piml,
  title={Physics-informed machine learning},
  author={Karniadakis, George Em and Kevrekidis, Ioannis G and Lu, Lu and Perdikaris, Paris and Wang, Sifan and Yang, Liu},
  journal={Nature Reviews Physics},
  volume={3},
  number={6},
  pages={422--440},
  year={2021},
  publisher={Nature Publishing Group UK London}
}

@article{raissi2019pinn,
  title={Physics-informed neural networks: A deep learning framework for solving forward and inverse problems involving nonlinear partial differential equations},
  author={Raissi, Maziar and Perdikaris, Paris and Karniadakis, George E},
  journal={Journal of Computational physics},
  volume={378},
  pages={686--707},
  year={2019},
  publisher={Elsevier}
}

@article{zhou2025dinowm,
  title={Dino-wm: World models on pre-trained visual features enable zero-shot planning},
  author={Zhou, Gaoyue and Pan, Hengkai and LeCun, Yann and Pinto, Lerrel},
  journal={arXiv preprint arXiv:2411.04983},
  year={2024}
}

@article{wang2026medicalwm,
  title={Medical world models: representing medical states, modelling clinical dynamics and guiding intervention policies},
  author={Liu, Ke and Li, Mengxuan and Bao, Yanyi and Zhang, Tianyun and Chu, Chong and Bu, Jiajun and Wang, Haishuai},
  journal={arXiv preprint arXiv:2606.16721},
  year={2026}
}

@article{nvidia2025cosmos,
  title={Cosmos world foundation model platform for physical ai},
  author={Agarwal, Niket and Ali, Arslan and Bala, Maciej and Balaji, Yogesh and Barker, Erik and Cai, Tiffany and Chattopadhyay, Prithvijit and Chen, Yongxin and Cui, Yin and Ding, Yifan and others},
  journal={arXiv preprint arXiv:2501.03575},
  year={2025}
}

@article{hafner2025dreamerv3,
  title={Mastering diverse control tasks through world models},
  author={Hafner, Danijar and Pasukonis, Jurgis and Ba, Jimmy and Lillicrap, Timothy},
  journal={Nature},
  volume={640},
  number={8059},
  pages={647--653},
  year={2025},
  publisher={Nature Publishing Group UK London}
}

@inproceedings{hafner2019planet,
  title={Learning latent dynamics for planning from pixels},
  author={Hafner, Danijar and Lillicrap, Timothy and Fischer, Ian and Villegas, Ruben and Ha, David and Lee, Honglak and Davidson, James},
  booktitle={International conference on machine learning},
  pages={2555--2565},
  year={2019},
  organization={PMLR}
}

@article{qiu2026dag,
  title={Dag: A dual correlation network for time series forecasting with exogenous variables},
  author={Qiu, Xiangfei and Zhu, Yuhan and Li, Zhengyu and Wu, Xingjian and Yang, Bin and Hu, Jilin},
  journal={arXiv preprint arXiv:2509.14933},
  year={2025}
}

@inproceedings{han2026unica,
  title={UniCA: Unified covariate adaptation for time series foundation model},
  author={Han, Lu and Liu, Yu and Li, Lan and Deng, Qiwen and Jiang, Jian and Sun, Yinbo and Yu, Zhe and Wang, Binfeng and Lu, Xingyu and Ma, Lintao and Ye, Han-Jia and others},
  booktitle={International Conference on Learning Representations},
  volume={2026},
  pages={6403--6443},
  year={2026}
}

@article{hoo2025tabpfnts,
  title={From tables to time: Extending tabpfn-v2 to time series forecasting},
  author={Hoo, Shi Bin and M{\"u}ller, Samuel and Salinas, David and Hutter, Frank},
  journal={arXiv preprint arXiv:2501.02945},
  year={2025}
}

@techreport{jin2017mmm,
    title	= {Bayesian Methods for Media Mix Modeling with Carryover and Shape Effects},
    author	= {Yuxue Jin and Yueqing Wang and Yunting Sun and David Chan and Jim Koehler},
    year	= {2017},
    institution	= {Google Inc.}
}

@article{crasson2024causal,
  title={Training and Evaluating Causal Forecasting Models for Time-Series},
  author={Crasson, Thomas and Nabet, Yacine and L{\'e}cuyer, Mathias},
  journal={arXiv preprint arXiv:2411.00126},
  year={2024}
}

@inproceedings{zhou2025crosslinear,
  title={Crosslinear: Plug-and-play cross-correlation embedding for time series forecasting with exogenous variables},
  author={Zhou, Pengfei and Liu, Yunlong and Liang, Junli and Song, Qi and Li, Xiangyang},
  booktitle={Proceedings of the 31st ACM SIGKDD Conference on Knowledge Discovery and Data Mining V. 2},
  pages={4120--4131},
  year={2025}
}

@article{zou2024hybrid,
  title={Hybrid$^2$ Neural ODE Causal Modeling and an Application to Glycemic Response},
  author={Zou, Bob Junyi and Levine, Matthew E and Zaharieva, Dessi P and Johari, Ramesh and Fox, Emily B},
  journal={arXiv preprint arXiv:2402.17233},
  year={2024}
}

@inproceedings{runje2023monotonic,
  title={Constrained monotonic neural networks},
  author={Runje, Davor and Shankaranarayana, Sharath M},
  booktitle={International Conference on Machine Learning},
  pages={29338--29353},
  year={2023},
  organization={PMLR}
}

@inproceedings{kitouni2023monotonic,
  title={Expressive monotonic neural networks},
  author={Nolte, Niklas and Kitouni, Ouail and Williams, Mike},
  booktitle={The Eleventh International Conference on Learning Representations},
  year={2023}
}

@article{yang2026latenttsf,
  title={From observations to states: Latent time series forecasting},
  author={Yang, Jie and Hu, Yifan and Li, Yuante and Zhang, Kexin and Ding, Kaize and Yu, Philip S},
  journal={arXiv preprint arXiv:2602.00297},
  year={2026}
}

@article{qin2025cora,
  title={Cora: Covariate-aware adaptation of time series foundation models},
  author={Qin, Guo and Chen, Zhi and Liu, Yong and Shi, Zhiyuan and Liu, Haixuan and Huang, Xiangdong and Wang, Jianmin and Long, Mingsheng},
  journal={arXiv preprint arXiv:2510.12681},
  year={2025}
}

@article{yamaguchi2026citras,
  title={Citras: Covariate-informed transformer for time series forecasting},
  author={Yamaguchi, Yosuke and Suemitsu, Issei and Wei, Wenpeng},
  journal={IEEE Access},
  year={2026},
  publisher={IEEE}
}

@inproceedings{melnychuk2022causal,
  title={Causal transformer for estimating counterfactual outcomes},
  author={Melnychuk, Valentyn and Frauen, Dennis and Feuerriegel, Stefan},
  booktitle={International conference on machine learning},
  pages={15293--15329},
  year={2022},
  organization={PMLR}
}

@article{bica2020crn,
  title={Estimating counterfactual treatment outcomes over time through adversarially balanced representations},
  author={Bica, Ioana and Alaa, Ahmed M and Jordon, James and Van Der Schaar, Mihaela},
  journal={arXiv preprint arXiv:2002.04083},
  year={2020}
}

@article{aksu2024gifteval,
  title={Gift-eval: A benchmark for general time series forecasting model evaluation},
  author={Aksu, Taha and Woo, Gerald and Liu, Juncheng and Liu, Xu and Liu, Chenghao and Savarese, Silvio and Xiong, Caiming and Sahoo, Doyen},
  journal={arXiv preprint arXiv:2410.10393},
  year={2024}
}

@article{che2018grud,
  title={Recurrent neural networks for multivariate time series with missing values},
  author={Che, Zhengping and Purushotham, Sanjay and Cho, Kyunghyun and Sontag, David and Liu, Yan},
  journal={Scientific reports},
  volume={8},
  number={1},
  pages={6085},
  year={2018},
  publisher={Nature Publishing Group UK London}
}

\appendix

\section{Datasets and Preprocessing}
\label{app:datasets}

This appendix supports Section~\ref{sec:datasets}. It describes each dataset of
Table~\ref{tab:datasets} under the three roles of Section~\ref{sec:setup}, state, action
and exogenous input, and then the two preprocessing steps shared by all methods, the
scaling of every channel and the split into training and held-out windows.

\subsection{The eight datasets}

\textbf{Greenhouse}~\citep{agc3} contains the climate records of six greenhouse
compartments from the third Autonomous Greenhouse Challenge, in which six teams each
grew a lettuce crop under their own control strategy over about six months, at 5-minute
resolution. The state consists of the indoor temperature, relative humidity, CO$_2$
concentration and photosynthetically active radiation. The seven continuous actions are
actuator intensities: heating-pipe temperature, leeward and windward vent apertures,
artificial lighting, energy screen, blackout screen and CO$_2$ dosing. The exogenous
inputs are eight outdoor weather variables, including temperature, humidity, irradiance,
wind and rain.
\textbf{PleiaData}~\citep{pleiadata} contains one year of operation data from a
university building at 10-minute resolution; a system pairs an indoor temperature sensor
with the HVAC unit serving its room. The state consists of the indoor temperature and the
energy consumed per step. The continuous action is the temperature setpoint entered by the
occupants, and two mode channels record the on/off status and the operating mode of the
unit (off, heat, cool, dry, fan, auto). The exogenous inputs are eight outdoor weather
variables.
\textbf{PREDIST}~\citep{predist} contains operational records of $58$ district-heating
substations at 10-minute resolution. The state consists of the secondary supply
temperature, the primary and network return temperatures, heat power and flow. The
continuous actions are the control-valve position and the supply-temperature setpoint
issued by the control unit; the mode channels are the pump status, the control mode and
the position of the domestic hot-water three-way valve. The exogenous inputs are the
outdoor temperature and the network supply temperature. We exclude maintenance events,
which are manual repairs rather than control signals.
\textbf{Wastewater}~\citep{agtrup} contains sensor and dosing records from a biological
reactor of the Agtrup wastewater treatment plant in Kolding, Denmark, at 2-minute
resolution. The state consists of the dissolved oxygen, ammonium and phosphate
concentrations in the tank. The continuous actions are two metal-salt dosing rates for
phosphate precipitation and a chemical dosing factor; the mode channels are the process
phases at the inlet and the outlet. The exogenous inputs are the influent flow and the
wastewater temperature. Since the dataset covers a single plant, we split it
in time rather than by system.
\textbf{VitalDB}~\citep{vitaldb} contains intraoperative vital signs of $3{,}443$
surgical cases at 2-second resolution. The state consists of mean arterial pressure, heart
rate, peripheral oxygen saturation and the bispectral index. The continuous actions are
the infusion rates of propofol and remifentanil, the hypnotic and the opioid used to
maintain anesthesia. There are no exogenous inputs.
\textbf{CGMacros}~\citep{cgmacros} contains continuous glucose monitoring traces of $45$
participants at 1-minute resolution together with their meal logs. The state is the
glucose concentration. The actions are events of four types (breakfast, lunch, dinner and
snack).
\textbf{Shanghai}~\citep{shanghai} contains continuous glucose monitoring traces of $112$
patients with type~1 or type~2 diabetes at 15-minute resolution together with their meal,
injection and oral-medication records. The state is the glucose concentration. The actions
are events of $31$ types: meals, $16$ injected agents (insulin preparations and GLP-1
agonists) and $14$ oral hypoglycemic drugs, several of which may be administered at the
same step.
\textbf{MIMIC-Cardio} is our extraction from MIMIC-IV~\citep{mimiciv} of the intensive-care
stays of $8{,}478$ patients, resampled to 30-minute resolution. The state consists of the
heart rate, respiratory rate, oxygen saturation, and systolic, diastolic and mean
non-invasive blood pressure. The $16$ continuous actions are the rates of nine continuous
infusions (norepinephrine, epinephrine, phenylephrine, vasopressin, dopamine, dobutamine,
milrinone, furosemide and intravenous fluid), the amounts of two boluses (furosemide and
fluid), and five ventilator settings (FiO$_2$, PEEP, oxygen flow, set tidal volume and set
respiratory rate). The event channel records the two bolus types as pushes. We exclude the
invasive arterial-pressure channels, which are missing for $52.6\%$ of stays, and $89$
epinephrine pushes whose recorded amounts are unit-entry errors.

\subsection{Scaling ranges}
The scaling range of a channel is fixed once per dataset before training. Greenhouse,
PleiaData, VitalDB, CGMacros, Shanghai and the ventilator channels of MIMIC-Cardio use
constants declared from instrument or physiological limits, e.g.\ $40$ to $400$\,mg/dL
for glucose. Wastewater uses the 1st and 99th percentiles of the training segment, PREDIST
uses the 1st and 99th percentiles of each substation's own series, and the drug channels
of MIMIC-Cardio use the 99.5th percentile of the non-zero doses over the cohort. The
continuous actions of Greenhouse, VitalDB and MIMIC-Cardio have declared ranges whose lower
end is an off position and map to $[0,1]$. All other channels map to $[-1,1]$, including
the PREDIST valve position and the Wastewater dosing rates, whose ranges are percentiles
of the data, so their zero, although it is an off position, is not mapped to a fixed
point.

\subsection{Splits and windows}
Seven datasets contain several systems of one kind, and we split them by system: all
windows of one system go either to the training set or to the held-out set, so that a
model is tested on systems it has never seen and the $N$ training windows, which overlap
in time within a system, are not independent samples. The system is the team for
Greenhouse, the sensor--HVAC pair for PleiaData, the substation for PREDIST, the surgical
case for VitalDB, the subject for CGMacros and Shanghai, and the patient for MIMIC-Cardio;
the train/held-out counts are in Table~\ref{tab:datasets}. Wastewater covers a single
plant and is split in time, the first $80\%$ for training and the rest held out. Within a
system, the series is cut wherever two consecutive samples are further apart than a
tolerance, and every window lies entirely inside one of the resulting segments. The
tolerance is $30$ minutes for Greenhouse, PleiaData, PREDIST, CGMacros and Shanghai,
$10$ minutes for Wastewater and $2$ minutes for VitalDB; a MIMIC-Cardio stay is a single
contiguous segment on its 30-minute grid, so it is never cut.

\begingroup
\raggedbottom
\section{Model Design Space: Formulas and Implementation}
\label{app:design-space}

This appendix supports Section~\ref{sec:design-space}. It gives the formula of every
option of the three design choices in the notation of Section~\ref{sec:setup} and the
implementation details that the formulas leave open.

\subsection{Formulas}

Tables~\ref{tab:fusion-formulas} and \ref{tab:encoding-formulas} give the
formula of every fusion method and every plan encoder in the notation of
Section~\ref{sec:setup}, for the window $W_t$ defined there: the
history covers steps $t-L+1$ to $t$ and the horizon covers steps $t+k$ for
$k\in[H]$. We write $\va_s\in\R^{C}$ for step $s$ of the plan $a$ as the model
receives it, i.e., the exogenous input $\ve_s$ and the continuous actions
$\vu^{\mathrm{c}}_s$ stacked with the embeddings of the mode indices
$\vu^{\mathrm{m}}_s$ and of the event value $u^{\mathrm{e}}_s$. The action
encoder maps $\va_{t-L+1:t+H}$ to $\tilde\va_{t-L+1:t+H}$ of the same shape,
causally, so that $\tilde\va_s$ depends on $\va_{t-L+1:s}$ only. The state
encoder maps every state $\vx_s$ to a latent $\vz_s\in\R^{32}$; in observation
space $\vz_s=\vx_s$. The backbone maps the latent history $\vz_{t-L+1:t}$ to
the plan-blind forecast $\hat\vz_{t+1:t+H}$, the fusion method maps
$\hat\vz_{t+k}$ to the adjusted forecast $\vz'_{t+k}$, and the decoder maps
$\vz'_{t+k}$ to the predicted state $\hat\vx_{t+k}$. The stepwise fusion
methods share a history summary
$\vr=\mathrm{ReLU}\big(W_r[\vz_t;\bar\vz;\tilde\va_t;\bar{\tilde\va}]\big)$,
where a bar denotes the mean over the $L$ history steps, and a per-step feature
$\bm{\phi}_k=[\vr;\tilde\va_{t+k};\vp_k]$ with a learned position embedding
$\vp_k$; $g$ is a two-layer MLP with a ReLU hidden layer of width $128$, and
$\odot$ is the elementwise product. A zero-initialized layer makes its method
the identity at the start of training, i.e., $\vz'_{t+k}=\hat\vz_{t+k}$ or
$\tilde\va_s=\va_s$.

\begin{table}[t]
    \caption{Plan-fusion methods. $\hat\vz_{t+k}$ is the plan-blind forecast
    of the backbone at horizon step $t+k$ and $\vz'_{t+k}$ the adjusted
    forecast that is decoded; $\tilde\va$ is the encoded plan. \textsc{Concat}
    is the only method whose plan enters the backbone itself.}
    \label{tab:fusion-formulas}
    \centering
    \small
    \setlength{\tabcolsep}{4pt}
    \begin{tabular}{@{}llp{0.48\textwidth}@{}}
    \toprule
    Method & Adjusted forecast $\vz'_{t+k}$ & Details \\
    \midrule
    \textsc{None}   & $\hat\vz_{t+k}$ & Plan discarded. \\
    \textsc{Concat} & $\hat\vz_{t+k}$, backbone input $[\vz_s;\tilde\va_s]$ &
        The history plan enters with the latent history; the future plan
        $\tilde\va_{t+k}$ enters through the native covariate interface of
        TimeXer and TiDE, and through $\mathrm{MLP}([\hat\vz_{t+k};\tilde\va_{t+k}])$
        otherwise. \\
    \textsc{Res}    & $\hat\vz_{t+k} + g(\bm{\phi}_k)$ & Standard initialization. \\
    \textsc{Res-0}  & $\hat\vz_{t+k} + g(\bm{\phi}_k)$ & Output layer of $g$ zero-initialized. \\
    \textsc{FiLM-0} & $(1+\bm{\gamma}_k)\odot\hat\vz_{t+k}+\bm{\beta}_k$ &
        $[\bm{\gamma}_k;\bm{\beta}_k]=g(\bm{\phi}_k)$, output layer zero-initialized. \\
    \textsc{Gate}   & $\hat\vz_{t+k} + \bm{\alpha}_k\odot\sigma(\vb_k)$ &
        $[\bm{\alpha}_k;\vb_k]=g([\bm{\phi}_k;\hat\vz_{t+k}])$, $\sigma$ the
        logistic function. \\
    \textsc{X-attn} & $\hat\vz_{t+k} + W_o\,\mathrm{MHA}(\vq_k,\,K,\,K)$ &
        Query $\vq_k=W_q\hat\vz_{t+k}+\vp_k$; keys and values
        $K=W_{kv}\tilde\va_{t-L+1:t+H}+P'$ over the whole plan, with position
        embeddings $P'$; four heads, width $128$. \\
    \bottomrule
    \end{tabular}
\end{table}

\begin{table}[t]
    \caption{Plan encoders. Each maps the plan $\va_{t-L+1:t+H}$ to
    $\tilde\va_{t-L+1:t+H}$ of the same width, causally and per channel unless
    stated otherwise; all parameters are per-channel vectors in $\R^{C}$
    except for \textsc{Attn}.}
    \label{tab:encoding-formulas}
    \centering
    \small
    \setlength{\tabcolsep}{4pt}
    \begin{tabular}{@{}llp{0.50\textwidth}@{}}
    \toprule
    Encoder & Encoded plan $\tilde\va_s$ & Details \\
    \midrule
    \textsc{Instant} & $\va_s$ & Identity. \\
    \textsc{Decay}   & $\va_s + \bm{\eta}\odot(\vm_s-\va_s)$ &
        $\vm_s=(1-\bm{\alpha})\odot\va_s+\bm{\alpha}\odot\vm_{s-1}$ with
        $\bm{\alpha}=\sigma(\bm{\theta})$ learned per channel; the gate
        $\bm{\eta}$ is zero-initialized. \\
    \textsc{Conv}    & $\vv^{(6)}_s$, with $\vv^{(0)}=\va$ &
        Six residual blocks ${\vv^{(l)}_s=\vv^{(l-1)}_s+\mathrm{GELU}(\vc^{(l)}_s)}$
        with ${\vc^{(l)}_s=\sum_{j=0}^{7} \vw^{(l)}_j\odot \vv^{(l-1)}_{s-2^{l-1}j}}$:
        depthwise kernels of eight taps with dilation $2^{l-1}$,
        zero-initialized; receptive field $442$ steps. \\
    \textsc{Attn}    & $\va_s + W_o\,\mathrm{MHA}(\vo_s,\,\vo_{\le s},\,\vo_{\le s})$ &
        $\vo_s=W_i\va_s+\vp'_s$ with position embeddings $\vp'_s$; causal
        mask; four heads, width $64$; $W_o$ zero-initialized. \\
    \bottomrule
    \end{tabular}
\end{table}

\subsection{Implementation}

The \textsc{AE} pretraining objective is reconstruction MSE. The
\textsc{VAE} adds a KL penalty with weight $10^{-3}$ and uses the posterior
mean when forecasting. The pretrained codec is frozen and shared across the
five forecasting seeds. Our \textsc{JEPA} variant is trained jointly from
scratch with observation-space forecasting MSE plus a latent MSE term of
weight $0.02$. The latter compares the predicted latent, after plan fusion,
with an exponential-moving-average encoder of the true future states; its
momentum rises from $0.996$ to $1$ on a cosine schedule. Built-in backbone
regularizers are retained, including DUET's expert load-balancing term.
All runs share the protocol of Table~\ref{tab:hyperparams}; the held-out set
of the fixed split selects the checkpoint and provides the reported MAE. The
plan carries actions and exogenous inputs together, and the three experiments
do not separate their contributions.

\textsc{Concat} concatenates the plan history with the state
representation. The future plan enters TimeXer and TiDE through their native
covariate interfaces, and the other five backbones through a shared output MLP.
The other fusion methods use a backbone without action or exogenous inputs.
\textsc{Res}, \textsc{Res-0}, \textsc{FiLM-0}, and \textsc{Gate} also
receive last-value and mean summaries of state and plan history.
The additional \textsc{Decay} encoder computes a learned per-channel
exponential moving average and mixes it with the current input through a
learned gate, initialized at zero. \textsc{Conv} uses six residual, dilated depthwise convolution layers
with GELU activations. Both it and \textsc{Attn} are causal over the
concatenated history and future plan.

\section{Training and Evaluation Protocol}
\label{app:protocol}

This appendix supports Section~\ref{sec:benchmark-setup}. In all three main
experiments, the data split and forecast
windows are fixed across methods. The \textsc{AE} entries in Experiment~1
are reused as \textsc{Concat} in Experiment~2. Experiment~3 reuses the
\textsc{Gate} and \textsc{FiLM-0} entries as its \textsc{Instant} baselines,
and the diagnostic of Appendix~\ref{app:auxiliary} reuses \textsc{Gate} as its
$\lambda=0$ baseline. Their
stored prediction metrics match exactly. These $1{,}120$ shared entries
explain the difference between $6{,}160$ grid entries and $5{,}040$ distinct
forecasting runs; the counts exclude codec pretraining and rollout evaluation.

\subsection{Hyperparameters}
\label{app:hyperparams}

Table~\ref{tab:hyperparams} lists every hyperparameter of the benchmark, and
the mechanism-consistency probe of Sections~\ref{sec:mechanism-consistency} and~\ref{sec:method}.
None was tuned per dataset or per backbone: the forecasting protocol is shared
by all $5{,}040$ runs, the codecs are pretrained once per dataset and frozen,
and the backbones use the default configuration of their reference
implementation in the Time series Forecasting Benchmark (TFB) codebase with instance
normalization enabled.

\begin{table}[H]
    \caption{Hyperparameters. Shared by every dataset, backbone,
    design choice and seed unless a row says otherwise. $L$, $H$ and $d_c$ are
    as in Section~\ref{sec:setup}; $d$ is the latent width; $\mathrm{MAD}$ is
    the median absolute deviation.}
    \label{tab:hyperparams}
    \centering
    \small
    \setlength{\tabcolsep}{4pt}
    \renewcommand{\arraystretch}{1.0}
    \begin{tabular}{@{}lp{0.62\textwidth}@{}}
    \toprule
    \multicolumn{2}{@{}p{0.88\textwidth}@{}}{\emph{Forecasting protocol (Experiments 1--3, the diagnostic of Appendix~\ref{app:auxiliary} and Section~\ref{sec:method}, whose added term is in Appendix~\ref{app:directional})}} \\
    Window & context $L=256$, horizon $H=16$, stride $80$ \\
    Optimizer & Adam, learning rate $10^{-3}$, no weight decay, batch size $128$ \\
    Schedule & halve the learning rate after $5$ epochs without held-out improvement \\
    Stopping & at most $100$ epochs; stop after $20$ epochs without held-out improvement; the checkpoint with the lowest held-out loss is kept \\
    Loss & MSE between the decoded forecast and the observed future states; gradients clipped at norm $1$ \\
    Seeds & training seeds $0$--$4$ (initialization, minibatch order, dropout); the split is fixed at seed $0$ with a held-out fraction of $0.2$ \\
    Backbones & TFB reference implementations with their default configuration; instance normalization on; built-in regularizers kept (e.g.\ DUET's load balancing) \\
    \midrule
    \multicolumn{2}{@{}l}{\emph{Prediction space}} \\
    \textsc{AE}, \textsc{VAE} & step-wise MLP codec, $d_x\to64\to d$ encoder and $d\to64\to d_x$ decoder with ReLU, $d=32$; pretrained on the state channels with reconstruction MSE (plus $10^{-3}\times$KL for the \textsc{VAE}, posterior mean used at forecast time); Adam $10^{-3}$, batch $128$, at most $100$ epochs, patience $15$, same windows and split as above; frozen and shared across seeds \\
    \textsc{JEPA} & same codec shape, $d=32$, trained jointly from scratch; latent prediction loss weight $0.02$; target encoder is an EMA of the online encoder with momentum $0.996$ ramped to $1$ on a cosine schedule; the latent term is excluded from early stopping \\
    \midrule
    \multicolumn{2}{@{}l}{\emph{Plan fusion}} \\
    \textsc{Res}, \textsc{Res-0}, \textsc{FiLM-0}, \textsc{Gate} & adapter MLP $g$ with one ReLU hidden layer of width $128$; position embedding of width $32$; history summary from the last value and the mean of the state and plan histories; output layer zero-initialized for \textsc{Res-0} and \textsc{FiLM-0} \\
    \textsc{X-attn} & four heads, width $128$, keys and values over the whole plan with position embeddings \\
    \textsc{Concat} & native covariate interface of TimeXer and TiDE; for the other backbones an output MLP with one hidden layer of width $256$ \\
    \midrule
    \multicolumn{2}{@{}l}{\emph{Plan encoding}} \\
    \textsc{Decay} & per-channel geometric kernel with $\bm{\alpha}=\sigma(\bm{\theta})$, $\bm{\theta}$ and the mixing gate initialized at $0$ \\
    \textsc{Conv} & six residual depthwise causal blocks, eight taps, dilation $2^{l-1}$, GELU, zero-initialized; receptive field $442$ steps \\
    \textsc{Attn} & causal self-attention, four heads, width $64$, output projection zero-initialized \\
    \midrule
    \multicolumn{2}{@{}l}{\emph{Auxiliary loss (diagnostic of Appendix~\ref{app:auxiliary})}} \\
    $\lambda$ & $\{0.1,\,0.3\}$ on the pre-fusion forecast, decoded to observation space; $\lambda=0$ is the \textsc{Gate} baseline; the auxiliary term is excluded from early stopping \\
    \midrule
    \multicolumn{2}{@{}l}{\emph{Mechanism-consistency probe (Sections~\ref{sec:mechanism-consistency} and~\ref{sec:method})}} \\
    Shift & $|\delta|=0.1$ along the empirical quantiles of the action channel, both directions \\
    Counted windows & the shift changes the plan, and $|\Delta^{i\to c}_\theta|>10^{-3}\,\mathrm{MAD}_c$ with $\mathrm{MAD}_c$ floored at $10^{-3}$; a mode condition must hold on at least half of the window's steps \\
    \bottomrule
    \end{tabular}
\end{table}

\subsection{Result aggregation and uncertainty}
\label{app:statistics}

Table~\ref{tab:sweeps} pools individual run scores before calculating their mean and
sample SD. For scores $m_1,\ldots,m_n$, we report
$\bar m=n^{-1}\sum_i m_i$ and
$s=\sqrt{\sum_i(m_i-\bar m)^2/(n-1)}$, with $n=35$ for prediction space
and fusion and $n=70$ for encoding. These SDs include differences between
backbones and, for encoding, fusion methods. They are distinct from both
within-configuration seed SDs and the dataset-level uncertainty estimates below.
Mean ranks are computed after averaging seeds within each dataset--backbone
configuration, and within each fusion method in the encoding experiment.

\paragraph{Paired effects.}
We average the five training seeds before comparing methods within each
configuration. A relative effect is the geometric mean of the paired MAE
ratios over dataset--backbone cells (Experiment~3 also fixes the fusion method), minus
one. Its $95\%$ interval is a percentile bootstrap over the eight datasets, which
resamples whole datasets with all their cells, so that an interval excluding zero marks
an effect that survives a change in the set of datasets. Next to it we count the
datasets on which the effect has the same sign.

\section{Extended Results of the Design-Choice Sweeps}
\label{app:sweeps}

This appendix supports Section~\ref{sec:benchmark-results}. It reports, for the three
sweeps of Table~\ref{tab:sweeps}, the paired effects with their uncertainty, computed as
described in Appendix~\ref{app:statistics}, and how the effects interact with the
backbone.

\subsection{Paired effects and uncertainty}

\begin{table}[H]
    \caption{Additional summaries of the three sweeps.
    References are observation space, \textsc{Concat}, and \textsc{Instant},
    respectively. Relative MAE and ranks use seed-averaged paired cells. Parameter counts are
    dataset-averaged fusion-adapter or temporal-encoder sizes, excluding the
    backbone, codec, and shared embeddings.}
    \label{tab:axis-summaries}
    \centering
    \small
    \setlength{\tabcolsep}{4pt}
    \begin{tabular}{llrrr}
\toprule
Sweep & Method & Mean rank & $\Delta$MAE vs.\ ref. & Params.  \\
\midrule
Prediction space & Observation & 2.52 & -- & --  \\
 & \textsc{AE} & 1.93 & $-9.87\%$ & --  \\
 & \textsc{VAE} & 1.95 & $-7.54\%$ & --  \\
 & \textsc{JEPA} & 3.61 & $+20.21\%$ & --  \\
\addlinespace
Plan fusion & \textsc{None} & 4.89 & $+5.73\%$ & 0  \\
 & \textsc{Concat} & 5.70 & -- & --  \\
 & \textsc{Res} & 3.43 & $-11.73\%$ & 38k  \\
 & \textsc{Res-0} & 3.66 & $-11.62\%$ & 38k  \\
 & \textsc{FiLM-0} & 2.91 & $-12.58\%$ & 43k  \\
 & \textsc{Gate} & 2.52 & $-12.73\%$ & 47k  \\
 & \textsc{X-attn} & 4.89 & $-7.27\%$ & 113k  \\
\addlinespace
Plan encoding & \textsc{Instant} & 2.79 & -- & 0  \\
 & \textsc{Decay} & 2.40 & $-0.81\%$ & 26  \\
 & \textsc{Conv} & 2.06 & $-2.20\%$ & 612  \\
 & \textsc{Attn} & 2.74 & $-1.15\%$ & 35.8k  \\
\bottomrule
\end{tabular}

\end{table}

\begin{table}[H]
    \caption{Paired effects of the three sweeps. $\Delta$MAE is the geometric-mean
    change in validation MAE of the first method against the second, with its $95\%$
    bootstrap interval over datasets; Datasets counts the datasets on which the first
    method is better. Negative values favor the first method.}
    \label{tab:benchmark-statistics}
    \centering
    \small
    \setlength{\tabcolsep}{4pt}
\begin{tabular}{llrcr}
\toprule
Exp. & Contrast & $\Delta$MAE & $95\%$ CI (bootstrap over datasets) & Datasets  \\
\midrule
1 & \textsc{AE} vs.\ observation & $-9.87\%$ & $[-17.71,-3.93]\%$ & 8/8  \\
1 & \textsc{VAE} vs.\ observation & $-7.54\%$ & $[-16.35,+1.96]\%$ & 7/8  \\
1 & \textsc{JEPA} vs.\ observation & $+20.21\%$ & $[+9.28,+37.40]\%$ & 0/8  \\
1 & \textsc{VAE} vs.\ \textsc{AE} & $+2.59\%$ & $[-1.93,+9.75]\%$ & 3/8  \\
\addlinespace
2 & \textsc{Gate} vs.\ \textsc{Concat} & $-12.73\%$ & $[-19.03,-7.12]\%$ & 8/8  \\
2 & \textsc{Gate} vs.\ \textsc{None} & $-17.47\%$ & $[-24.22,-10.90]\%$ & 8/8  \\
2 & \textsc{FiLM-0} vs.\ \textsc{Concat} & $-12.58\%$ & $[-19.19,-6.75]\%$ & 8/8  \\
2 & \textsc{FiLM-0} vs.\ \textsc{None} & $-17.32\%$ & $[-24.31,-10.36]\%$ & 8/8  \\
2 & \textsc{Gate} vs.\ \textsc{FiLM-0} & $-0.17\%$ & $[-0.84,+0.61]\%$ & 6/8  \\
\addlinespace
3 & \textsc{Decay} vs.\ \textsc{Instant} & $-0.81\%$ & $[-2.37,+0.44]\%$ & 4/8  \\
3 & \textsc{Conv} vs.\ \textsc{Instant} & $-2.20\%$ & $[-4.33,-0.13]\%$ & 4/8  \\
3 & \textsc{Attn} vs.\ \textsc{Instant} & $-1.15\%$ & $[-4.80,+1.99]\%$ & 3/8  \\
\bottomrule
\end{tabular}

\end{table}

\paragraph{Close pairs.} The two pairs that Section~\ref{sec:benchmark-results} calls
close are close on every column: \textsc{VAE} differs from \textsc{AE} by $+2.59\%$
with an interval of $[-1.93,+9.75]\%$ and is better on three of the eight datasets,
and \textsc{Gate} differs from \textsc{FiLM-0} by $-0.17\%$ with an interval of
$[-0.84,+0.61]\%$.

\subsection{Interaction with the backbone}

The complete grids of Appendix~\ref{app:grids} list the std over the five training
seeds of every cell. It is largest for \textsc{JEPA}, whose jointly trained encoder
makes some cells vary by more than their mean, and smaller for the output adapters than
for \textsc{Concat}; the frozen \textsc{AE} and \textsc{VAE} codecs are shared across
seeds, so their variation excludes codec retraining.

\begin{table}[H]
    \caption{How design choices and backbone rankings interact.
    Panel (a) ranks design choices within each backbone, then averages over
    datasets. Panel (b) ranks backbones within each choice. Lower is better;
    bold marks the best entry in the relevant comparison. An output adapter
    ranks first for each backbone, but the ordering of backbones changes with
    fusion. For example, the \textsc{None} and \textsc{Concat} backbone
    rankings have Kendall $\tau=0.05$.}
    \label{tab:per-backbone}
    \centering
    \scriptsize
    \setlength{\tabcolsep}{1.55pt}
    \textbf{(a) Mean rank of the arms within each backbone} (1 = that backbone's best arm)\\[1pt]
\begin{tabular}{lrrrr@{\hskip 9pt}rrrrrrr@{\hskip 9pt}rrrr}
\toprule
 & \multicolumn{4}{c}{Exp.\ 1: prediction space} & \multicolumn{7}{c}{Exp.\ 2: fusion} & \multicolumn{4}{c}{Exp.\ 3: encoder} \\
\cmidrule(lr){2-5}\cmidrule(lr){6-12}\cmidrule(lr){13-16}
Backbone & Observation & \textsc{AE} & \textsc{VAE} & \textsc{JEPA} & \textsc{None} & \textsc{Concat} & \textsc{Res} & \textsc{Res-0} & \textsc{FiLM-0} & \textsc{Gate} & \textsc{X-attn} & \textsc{Instant} & \textsc{Decay} & \textsc{Conv} & \textsc{Attn} \\
\midrule
TimeXer & 2.00 & \textbf{1.88} & 2.50 & 3.62 & 4.50 & 6.12 & 2.88 & 3.88 & 3.25 & \textbf{2.00} & 5.38 & 2.50 & 2.31 & \textbf{2.19} & 3.00 \\
TiDE & \textbf{1.75} & 2.00 & 2.25 & 4.00 & 5.00 & 5.50 & 3.50 & 3.38 & \textbf{3.00} & 3.00 & 4.62 & 2.88 & 2.75 & \textbf{1.88} & 2.50 \\
DUET & 2.75 & 1.75 & \textbf{1.50} & 4.00 & 5.00 & 6.88 & 3.00 & 3.25 & \textbf{1.88} & 3.12 & 4.88 & 2.50 & 2.44 & \textbf{1.94} & 3.12 \\
PatchTST & 3.12 & 1.88 & \textbf{1.50} & 3.50 & 3.88 & 6.62 & 4.25 & 3.38 & 2.88 & \textbf{2.25} & 4.75 & 3.12 & 2.25 & \textbf{1.88} & 2.75 \\
TimeKAN & 2.25 & \textbf{1.62} & 2.12 & 4.00 & 4.88 & 5.38 & 3.62 & 3.12 & 3.12 & \textbf{2.75} & 5.12 & 2.75 & 2.50 & \textbf{2.06} & 2.69 \\
CrossLinear & 3.62 & 2.25 & \textbf{2.00} & 2.12 & 7.00 & 3.50 & 3.12 & 4.50 & 2.38 & \textbf{2.00} & 5.50 & 2.69 & 2.44 & \textbf{2.12} & 2.75 \\
Amplifier & 2.12 & 2.12 & \textbf{1.75} & 4.00 & 4.00 & 5.88 & 3.62 & 4.12 & 3.88 & \textbf{2.50} & 4.00 & 3.12 & \textbf{2.12} & 2.38 & 2.38 \\
\bottomrule
\end{tabular}

\vspace{6pt}

\textbf{(b) Mean rank of the backbones within each arm} (1 = that arm's best backbone)\\[1pt]
\begin{tabular}{lrrrr@{\hskip 9pt}rrrrrrr@{\hskip 9pt}rrrr}
\toprule
 & \multicolumn{4}{c}{Exp.\ 1: prediction space} & \multicolumn{7}{c}{Exp.\ 2: fusion} & \multicolumn{4}{c}{Exp.\ 3: encoder} \\
\cmidrule(lr){2-5}\cmidrule(lr){6-12}\cmidrule(lr){13-16}
Backbone & Observation & \textsc{AE} & \textsc{VAE} & \textsc{JEPA} & \textsc{None} & \textsc{Concat} & \textsc{Res} & \textsc{Res-0} & \textsc{FiLM-0} & \textsc{Gate} & \textsc{X-attn} & \textsc{Instant} & \textsc{Decay} & \textsc{Conv} & \textsc{Attn} \\
\midrule
TimeXer & 3.00 & 3.88 & 5.12 & 3.38 & 3.62 & 3.88 & 3.12 & 3.62 & 3.62 & 3.50 & 3.88 & 3.38 & 3.38 & 3.75 & 3.88 \\
TiDE & \textbf{2.00} & 2.88 & 3.50 & 5.62 & 3.38 & 2.88 & 3.88 & 3.62 & 4.12 & 3.88 & 2.88 & 4.25 & 4.12 & 3.75 & 3.50 \\
DUET & 6.38 & 6.50 & 6.25 & 6.62 & 4.50 & 6.50 & 5.12 & 5.38 & 4.62 & 5.50 & 4.44 & 5.12 & 5.25 & 5.50 & 5.00 \\
PatchTST & 4.62 & 4.25 & 3.75 & 3.25 & \textbf{2.00} & 4.25 & 3.75 & 3.00 & \textbf{2.88} & \textbf{2.38} & \textbf{2.75} & \textbf{2.62} & \textbf{2.62} & \textbf{2.38} & \textbf{2.38} \\
TimeKAN & 2.75 & 3.12 & 3.12 & 3.38 & 3.75 & 3.12 & \textbf{2.62} & \textbf{2.88} & 3.38 & 3.25 & 3.12 & 3.12 & 3.25 & 3.12 & 3.75 \\
CrossLinear & 5.50 & \textbf{2.62} & \textbf{2.12} & \textbf{1.00} & 6.62 & \textbf{2.62} & 5.25 & 5.25 & 4.62 & 4.75 & 6.25 & 4.62 & 4.62 & 4.88 & 4.75 \\
Amplifier & 3.75 & 4.75 & 4.12 & 4.75 & 4.12 & 4.75 & 4.25 & 4.25 & 4.75 & 4.75 & 4.69 & 4.88 & 4.75 & 4.62 & 4.75 \\
\bottomrule
\end{tabular}

\end{table}

In Experiment~3, \textsc{Conv} ranks first or second for each backbone.
Its overall gain is similar under \textsc{Gate} and \textsc{FiLM-0}.
To assess that interaction, we compute each encoder's log MAE ratio to
\textsc{Instant} under each fusion method, subtract the two ratios, and
average over backbones within each dataset. The two ratios differ by less than $2\%$ on average for \textsc{Decay},
\textsc{Conv} and \textsc{Attn}, so the average encoding effect is similar under the
two retained fusion methods, while the dataset-specific gains remain heterogeneous.

\section{Mechanism Consistency}
\label{app:mechanism-consistency}

This appendix supports Table~\ref{tab:declared-mechanisms} of Section~\ref{sec:datasets}
and the mechanism-consistency paragraph of Section~\ref{sec:benchmark-results}. It states which
declared relations enter the metric and why the others are set aside, which windows are
counted, and gives the numbers behind Figure~\ref{fig:sec4-mechanism} together with the
score of every relation.

\subsection{Declared mechanisms and exclusions}
\label{app:mc-declared}

The six datasets with a continuous action
declare \McNpriorDecl{} relations, read from the documented process for the
engineered datasets and from the pharmacology of the drug for the clinical
ones. Two are set aside before any model is evaluated. The greenhouse
assimilation lamp runs at full output for $98.8\%$ of the held-out steps, so
the upward quantile shift $\tau_{i,\delta}$ changes the plan in almost no
window and the conditioning event of $\mathrm{MC}$ nearly never holds. The
PleiaData on/off relation is categorical, and $\tau_{i,\delta}$ is defined for
continuous channels only; Appendix~\ref{app:directional} probes it by switching the
unit on and reports it separately. The remaining \McNpriorAll{} are listed in
Table~\ref{tab:mechanism-priors}. One of them, the single relation of
Wastewater, is set aside after evaluation and is reported in the table but not
counted. It states that metal-salt dosing lowers the phosphate concentration in
the tank by precipitation, a mechanism that acts through the reaction and the
residence of the water in the tank, on a timescale beyond the $32$-minute
horizon ($H=16$ steps at $2$\,min) over which $\mathrm{MC}$ averages the
response, so the declared sign is not expected to appear within the window the
metric scores. Its scores, $\McWwMin$ to $\McWwMax$ across the design choices,
are the lowest in the grid; we report them so that this exclusion, made after
the numbers were seen, stays visible. Section~\ref{sec:mechanism-consistency}
therefore averages the \McNprior{} relations of the remaining five datasets.

\subsection{Counted windows and reproducibility check}

$\mathrm{MC}$ conditions on the
perturbation changing the plan and on the forecast moving. Both conditions bite.
A channel already at its maximum cannot be shifted up, and a model that routes
the plan into a path with no influence on the output produces
$|\Delta^{i\to c}_\theta|$ below the threshold in every window; those cells have
no defined consistency and are reported as such rather than counted as chance.
\textsc{None} is the extreme case, since it never receives the plan, but it is
not the only one: $\McDeadCells$ of the $\McDeadTotal$ cells that do receive
the plan also fail to move, and every one of them is TiDE under a prediction
space that routes the plan through the backbone rather than an adapter.

Every number of Section~\ref{sec:mechanism-consistency} comes from the checkpoints of the three experiments; no model is
retrained. Reloading a checkpoint and recomputing its validation MAE reproduces
the stored value to $\McSelfcheck$ for every backbone but DUET, whose
mixture-of-experts router samples at inference and drifts by at most
$\McSelfcheckDuet$. Within one probed cell the router is held to a single sample
across the factual and the perturbed forecast, which is what a counterfactual
requires: the same unit, the same history, and the same draw of any internal
randomness, with only the plan changed. The factual forecast is verified to be
bitwise identical when the plan is passed twice.

\subsection{Results per design choice and per mechanism}

Table~\ref{tab:mechanism-consistency} gives the numbers drawn in
Figure~\ref{fig:sec4-mechanism}, and Table~\ref{tab:mechanism-priors} the score of
every declared relation under the three fusion methods that
Section~\ref{sec:benchmark-results} discusses.

\begin{table}[H]
    \caption{Mechanism consistency of the three sweeps, the numbers behind
    Figure~\ref{fig:sec4-mechanism}. The
    $\mathrm{MC}$ metric of Section~\ref{sec:setup} at $\delta=\pm0.1$, averaged
    over both directions and over the mechanisms a dataset declares, as mean
    $\pm$ std over the $35$ backbone--seed runs of one dataset and one option,
    or $70$ for encoding, which pools \textsc{Gate} and \textsc{FiLM-0}. Chance
    is $0.5$ and $1$ is perfect agreement with the
    declared mechanism. \textsc{None} discards the plan, so its forecast does
    not move and it is omitted. Bold marks the highest unrounded mean of each
    block in each column; Mean averages the five datasets. Wastewater is set
    aside, see Appendix~\ref{app:mc-declared} and Table~\ref{tab:mechanism-priors}. Higher is
    better.}
    \label{tab:mechanism-consistency}
    \centering
    \small
    \setlength{\tabcolsep}{4pt}
    \renewcommand{\arraystretch}{1.0}
\begin{tabular}{llcccccc}
\toprule
 & Option & Greenhouse & PleiaData & PREDIST & VitalDB & MIMIC-Cardio & Mean \\
\midrule
\multirow{4}{*}{\rotatebox[origin=c]{90}{Space}} & \textsc{Observation} & \textbf{.61}$_{\pm.13}$ & .49$_{\pm.10}$ & .97$_{\pm.03}$ & .51$_{\pm.10}$ & \textbf{.35}$_{\pm.14}$ & \textbf{.59} \\
 & \textsc{AE} & .55$_{\pm.17}$ & .51$_{\pm.11}$ & .92$_{\pm.15}$ & \textbf{.55}$_{\pm.10}$ & .33$_{\pm.17}$ & .57 \\
 & \textsc{VAE} & .47$_{\pm.26}$ & \textbf{.53}$_{\pm.15}$ & \textbf{.98}$_{\pm.03}$ & .47$_{\pm.05}$ & .33$_{\pm.08}$ & .55 \\
 & \textsc{JEPA} & .55$_{\pm.15}$ & .44$_{\pm.16}$ & .88$_{\pm.13}$ & .51$_{\pm.11}$ & .33$_{\pm.10}$ & .54 \\
\midrule
\multirow{6}{*}{\rotatebox[origin=c]{90}{Fusion}} & \textsc{Concat} & \textbf{.55}$_{\pm.17}$ & .51$_{\pm.11}$ & .92$_{\pm.15}$ & \textbf{.55}$_{\pm.10}$ & .33$_{\pm.17}$ & \textbf{.57} \\
 & \textsc{Res} & .49$_{\pm.11}$ & .46$_{\pm.10}$ & .97$_{\pm.02}$ & .35$_{\pm.09}$ & \textbf{.48}$_{\pm.09}$ & .55 \\
 & \textsc{Res-0} & .41$_{\pm.15}$ & .47$_{\pm.13}$ & .97$_{\pm.02}$ & .36$_{\pm.11}$ & .40$_{\pm.12}$ & .52 \\
 & \textsc{FiLM-0} & .43$_{\pm.14}$ & .46$_{\pm.07}$ & .98$_{\pm.02}$ & .36$_{\pm.11}$ & .42$_{\pm.08}$ & .53 \\
 & \textsc{Gate} & .47$_{\pm.11}$ & .49$_{\pm.12}$ & .98$_{\pm.02}$ & .36$_{\pm.08}$ & .41$_{\pm.09}$ & .54 \\
 & \textsc{X-attn} & .42$_{\pm.13}$ & \textbf{.52}$_{\pm.06}$ & \textbf{.99}$_{\pm.01}$ & .15$_{\pm.19}$ & .21$_{\pm.11}$ & .46 \\
\midrule
\multirow{4}{*}{\rotatebox[origin=c]{90}{Encoding}} & \textsc{Instant} & .45$_{\pm.12}$ & .47$_{\pm.10}$ & .98$_{\pm.02}$ & \textbf{.36}$_{\pm.10}$ & .42$_{\pm.09}$ & .54 \\
 & \textsc{Decay} & .47$_{\pm.12}$ & \textbf{.50}$_{\pm.10}$ & .93$_{\pm.03}$ & .34$_{\pm.10}$ & .42$_{\pm.09}$ & .53 \\
 & \textsc{Conv} & \textbf{.56}$_{\pm.15}$ & .50$_{\pm.10}$ & .96$_{\pm.02}$ & .25$_{\pm.11}$ & \textbf{.52}$_{\pm.10}$ & \textbf{.56} \\
 & \textsc{Attn} & .44$_{\pm.12}$ & .49$_{\pm.06}$ & \textbf{.98}$_{\pm.01}$ & .30$_{\pm.14}$ & .44$_{\pm.10}$ & .53 \\
\bottomrule
\end{tabular}

\end{table}

\begin{table}[H]
    \caption{Mechanism consistency per declared relation under the three
    fusion methods that Section~\ref{sec:benchmark-results} discusses. Mean over
    $35$ backbone--seed cells; the four engineered datasets first, then the two
    clinical ones. Sign is the declared direction. Seed range is the median over backbones of the spread of the five
    seeds of one cell configuration on that relation.}
    \label{tab:mechanism-priors}
    \centering
    \small
    \setlength{\tabcolsep}{4pt}
    \renewcommand{\arraystretch}{1.0}
\begin{tabular}{llccccc}
\toprule
Dataset & Mechanism & Sign & \textsc{Concat} & \textsc{Gate} & \textsc{FiLM-0} & seed range \\
\midrule
\multirow{4}{*}{Greenhouse} & Tpipe$\to$Tair & $+$ & .83 & .97 & .99 & .03 \\
 & CO2dosing$\to$CO2air & $+$ & .48 & .08 & .17 & .29 \\
 & VentLee$\to$Tair & $-$ & .61 & .64 & .36 & .77 \\
 & VentWind$\to$Tair & $-$ & .28 & .19 & .22 & .66 \\
\midrule
\multirow{2}{*}{PleiaData} & setpoint\_norm@mode=1$\to$indoor\_temp & $+$ & .80 & .60 & .50 & .75 \\
 & setpoint\_norm@mode=2$\to$indoor\_temp & $-$ & .21 & .37 & .42 & .57 \\
\midrule
\multirow{3}{*}{PREDIST} & valve\_pos$\to$flow & $+$ & .91 & .97 & .96 & .07 \\
 & valve\_pos$\to$heat\_power & $+$ & .92 & .97 & .97 & .05 \\
 & supply\_setpoint$\to$s\_hc1\_supply\_temp & $+$ & .92 & 1.00 & 1.00 & .01 \\
\midrule
\multirow{1}{*}{Wastewater} & metal\_dosing$\to$po4 & $-$ & .15 & .03 & .07 & .10 \\
\midrule
\multirow{4}{*}{VitalDB} & propofol$\to$BIS & $-$ & .63 & .24 & .34 & .31 \\
 & propofol$\to$ART\_MBP & $-$ & .38 & .56 & .52 & .49 \\
 & remifentanil$\to$HR & $-$ & .54 & .35 & .36 & .33 \\
 & remifentanil$\to$ART\_MBP & $-$ & .65 & .27 & .23 & .26 \\
\midrule
\multirow{8}{*}{MIMIC-Cardio} & norepinephrine$\to$NBPs & $+$ & .21 & .20 & .14 & .24 \\
 & norepinephrine$\to$NBPm & $+$ & .15 & .14 & .13 & .18 \\
 & phenylephrine$\to$NBPs & $+$ & .29 & .15 & .27 & .29 \\
 & phenylephrine$\to$NBPm & $+$ & .29 & .11 & .16 & .26 \\
 & vasopressin$\to$NBPm & $+$ & .34 & .62 & .53 & .42 \\
 & dobutamine$\to$HR & $+$ & .55 & .86 & .92 & .33 \\
 & dopamine$\to$HR & $+$ & .42 & .99 & .92 & .05 \\
 & FiO2$\to$SpO2 & $+$ & .26 & .22 & .30 & .41 \\
\bottomrule
\end{tabular}

\end{table}

\paragraph{Two relations on one actuator.} Averaging a dataset's relations hides a
sharper failure. PleiaData declares opposite signs for the same setpoint channel
in the two modes of the heat pump, so a model that has learned the mechanism
must get both. Across the cells of that dataset only $\McPlJoint$ do, and the two
relations are anticorrelated across cells at $\rho=\McPlRho$: a cell that gets
one right gets the other wrong, which is the signature of a single
mode-independent sign rather than a mode-dependent mechanism. The two greenhouse
vents, whose apertures are nearly collinear in the training data, are also rarely both
right, $\McGhJoint$ cells, but there the two relations are uncorrelated
across cells rather than anticorrelated, so that pair is a weaker instance of
the same point.

\par
\endgroup

\begingroup
\raggedbottom
\section{Directional Supervision: Details and Extended Results}
\label{app:directional}

This appendix supports Section~\ref{sec:method}. It gives the hyperparameters of the
added term and how $\rho$ and the penalized set were fixed, the score of every mechanism
under both shift directions, and the prediction error of every dataset including the
null control.

\subsection{Hyperparameters and the penalized set}

Directional supervision adds the term of Equation~\ref{eq:penalty} to the forecasting
protocol of Table~\ref{tab:hyperparams} and changes nothing else: the same
\textsc{AE}--\textsc{Gate}--\textsc{Instant} model, backbones, seeds, split, windows,
optimizer and stopping rule; each mechanism costs one extra pass through $G_\theta$
and $D$ per batch. The term uses $\rho=0.1$, one shift size per batch drawn
from $\delta\sim\mathcal{U}(0.05,0.30)$ and applied upward only, $\varepsilon=10^{-6}$,
and the quantiles of $\tau_{i,\delta}$ estimated on the training actions; it is
excluded from early stopping, which uses the forecasting loss alone. The value of
$\rho$ follows a rule fixed before the grid was run: the smallest value in
$\{0.1,1,10\}$ whose consistency on the penalized mechanisms of the training windows
reached $0.95$ in a Greenhouse pilot with three backbones. All three values reached
$1.000$, so the rule selects $\rho=0.1$.

The penalized set holds twelve of the \McNprior{} mechanisms of
Table~\ref{tab:declared-mechanisms} and was fixed before any supervised model was
trained. Mechanisms were assigned by action channel, so that the two targets of the
PREDIST valve and the two targets of norepinephrine are on the same side and no
held-out mechanism shares its actuator with a penalized one; the single exception is
the PleiaData cooling-mode setpoint, which shares its channel with the penalized
heating-mode mechanism and differs only in the mode condition, so it tests whether
the learned response follows the mode. On the clinical datasets the held-out side was
chosen to contain both a mechanism that \textsc{Gate} already gets right, dopamine,
and mechanisms it gets wrong, phenylephrine and remifentanil. Wastewater has no
penalized mechanism: its single relation is set aside in
Appendix~\ref{app:mc-declared}, so the added term is identically zero there and the
dataset serves as a null control. The PleiaData on/off relation is listed with the
penalized set in the code but the term is defined for continuous channels only and is
zero for it, so Figure~\ref{fig:soft} leaves it out of the PleiaData mean. It is probed
by switching the unit from off to on instead of by the quantile shift, and
Table~\ref{tab:directional-mechanisms} reports it with Wastewater in a separate block.

\subsection{Results per mechanism}

\begin{table}[t]
    \caption{Mechanism consistency of every declared relation with and
    without directional supervision. Mean over the $35$ backbone--seed runs of
    \textsc{Gate} and of \textsc{Gate} $+$ directional supervision ($+$DS) under the
    upward and the downward shift; Pen.\ marks the twelve penalized mechanisms, the
    training shift is upward only. The last column is the lowest of the $35$ runs of
    $+$DS under the upward shift. The last block holds the two relations that
    Section~\ref{sec:datasets} sets aside; the on/off relation is categorical and has
    no downward shift.}
    \label{tab:directional-mechanisms}
    \centering
    \small
    \setlength{\tabcolsep}{1.8pt}
    \resizebox{\textwidth}{!}{
\begin{tabular}{llcc@{\hskip 10pt}cc@{\hskip 10pt}cc@{\hskip 10pt}c}
\toprule
 & & & & \multicolumn{2}{c}{$\delta=+0.1$} & \multicolumn{2}{c}{$\delta=-0.1$} & lowest cell \\
\cmidrule(lr){5-6}\cmidrule(lr){7-8}
Dataset & Action $\to$ State & Sign & Pen. & \textsc{Gate} & $+$DS & \textsc{Gate} & $+$DS & $+$DS, $\delta=+0.1$ \\
\midrule
Greenhouse & heating-pipe temp. $\to$ air temp. & $+$ & $\checkmark$ & 0.973 & 1.000 & 0.974 & 1.000 & 1.000 \\
Greenhouse & leeward vent $\to$ air temp. & $-$ & $\checkmark$ & 0.653 & 1.000 & 0.634 & 1.000 & 0.991 \\
Greenhouse & windward vent $\to$ air temp. & $-$ & $\checkmark$ & 0.196 & 1.000 & 0.181 & 1.000 & 1.000 \\
Greenhouse & CO$_2$ dosing $\to$ CO$_2$ level & $+$ &  & 0.097 & 0.154 & 0.068 & 0.126 & 0.000 \\
\addlinespace[2pt]
PleiaData & setpoint (heating) $\to$ indoor temp. & $+$ & $\checkmark$ & 0.606 & 1.000 & 0.601 & 1.000 & 1.000 \\
PleiaData & setpoint (cooling) $\to$ indoor temp. & $-$ &  & 0.373 & 0.000 & 0.376 & 0.000 & 0.000 \\
\addlinespace[2pt]
PREDIST & valve position $\to$ flow & $+$ & $\checkmark$ & 0.966 & 1.000 & 0.965 & 1.000 & 0.999 \\
PREDIST & valve position $\to$ heat power & $+$ & $\checkmark$ & 0.969 & 1.000 & 0.970 & 1.000 & 0.997 \\
PREDIST & supply setpoint $\to$ supply temp. & $+$ &  & 0.999 & 0.998 & 0.999 & 0.997 & 0.989 \\
\addlinespace[2pt]
VitalDB & propofol $\to$ BIS & $-$ & $\checkmark$ & 0.218 & 1.000 & 0.258 & 1.000 & 1.000 \\
VitalDB & propofol $\to$ MAP & $-$ & $\checkmark$ & 0.574 & 1.000 & 0.551 & 1.000 & 1.000 \\
VitalDB & remifentanil $\to$ heart rate & $-$ &  & 0.370 & 0.306 & 0.330 & 0.270 & 0.083 \\
VitalDB & remifentanil $\to$ MAP & $-$ &  & 0.284 & 0.156 & 0.258 & 0.119 & 0.004 \\
\addlinespace[2pt]
MIMIC-Cardio & norepinephrine $\to$ systolic pressure & $+$ & $\checkmark$ & 0.218 & 1.000 & 0.188 & 1.000 & 1.000 \\
MIMIC-Cardio & norepinephrine $\to$ mean pressure & $+$ & $\checkmark$ & 0.150 & 1.000 & 0.127 & 1.000 & 1.000 \\
MIMIC-Cardio & phenylephrine $\to$ systolic pressure & $+$ &  & 0.180 & 0.236 & 0.127 & 0.198 & 0.004 \\
MIMIC-Cardio & phenylephrine $\to$ mean pressure & $+$ &  & 0.103 & 0.128 & 0.109 & 0.155 & 0.000 \\
MIMIC-Cardio & vasopressin $\to$ mean pressure & $+$ &  & 0.531 & 0.182 & 0.702 & 0.274 & 0.003 \\
MIMIC-Cardio & dobutamine $\to$ heart rate & $+$ & $\checkmark$ & 0.864 & 1.000 & 0.852 & 1.000 & 1.000 \\
MIMIC-Cardio & dopamine $\to$ heart rate & $+$ &  & 0.999 & 0.978 & 0.987 & 0.992 & 0.230 \\
MIMIC-Cardio & FiO$_2$ $\to$ oxygen saturation & $+$ & $\checkmark$ & 0.194 & 1.000 & 0.240 & 1.000 & 1.000 \\
\midrule
\multicolumn{9}{@{}l}{\emph{Set aside in Section~4 (Appendix~\ref{app:mechanism-consistency})}} \\
Wastewater & metal-salt dosing $\to$ phosphate & $-$ & & 0.031 & 0.033 & 0.028 & 0.029 & 0.000 \\
PleiaData & unit on/off $\to$ energy per step & $+$ & & 0.579 & 0.500 & --- & --- & 0.000 \\
\bottomrule
\end{tabular}
}
\end{table}

Table~\ref{tab:directional-mechanisms} gives the numbers behind the left panel of
Figure~\ref{fig:soft} and adds the downward shift and the held-out mechanisms. On
every penalized mechanism the mean of $+$DS rounds to $1.000$ in both directions, the
heating pipe included, which \textsc{Gate} already gets right at $0.973$; the lowest
single run is $0.991$, on the leeward vent. The downward shift, which the
term never sees, is as consistent as the upward one. On the held-out mechanisms the
picture is mixed and is reported in full. Two stay where \textsc{Gate} left them,
the PREDIST supply setpoint and dopamine, both already consistent. Four move by less
than $0.08$ under either shift and stay well below chance: CO$_2$ dosing and both
phenylephrine relations rise slightly, and remifentanil on heart rate falls slightly.
Three move away from the declared sign: the PleiaData cooling-mode setpoint falls from
$0.373$ to $0.000$, remifentanil on mean pressure from $0.284$ to $0.156$, and
vasopressin on mean pressure from $0.531$ to $0.182$. The cooling-mode case is the
conditional mechanism above: the learned heating-mode response extends to the cooling
mode, so the model has a single mode-independent response, the failure that
Appendix~\ref{app:mechanism-consistency} already found in the unsupervised models,
and supervising one mode does not correct it. Remifentanil and vasopressin on mean
pressure each share their state channel with a penalized mechanism, propofol and
norepinephrine respectively, but not its action channel; we have no evidence for why
their scores fall, and note only that the supervision does not constrain them. Directional supervision therefore enforces the
signs it is given and should not be expected to transfer to a mechanism it is not
given, including one that shares a channel with a penalized mechanism.

\subsection{Prediction error and the null control}

\begin{table}[t]
    \caption{Validation MAE with and without directional supervision. Mean
    over the $35$ backbone--seed runs of each dataset; Pen.\ is the number of
    penalized mechanisms. $\Delta$MAE is the median of the paired relative change and the interval is a
    $t$-based $90\%$ interval on the paired log ratios, as in Figure~\ref{fig:soft}. The
    pooled row pairs all $210$ runs. Wastewater has no penalized mechanism and is the
    null control.}
    \label{tab:directional-mae}
    \centering
    \small
    \setlength{\tabcolsep}{4pt}
\begin{tabular}{lrcccc}
\toprule
Dataset & Pen. & \textsc{Gate} & $+$DS & $\Delta$MAE (median) & $90\%$ interval \\
\midrule
Greenhouse & 3 & 0.0632 & 0.0639 & +0.03\% & $[-0.24,+2.11]\%$ \\
PleiaData & 1 & 0.0205 & 0.0198 & $-$1.54\% & $[-9.60,+3.34]\%$ \\
PREDIST & 2 & 0.1641 & 0.1641 & +0.35\% & $[-0.34,+0.38]\%$ \\
VitalDB & 2 & 0.0313 & 0.0312 & $-$0.07\% & $[-0.70,+0.33]\%$ \\
MIMIC-Cardio & 4 & 0.0543 & 0.0542 & $-$0.15\% & $[-0.26,-0.03]\%$ \\
Wastewater & 0 & 0.1500 & 0.1498 & +0.00\% & $[-0.59,+0.34]\%$ \\
\midrule
Pooled & 12 & 0.0806 & 0.0805 & $-$0.05\% & $[-1.58,+0.63]\%$ \\
\bottomrule
\end{tabular}

\end{table}

Table~\ref{tab:directional-mae} gives the numbers behind the right panel of
Figure~\ref{fig:soft} and adds Wastewater. On the null control the two arms are the
same model trained twice under the same seed, since the added term is identically
zero, and their MAE agrees to $0.0002$ with a median paired change of $0.00\%$; this
verifies that the training path of the supervised arm matches \textsc{Gate} apart
from the term itself. On the five datasets with penalized mechanisms the median
change lies between $-1.54\%$ and $+0.35\%$, and every interval lies within $\pm5\%$
except on PleiaData, whose spread across runs is the widest; MIMIC-Cardio shows a
small decrease.

\par
\endgroup

\begingroup
\raggedbottom
\section{Auxiliary Prediction Loss and Rollout Diagnostic}
\label{app:auxiliary}

This appendix reports the diagnostic that Section~\ref{sec:benchmark-setup} mentions.
As a separate diagnostic, we fix \textsc{AE}, \textsc{Gate}, and
\textsc{Instant} and add supervision to the pre-fusion state forecast:
\begin{equation}
    \mathcal{L}_{\mathrm{prediction}}
    = \mathrm{MSE}(\hat y,y)
    + \lambda\,\mathrm{MSE}\bigl(D(z_{\mathrm{pre}}),y\bigr),
    \qquad \lambda\in\{0,0.1,0.3\}.
    \label{eq:benchmark-auxiliary}
\end{equation}
Here $z_{\mathrm{pre}}$ is the backbone forecast before fusion, $D$ is the
frozen decoder, and $\hat y$ is the full model's output. Both terms are in
observation space; backbone-specific training regularizers are retained.
This is an additional objective comparison, not a fourth architectural sweep.

Equation~\ref{eq:benchmark-auxiliary} adds observation-space supervision to
the decoded pre-fusion forecast. The validation criterion used for checkpoint
selection excludes this added term, retaining the full-model prediction
criterion. The bypass metric decodes the pre-fusion forecast alone for a single
$16$-step block. It is not a forecast of a no-action intervention: its training
target is still the observed outcome under the recorded plan.
At $\lambda=0.1$, the bypass MAE improves by $44.73\%$ while the full-model
MAE changes by $+1.75\%$. The latter has a dataset-level $90\%$ interval of
$[-0.20,+3.74]\%$, within $\pm5\%$. At $\lambda=0.3$, full-model MAE changes by
$+2.78\%$, with an interval of $[-0.19,+5.84]\%$ that reaches beyond $5\%$.

\begin{table}[H]
    \caption{Auxiliary-loss results by dataset.
    The first two blocks report $16$-step full-model and pre-fusion MAE over
    seven backbones and five seeds. Rollout error is for steps $49$--$64$,
    not the cumulative horizon. It uses $273$ matched triples: $35$ per dataset
    except VitalDB ($28$; all TimeXer runs and two TimeKAN seeds missing).
    Subscripts are SDs over five seed means of available backbones; VitalDB
    coverage varies by seed. Bold marks the lowest unrounded mean of each row.
    Relative changes use seed-averaged cells in the first two blocks and
    matched backbone--seed triples for rollout.}
    \label{tab:exp4-autonomy}
    \centering
    \small
    \setlength{\tabcolsep}{8pt}
    \begin{tabular}{@{}lccc@{}}
\toprule
Dataset & $\lambda\!=\!0$ & $\lambda\!=\!0.1$ & $\lambda\!=\!0.3$ \\
\midrule
\multicolumn{4}{@{}l}{\emph{MAE (fused, $H\!=\!16$)}} \\
Greenhouse & \textbf{0.0632}$_{\pm.0005}$ & 0.0651$_{\pm.0004}$ & 0.0658$_{\pm.0009}$ \\
VitalDB & 0.0313$_{\pm.0002}$ & 0.0312$_{\pm.0001}$ & \textbf{0.0312}$_{\pm.0001}$ \\
CGMacros & \textbf{0.0086}$_{\pm.0001}$ & 0.0095$_{\pm.0001}$ & 0.0098$_{\pm.0002}$ \\
Shanghai & \textbf{0.1165}$_{\pm.0010}$ & 0.1174$_{\pm.0010}$ & 0.1177$_{\pm.0007}$ \\
MIMIC-Cardio & \textbf{0.0543}$_{\pm.0001}$ & 0.0545$_{\pm.0001}$ & 0.0545$_{\pm.0001}$ \\
PleiaData & 0.0205$_{\pm.0011}$ & \textbf{0.0204}$_{\pm.0007}$ & 0.0212$_{\pm.0005}$ \\
PREDIST & 0.1641$_{\pm.0006}$ & 0.1640$_{\pm.0013}$ & \textbf{0.1639}$_{\pm.0002}$ \\
Wastewater & 0.1500$_{\pm.0011}$ & 0.1505$_{\pm.0009}$ & \textbf{0.1499}$_{\pm.0011}$ \\
$\Delta$ vs.\ $\lambda\!=\!0$ & -- & $+1.75\%$ & $+2.78\%$ \\
\midrule
\multicolumn{4}{@{}l}{\emph{MAE$_{\mathrm{bypass}}$ (pre-fusion, $H\!=\!16$)}} \\
Greenhouse & 0.1637$_{\pm.0027}$ & 0.0859$_{\pm.0015}$ & \textbf{0.0814}$_{\pm.0026}$ \\
VitalDB & 0.0961$_{\pm.0002}$ & 0.0430$_{\pm.0003}$ & \textbf{0.0411}$_{\pm.0003}$ \\
CGMacros & 0.1397$_{\pm.0062}$ & 0.0253$_{\pm.0030}$ & \textbf{0.0203}$_{\pm.0024}$ \\
Shanghai & 0.1852$_{\pm.0022}$ & 0.1435$_{\pm.0042}$ & \textbf{0.1322}$_{\pm.0013}$ \\
MIMIC-Cardio & 0.1147$_{\pm.0003}$ & 0.0626$_{\pm.0002}$ & \textbf{0.0614}$_{\pm.0001}$ \\
PleiaData & 0.1735$_{\pm.0095}$ & 0.0361$_{\pm.0010}$ & \textbf{0.0325}$_{\pm.0007}$ \\
PREDIST & 0.2333$_{\pm.0009}$ & 0.1953$_{\pm.0008}$ & \textbf{0.1906}$_{\pm.0011}$ \\
Wastewater & 0.2832$_{\pm.0044}$ & 0.2122$_{\pm.0018}$ & \textbf{0.2047}$_{\pm.0014}$ \\
$\Delta$ vs.\ $\lambda\!=\!0$ & -- & $-44.73\%$ & $-47.83\%$ \\
\midrule
\multicolumn{4}{@{}l}{\emph{Rollout MAE (steps 49--64)}} \\
Greenhouse & \textbf{0.1160}$_{\pm.0013}$ & 0.1195$_{\pm.0017}$ & 0.1204$_{\pm.0031}$ \\
VitalDB & \textbf{0.0587}$_{\pm.0005}$ & 0.0592$_{\pm.0005}$ & 0.0590$_{\pm.0004}$ \\
CGMacros & \textbf{0.0733}$_{\pm.0003}$ & 0.0755$_{\pm.0009}$ & 0.0756$_{\pm.0011}$ \\
Shanghai & \textbf{0.1260}$_{\pm.0006}$ & 0.1278$_{\pm.0015}$ & 0.1295$_{\pm.0009}$ \\
MIMIC-Cardio & \textbf{0.0730}$_{\pm.0002}$ & 0.0732$_{\pm.0001}$ & 0.0735$_{\pm.0002}$ \\
PleiaData & \textbf{0.1203}$_{\pm.0015}$ & 0.1209$_{\pm.0007}$ & 0.1240$_{\pm.0006}$ \\
PREDIST & \textbf{0.1989}$_{\pm.0013}$ & 0.1998$_{\pm.0021}$ & 0.1997$_{\pm.0012}$ \\
Wastewater & \textbf{0.1939}$_{\pm.0022}$ & 0.1953$_{\pm.0017}$ & 0.1951$_{\pm.0013}$ \\
$\Delta$ vs.\ $\lambda\!=\!0$ & -- & $+1.19\%$ & $+1.71\%$ \\
\bottomrule
\end{tabular}

\end{table}

\begin{table}[H]
    \caption{Sensitivity to backbone inclusion.
    Each row removes one backbone and recomputes the geometric-mean change
    relative to $\lambda=0$. Negative values indicate lower MAE.
    Removing CrossLinear changes the average full-model effect to approximately
    zero while retaining a large bypass gain, showing where the observed cost
    is concentrated.}
    \label{tab:exp4-lobo}
    \centering
    \small
    \setlength{\tabcolsep}{4pt}
    \begin{tabular}{lrrrr}
\toprule
Backbone dropped & \multicolumn{2}{c}{$\Delta$MAE} & \multicolumn{2}{c}{$\Delta$MAE$_{\mathrm{bypass}}$} \\
\cmidrule(lr){2-3}\cmidrule(lr){4-5}
 & $\lambda\!=\!0.1$ & $\lambda\!=\!0.3$ & $\lambda\!=\!0.1$ & $\lambda\!=\!0.3$ \\
\midrule
\emph{none (full set)} & $+1.75\%$ & $+2.78\%$ & $-44.73\%$ & $-47.83\%$ \\
\addlinespace
TimeXer & $+2.35\%$ & $+3.73\%$ & $-45.24\%$ & $-48.53\%$ \\
TiDE & $+2.16\%$ & $+3.37\%$ & $-45.90\%$ & $-49.20\%$ \\
DUET & $+2.05\%$ & $+2.90\%$ & $-45.85\%$ & $-48.87\%$ \\
PatchTST & $+1.95\%$ & $+3.26\%$ & $-45.70\%$ & $-48.63\%$ \\
TimeKAN & $+1.93\%$ & $+3.03\%$ & $-46.06\%$ & $-49.26\%$ \\
CrossLinear & $-0.22\%$ & $-0.02\%$ & $-37.92\%$ & $-40.85\%$ \\
Amplifier & $+2.05\%$ & $+3.26\%$ & $-45.97\%$ & $-48.96\%$ \\
\bottomrule
\end{tabular}

\end{table}

Rollout feeds the full model's predictions back as state history, while
teacher forcing restores observed history; both retain the recorded plan.
The first block is identical. At each later block, we subtract the relative
change in teacher-forced MAE from the relative change in rollout MAE, both
against $\lambda=0$. Negative differences indicate a rollout-specific gain.
Neither weight shows a consistent rollout-specific advantage at the blocks ending
at steps $32$, $48$ and $64$; $\lambda=0$ also has the lowest final-block mean
on all eight datasets. These results apply to the available matched runs.

\section{Complete Validation Grids}
\label{app:grids}

Tables~\ref{tab:cells-exp1}--\ref{tab:cells-exp3-film} show every configuration
behind Table~\ref{tab:sweeps}: five-seed
means with seed SDs, with the lowest unrounded mean in each row in bold.

\begin{table}[H]
    \caption{Experiment 1, prediction space: the complete grid.
    Validation MAE, mean and SD over five training seeds.}
    \label{tab:cells-exp1}
    \centering
    \small
    \setlength{\tabcolsep}{4pt}
    \renewcommand{\arraystretch}{1.0}
    \begin{tabular}{llcccc}
\toprule
Dataset & Backbone & Observation & \textsc{AE} & \textsc{VAE} & \textsc{JEPA} \\
\midrule
Greenhouse & TimeXer & 0.0627$_{\pm.0008}$ & \textbf{0.0604}$_{\pm.0006}$ & 0.1051$_{\pm.0009}$ & 0.0882$_{\pm.0206}$ \\
 & TiDE & \textbf{0.0635}$_{\pm.0026}$ & 0.0658$_{\pm.0027}$ & 0.1094$_{\pm.0007}$ & 0.1158$_{\pm.0152}$ \\
 & DUET & 0.0920$_{\pm.0084}$ & \textbf{0.0709}$_{\pm.0016}$ & 0.0962$_{\pm.0134}$ & 0.1453$_{\pm.0261}$ \\
 & PatchTST & 0.0986$_{\pm.0334}$ & 0.0924$_{\pm.0334}$ & \textbf{0.0921}$_{\pm.0025}$ & 0.1019$_{\pm.0152}$ \\
 & TimeKAN & \textbf{0.0620}$_{\pm.0007}$ & 0.0642$_{\pm.0011}$ & 0.0770$_{\pm.0022}$ & 0.0894$_{\pm.0282}$ \\
 & CrossLinear & 0.1147$_{\pm.0327}$ & 0.1066$_{\pm.0024}$ & 0.0996$_{\pm.0142}$ & \textbf{0.0827}$_{\pm.0065}$ \\
 & Amplifier & 0.0814$_{\pm.0265}$ & \textbf{0.0723}$_{\pm.0092}$ & 0.0898$_{\pm.0164}$ & 0.1051$_{\pm.0196}$ \\
\addlinespace
VitalDB & TimeXer & 0.0357$_{\pm.0009}$ & \textbf{0.0326}$_{\pm.0014}$ & 0.0356$_{\pm.0012}$ & 0.0339$_{\pm.0043}$ \\
 & TiDE & 0.0303$_{\pm.0002}$ & 0.0310$_{\pm.0001}$ & \textbf{0.0303}$_{\pm.0002}$ & 0.0444$_{\pm.0096}$ \\
 & DUET & 0.0446$_{\pm.0043}$ & \textbf{0.0384}$_{\pm.0013}$ & 0.0417$_{\pm.0020}$ & 0.0631$_{\pm.0059}$ \\
 & PatchTST & 0.0334$_{\pm.0009}$ & \textbf{0.0318}$_{\pm.0004}$ & 0.0331$_{\pm.0003}$ & 0.0338$_{\pm.0022}$ \\
 & TimeKAN & 0.0310$_{\pm.0003}$ & \textbf{0.0307}$_{\pm.0001}$ & 0.0309$_{\pm.0002}$ & 0.0339$_{\pm.0031}$ \\
 & CrossLinear & 0.0562$_{\pm.0005}$ & 0.0321$_{\pm.0005}$ & \textbf{0.0310}$_{\pm.0002}$ & 0.0326$_{\pm.0004}$ \\
 & Amplifier & 0.0326$_{\pm.0004}$ & 0.0335$_{\pm.0008}$ & \textbf{0.0323}$_{\pm.0005}$ & 0.0403$_{\pm.0035}$ \\
\addlinespace
CGMacros & TimeXer & 0.0152$_{\pm.0006}$ & \textbf{0.0116}$_{\pm.0002}$ & 0.0119$_{\pm.0005}$ & 0.0206$_{\pm.0115}$ \\
 & TiDE & 0.0108$_{\pm.0010}$ & \textbf{0.0093}$_{\pm.0002}$ & 0.0105$_{\pm.0006}$ & 0.0538$_{\pm.0247}$ \\
 & DUET & 0.0233$_{\pm.0054}$ & 0.0219$_{\pm.0058}$ & \textbf{0.0152}$_{\pm.0029}$ & 0.0414$_{\pm.0161}$ \\
 & PatchTST & 0.0142$_{\pm.0037}$ & 0.0094$_{\pm.0005}$ & \textbf{0.0091}$_{\pm.0003}$ & 0.0164$_{\pm.0058}$ \\
 & TimeKAN & 0.0109$_{\pm.0012}$ & 0.0101$_{\pm.0016}$ & \textbf{0.0095}$_{\pm.0012}$ & 0.0117$_{\pm.0020}$ \\
 & CrossLinear & \textbf{0.0086}$_{\pm.0001}$ & 0.0090$_{\pm.0001}$ & 0.0088$_{\pm.0001}$ & 0.0088$_{\pm.0001}$ \\
 & Amplifier & 0.0087$_{\pm.0001}$ & 0.0084$_{\pm.0002}$ & \textbf{0.0084}$_{\pm.0001}$ & 0.0366$_{\pm.0374}$ \\
\addlinespace
Shanghai & TimeXer & 0.1279$_{\pm.0024}$ & \textbf{0.1236}$_{\pm.0029}$ & 0.1270$_{\pm.0025}$ & 0.1625$_{\pm.0238}$ \\
 & TiDE & 0.1220$_{\pm.0010}$ & 0.1205$_{\pm.0025}$ & \textbf{0.1197}$_{\pm.0011}$ & 0.1324$_{\pm.0038}$ \\
 & DUET & 0.1426$_{\pm.0153}$ & 0.1403$_{\pm.0158}$ & \textbf{0.1263}$_{\pm.0037}$ & 0.1630$_{\pm.0067}$ \\
 & PatchTST & 0.1556$_{\pm.0041}$ & 0.1235$_{\pm.0070}$ & \textbf{0.1172}$_{\pm.0018}$ & 0.1487$_{\pm.0224}$ \\
 & TimeKAN & 0.1187$_{\pm.0015}$ & 0.1201$_{\pm.0006}$ & \textbf{0.1182}$_{\pm.0018}$ & 0.1300$_{\pm.0211}$ \\
 & CrossLinear & 0.1330$_{\pm.0034}$ & 0.1175$_{\pm.0046}$ & \textbf{0.1145}$_{\pm.0018}$ & 0.1223$_{\pm.0084}$ \\
 & Amplifier & 0.1310$_{\pm.0029}$ & 0.1286$_{\pm.0031}$ & \textbf{0.1211}$_{\pm.0025}$ & 0.1335$_{\pm.0057}$ \\
\addlinespace
MIMIC-Cardio & TimeXer & \textbf{0.0542}$_{\pm.0007}$ & 0.0543$_{\pm.0006}$ & 0.0548$_{\pm.0003}$ & 0.0570$_{\pm.0039}$ \\
 & TiDE & \textbf{0.0534}$_{\pm.0001}$ & 0.0537$_{\pm.0002}$ & 0.0541$_{\pm.0002}$ & 0.0631$_{\pm.0041}$ \\
 & DUET & 0.0587$_{\pm.0019}$ & 0.0596$_{\pm.0023}$ & \textbf{0.0586}$_{\pm.0039}$ & 0.0754$_{\pm.0110}$ \\
 & PatchTST & 0.0554$_{\pm.0004}$ & \textbf{0.0543}$_{\pm.0003}$ & 0.0546$_{\pm.0005}$ & 0.0582$_{\pm.0031}$ \\
 & TimeKAN & \textbf{0.0541}$_{\pm.0002}$ & 0.0542$_{\pm.0002}$ & 0.0543$_{\pm.0000}$ & 0.0608$_{\pm.0124}$ \\
 & CrossLinear & 0.0618$_{\pm.0004}$ & 0.0543$_{\pm.0002}$ & \textbf{0.0543}$_{\pm.0003}$ & 0.0558$_{\pm.0012}$ \\
 & Amplifier & 0.0557$_{\pm.0001}$ & 0.0558$_{\pm.0002}$ & \textbf{0.0555}$_{\pm.0002}$ & 0.0645$_{\pm.0106}$ \\
\addlinespace
PleiaData & TimeXer & \textbf{0.0281}$_{\pm.0108}$ & 0.0337$_{\pm.0099}$ & 0.0312$_{\pm.0070}$ & 0.0564$_{\pm.0125}$ \\
 & TiDE & 0.0121$_{\pm.0021}$ & \textbf{0.0115}$_{\pm.0009}$ & 0.0123$_{\pm.0007}$ & 0.0866$_{\pm.0260}$ \\
 & DUET & 0.0923$_{\pm.0057}$ & 0.0795$_{\pm.0058}$ & \textbf{0.0763}$_{\pm.0101}$ & 0.0936$_{\pm.0065}$ \\
 & PatchTST & 0.0448$_{\pm.0232}$ & 0.0318$_{\pm.0140}$ & \textbf{0.0313}$_{\pm.0162}$ & 0.0663$_{\pm.0186}$ \\
 & TimeKAN & 0.0643$_{\pm.0007}$ & \textbf{0.0552}$_{\pm.0015}$ & 0.0580$_{\pm.0007}$ & 0.0874$_{\pm.0329}$ \\
 & CrossLinear & 0.0766$_{\pm.0023}$ & 0.0075$_{\pm.0005}$ & 0.0081$_{\pm.0004}$ & \textbf{0.0070}$_{\pm.0001}$ \\
 & Amplifier & \textbf{0.0155}$_{\pm.0004}$ & 0.0166$_{\pm.0004}$ & 0.0198$_{\pm.0009}$ & 0.0273$_{\pm.0227}$ \\
\addlinespace
PREDIST & TimeXer & \textbf{0.1737}$_{\pm.0057}$ & 0.1799$_{\pm.0021}$ & 0.1799$_{\pm.0051}$ & 0.1864$_{\pm.0038}$ \\
 & TiDE & \textbf{0.1773}$_{\pm.0039}$ & 0.1775$_{\pm.0006}$ & 0.1784$_{\pm.0047}$ & 0.2593$_{\pm.0193}$ \\
 & DUET & 0.2013$_{\pm.0076}$ & \textbf{0.1878}$_{\pm.0029}$ & 0.1909$_{\pm.0025}$ & 0.2257$_{\pm.0063}$ \\
 & PatchTST & 0.1884$_{\pm.0027}$ & \textbf{0.1816}$_{\pm.0048}$ & 0.1853$_{\pm.0017}$ & 0.1999$_{\pm.0073}$ \\
 & TimeKAN & 0.1837$_{\pm.0028}$ & \textbf{0.1827}$_{\pm.0029}$ & 0.1842$_{\pm.0017}$ & 0.2050$_{\pm.0389}$ \\
 & CrossLinear & 0.1955$_{\pm.0009}$ & \textbf{0.1774}$_{\pm.0024}$ & 0.1784$_{\pm.0013}$ & 0.1864$_{\pm.0052}$ \\
 & Amplifier & \textbf{0.1863}$_{\pm.0022}$ & 0.1875$_{\pm.0019}$ & 0.1881$_{\pm.0010}$ & 0.2377$_{\pm.0373}$ \\
\addlinespace
Wastewater & TimeXer & \textbf{0.1626}$_{\pm.0050}$ & 0.1681$_{\pm.0027}$ & 0.1674$_{\pm.0065}$ & 0.1814$_{\pm.0107}$ \\
 & TiDE & \textbf{0.1657}$_{\pm.0076}$ & 0.1809$_{\pm.0044}$ & 0.1806$_{\pm.0088}$ & 0.2109$_{\pm.0347}$ \\
 & DUET & 0.1824$_{\pm.0237}$ & 0.1801$_{\pm.0152}$ & \textbf{0.1722}$_{\pm.0061}$ & 0.2126$_{\pm.0134}$ \\
 & PatchTST & 0.1651$_{\pm.0015}$ & 0.1699$_{\pm.0051}$ & 0.1615$_{\pm.0049}$ & \textbf{0.1581}$_{\pm.0053}$ \\
 & TimeKAN & 0.1702$_{\pm.0020}$ & \textbf{0.1574}$_{\pm.0050}$ & 0.1605$_{\pm.0027}$ & 0.1801$_{\pm.0063}$ \\
 & CrossLinear & 0.1754$_{\pm.0016}$ & 0.1520$_{\pm.0026}$ & 0.1520$_{\pm.0031}$ & \textbf{0.1478}$_{\pm.0028}$ \\
 & Amplifier & 0.1811$_{\pm.0027}$ & 0.1783$_{\pm.0018}$ & \textbf{0.1762}$_{\pm.0041}$ & 0.1861$_{\pm.0067}$ \\
\bottomrule
\end{tabular}

\end{table}

\begin{table}[H]
    \caption{Experiment 2, plan fusion: the complete grid.
    The \textsc{Concat} entries reuse the \textsc{AE} results from
    Table~\ref{tab:cells-exp1}.}
    \label{tab:cells-exp2}
    \centering
    \scriptsize
    \setlength{\tabcolsep}{1.6pt}
    \renewcommand{\arraystretch}{1.0}
    \begin{tabular}{llccccccc}
\toprule
Dataset & Backbone & \textsc{None} & \textsc{Concat} & \textsc{Res} & \textsc{Res-0} & \textsc{FiLM-0} & \textsc{Gate} & \textsc{X-attn} \\
\midrule
Greenhouse & TimeXer & \textbf{0.0587}$_{\pm.0006}$ & 0.0604$_{\pm.0006}$ & 0.0625$_{\pm.0019}$ & 0.0621$_{\pm.0013}$ & 0.0617$_{\pm.0008}$ & 0.0608$_{\pm.0012}$ & 0.0651$_{\pm.0021}$ \\
 & TiDE & \textbf{0.0609}$_{\pm.0005}$ & 0.0658$_{\pm.0027}$ & 0.0636$_{\pm.0017}$ & 0.0635$_{\pm.0020}$ & 0.0629$_{\pm.0012}$ & 0.0615$_{\pm.0006}$ & 0.0641$_{\pm.0011}$ \\
 & DUET & 0.0657$_{\pm.0006}$ & 0.0709$_{\pm.0016}$ & 0.0670$_{\pm.0013}$ & 0.0678$_{\pm.0013}$ & \textbf{0.0642}$_{\pm.0006}$ & 0.0649$_{\pm.0017}$ & 0.0694$_{\pm.0014}$ \\
 & PatchTST & \textbf{0.0563}$_{\pm.0008}$ & 0.0924$_{\pm.0334}$ & 0.0606$_{\pm.0013}$ & 0.0612$_{\pm.0013}$ & 0.0605$_{\pm.0026}$ & 0.0590$_{\pm.0014}$ & 0.0694$_{\pm.0039}$ \\
 & TimeKAN & \textbf{0.0587}$_{\pm.0004}$ & 0.0642$_{\pm.0011}$ & 0.0656$_{\pm.0020}$ & 0.0650$_{\pm.0022}$ & 0.0645$_{\pm.0014}$ & 0.0636$_{\pm.0017}$ & 0.0659$_{\pm.0026}$ \\
 & CrossLinear & 0.1626$_{\pm.0056}$ & 0.1066$_{\pm.0024}$ & 0.0731$_{\pm.0009}$ & 0.0732$_{\pm.0016}$ & 0.0721$_{\pm.0039}$ & \textbf{0.0703}$_{\pm.0019}$ & 0.1310$_{\pm.0060}$ \\
 & Amplifier & 0.0627$_{\pm.0006}$ & 0.0723$_{\pm.0092}$ & 0.0631$_{\pm.0008}$ & 0.0638$_{\pm.0007}$ & 0.0636$_{\pm.0010}$ & \textbf{0.0625}$_{\pm.0008}$ & 0.0697$_{\pm.0039}$ \\
\addlinespace
VitalDB & TimeXer & 0.0326$_{\pm.0013}$ & 0.0326$_{\pm.0014}$ & 0.0310$_{\pm.0001}$ & 0.0311$_{\pm.0002}$ & 0.0311$_{\pm.0004}$ & \textbf{0.0308}$_{\pm.0002}$ & 0.0328$_{\pm.0007}$ \\
 & TiDE & 0.0312$_{\pm.0002}$ & 0.0310$_{\pm.0001}$ & \textbf{0.0308}$_{\pm.0002}$ & 0.0309$_{\pm.0003}$ & 0.0309$_{\pm.0003}$ & 0.0310$_{\pm.0001}$ & 0.0317$_{\pm.0004}$ \\
 & DUET & 0.0342$_{\pm.0016}$ & 0.0384$_{\pm.0013}$ & 0.0312$_{\pm.0001}$ & 0.0312$_{\pm.0001}$ & 0.0312$_{\pm.0001}$ & \textbf{0.0311}$_{\pm.0001}$ & 0.0323$_{\pm.0001}$ \\
 & PatchTST & 0.0316$_{\pm.0003}$ & 0.0318$_{\pm.0004}$ & 0.0310$_{\pm.0003}$ & 0.0309$_{\pm.0002}$ & 0.0309$_{\pm.0003}$ & \textbf{0.0307}$_{\pm.0002}$ & 0.0322$_{\pm.0005}$ \\
 & TimeKAN & 0.0308$_{\pm.0002}$ & 0.0307$_{\pm.0001}$ & 0.0308$_{\pm.0001}$ & 0.0308$_{\pm.0002}$ & 0.0306$_{\pm.0001}$ & \textbf{0.0304}$_{\pm.0001}$ & 0.0312$_{\pm.0002}$ \\
 & CrossLinear & 0.0846$_{\pm.0005}$ & \textbf{0.0321}$_{\pm.0005}$ & 0.0328$_{\pm.0005}$ & 0.0331$_{\pm.0006}$ & 0.0329$_{\pm.0006}$ & 0.0329$_{\pm.0008}$ & 0.0335$_{\pm.0006}$ \\
 & Amplifier & 0.0321$_{\pm.0002}$ & 0.0335$_{\pm.0008}$ & 0.0324$_{\pm.0005}$ & 0.0324$_{\pm.0002}$ & 0.0323$_{\pm.0004}$ & \textbf{0.0319}$_{\pm.0004}$ & 0.0323$_{\pm.0004}$ \\
\addlinespace
CGMacros & TimeXer & 0.0109$_{\pm.0007}$ & 0.0116$_{\pm.0002}$ & \textbf{0.0085}$_{\pm.0001}$ & 0.0086$_{\pm.0002}$ & 0.0086$_{\pm.0002}$ & 0.0085$_{\pm.0002}$ & 0.0108$_{\pm.0006}$ \\
 & TiDE & 0.0095$_{\pm.0004}$ & 0.0093$_{\pm.0002}$ & 0.0093$_{\pm.0002}$ & 0.0093$_{\pm.0000}$ & \textbf{0.0092}$_{\pm.0002}$ & 0.0092$_{\pm.0002}$ & 0.0104$_{\pm.0002}$ \\
 & DUET & 0.0116$_{\pm.0007}$ & 0.0219$_{\pm.0058}$ & \textbf{0.0085}$_{\pm.0001}$ & 0.0086$_{\pm.0001}$ & 0.0087$_{\pm.0003}$ & 0.0088$_{\pm.0002}$ & 0.0118$_{\pm.0005}$ \\
 & PatchTST & 0.0092$_{\pm.0002}$ & 0.0094$_{\pm.0005}$ & 0.0086$_{\pm.0002}$ & 0.0085$_{\pm.0003}$ & 0.0084$_{\pm.0001}$ & \textbf{0.0084}$_{\pm.0002}$ & 0.0098$_{\pm.0002}$ \\
 & TimeKAN & 0.0099$_{\pm.0013}$ & 0.0101$_{\pm.0016}$ & \textbf{0.0085}$_{\pm.0004}$ & 0.0090$_{\pm.0005}$ & 0.0094$_{\pm.0005}$ & 0.0087$_{\pm.0008}$ & 0.0105$_{\pm.0009}$ \\
 & CrossLinear & 0.0624$_{\pm.0011}$ & 0.0090$_{\pm.0001}$ & 0.0085$_{\pm.0001}$ & 0.0085$_{\pm.0001}$ & 0.0084$_{\pm.0001}$ & \textbf{0.0084}$_{\pm.0000}$ & 0.0121$_{\pm.0009}$ \\
 & Amplifier & 0.0084$_{\pm.0001}$ & 0.0084$_{\pm.0002}$ & \textbf{0.0084}$_{\pm.0001}$ & 0.0085$_{\pm.0001}$ & 0.0085$_{\pm.0001}$ & 0.0085$_{\pm.0000}$ & 0.0095$_{\pm.0003}$ \\
\addlinespace
Shanghai & TimeXer & 0.1204$_{\pm.0013}$ & 0.1236$_{\pm.0029}$ & 0.1197$_{\pm.0011}$ & 0.1199$_{\pm.0021}$ & 0.1195$_{\pm.0015}$ & \textbf{0.1174}$_{\pm.0030}$ & 0.1211$_{\pm.0018}$ \\
 & TiDE & 0.1175$_{\pm.0012}$ & 0.1205$_{\pm.0025}$ & 0.1151$_{\pm.0004}$ & 0.1144$_{\pm.0008}$ & 0.1148$_{\pm.0007}$ & \textbf{0.1142}$_{\pm.0031}$ & 0.1199$_{\pm.0008}$ \\
 & DUET & 0.1260$_{\pm.0023}$ & 0.1403$_{\pm.0158}$ & 0.1220$_{\pm.0021}$ & \textbf{0.1201}$_{\pm.0023}$ & 0.1214$_{\pm.0013}$ & 0.1221$_{\pm.0008}$ & 0.1250$_{\pm.0023}$ \\
 & PatchTST & \textbf{0.1164}$_{\pm.0012}$ & 0.1235$_{\pm.0070}$ & 0.1172$_{\pm.0022}$ & 0.1171$_{\pm.0021}$ & 0.1164$_{\pm.0010}$ & 0.1169$_{\pm.0023}$ & 0.1179$_{\pm.0022}$ \\
 & TimeKAN & 0.1208$_{\pm.0017}$ & 0.1201$_{\pm.0006}$ & 0.1148$_{\pm.0017}$ & \textbf{0.1138}$_{\pm.0015}$ & 0.1142$_{\pm.0026}$ & 0.1141$_{\pm.0014}$ & 0.1268$_{\pm.0021}$ \\
 & CrossLinear & 0.1623$_{\pm.0008}$ & 0.1175$_{\pm.0046}$ & \textbf{0.1149}$_{\pm.0013}$ & 0.1156$_{\pm.0005}$ & 0.1153$_{\pm.0009}$ & 0.1155$_{\pm.0014}$ & 0.1292$_{\pm.0069}$ \\
 & Amplifier & 0.1162$_{\pm.0005}$ & 0.1286$_{\pm.0031}$ & 0.1151$_{\pm.0017}$ & 0.1155$_{\pm.0012}$ & 0.1169$_{\pm.0009}$ & \textbf{0.1151}$_{\pm.0012}$ & 0.1254$_{\pm.0038}$ \\
\addlinespace
MIMIC-Cardio & TimeXer & \textbf{0.0535}$_{\pm.0001}$ & 0.0543$_{\pm.0006}$ & 0.0538$_{\pm.0001}$ & 0.0539$_{\pm.0001}$ & 0.0538$_{\pm.0001}$ & 0.0535$_{\pm.0001}$ & 0.0542$_{\pm.0001}$ \\
 & TiDE & 0.0540$_{\pm.0003}$ & \textbf{0.0537}$_{\pm.0002}$ & 0.0542$_{\pm.0001}$ & 0.0543$_{\pm.0004}$ & 0.0544$_{\pm.0003}$ & 0.0543$_{\pm.0002}$ & 0.0542$_{\pm.0002}$ \\
 & DUET & \textbf{0.0540}$_{\pm.0003}$ & 0.0596$_{\pm.0023}$ & 0.0542$_{\pm.0001}$ & 0.0543$_{\pm.0001}$ & 0.0542$_{\pm.0001}$ & 0.0541$_{\pm.0001}$ & 0.0548$_{\pm.0002}$ \\
 & PatchTST & \textbf{0.0533}$_{\pm.0002}$ & 0.0543$_{\pm.0003}$ & 0.0541$_{\pm.0003}$ & 0.0540$_{\pm.0003}$ & 0.0541$_{\pm.0001}$ & 0.0539$_{\pm.0003}$ & 0.0541$_{\pm.0002}$ \\
 & TimeKAN & \textbf{0.0534}$_{\pm.0001}$ & 0.0542$_{\pm.0002}$ & 0.0538$_{\pm.0002}$ & 0.0537$_{\pm.0001}$ & 0.0539$_{\pm.0003}$ & 0.0539$_{\pm.0001}$ & 0.0540$_{\pm.0002}$ \\
 & CrossLinear & 0.1029$_{\pm.0004}$ & \textbf{0.0543}$_{\pm.0002}$ & 0.0557$_{\pm.0005}$ & 0.0558$_{\pm.0003}$ & 0.0558$_{\pm.0007}$ & 0.0553$_{\pm.0001}$ & 0.0573$_{\pm.0007}$ \\
 & Amplifier & \textbf{0.0546}$_{\pm.0002}$ & 0.0558$_{\pm.0002}$ & 0.0551$_{\pm.0001}$ & 0.0550$_{\pm.0001}$ & 0.0551$_{\pm.0002}$ & 0.0551$_{\pm.0002}$ & 0.0550$_{\pm.0001}$ \\
\addlinespace
PleiaData & TimeXer & 0.0284$_{\pm.0010}$ & 0.0337$_{\pm.0099}$ & 0.0263$_{\pm.0008}$ & 0.0270$_{\pm.0019}$ & \textbf{0.0251}$_{\pm.0009}$ & 0.0251$_{\pm.0068}$ & 0.0294$_{\pm.0046}$ \\
 & TiDE & 0.0116$_{\pm.0013}$ & 0.0115$_{\pm.0009}$ & 0.0105$_{\pm.0003}$ & 0.0108$_{\pm.0008}$ & \textbf{0.0103}$_{\pm.0001}$ & 0.0110$_{\pm.0009}$ & 0.0104$_{\pm.0007}$ \\
 & DUET & 0.0250$_{\pm.0030}$ & 0.0795$_{\pm.0058}$ & 0.0221$_{\pm.0031}$ & 0.0227$_{\pm.0030}$ & \textbf{0.0220}$_{\pm.0028}$ & 0.0230$_{\pm.0051}$ & 0.0236$_{\pm.0032}$ \\
 & PatchTST & 0.0226$_{\pm.0053}$ & 0.0318$_{\pm.0140}$ & 0.0238$_{\pm.0030}$ & 0.0207$_{\pm.0011}$ & 0.0203$_{\pm.0011}$ & 0.0219$_{\pm.0025}$ & \textbf{0.0196}$_{\pm.0037}$ \\
 & TimeKAN & 0.0685$_{\pm.0006}$ & 0.0552$_{\pm.0015}$ & 0.0435$_{\pm.0016}$ & 0.0434$_{\pm.0019}$ & 0.0411$_{\pm.0012}$ & 0.0392$_{\pm.0019}$ & \textbf{0.0364}$_{\pm.0021}$ \\
 & CrossLinear & 0.0254$_{\pm.0014}$ & 0.0075$_{\pm.0005}$ & 0.0076$_{\pm.0002}$ & 0.0076$_{\pm.0001}$ & \textbf{0.0070}$_{\pm.0002}$ & 0.0071$_{\pm.0002}$ & 0.0074$_{\pm.0002}$ \\
 & Amplifier & 0.0229$_{\pm.0003}$ & 0.0166$_{\pm.0004}$ & 0.0182$_{\pm.0004}$ & 0.0187$_{\pm.0008}$ & 0.0160$_{\pm.0004}$ & 0.0159$_{\pm.0006}$ & \textbf{0.0154}$_{\pm.0002}$ \\
\addlinespace
PREDIST & TimeXer & 0.1756$_{\pm.0014}$ & 0.1799$_{\pm.0021}$ & 0.1633$_{\pm.0018}$ & 0.1630$_{\pm.0018}$ & 0.1650$_{\pm.0030}$ & 0.1646$_{\pm.0005}$ & \textbf{0.1533}$_{\pm.0013}$ \\
 & TiDE & 0.1773$_{\pm.0012}$ & 0.1775$_{\pm.0006}$ & 0.1649$_{\pm.0015}$ & 0.1641$_{\pm.0009}$ & 0.1663$_{\pm.0011}$ & 0.1636$_{\pm.0018}$ & \textbf{0.1545}$_{\pm.0009}$ \\
 & DUET & 0.1752$_{\pm.0006}$ & 0.1878$_{\pm.0029}$ & 0.1651$_{\pm.0013}$ & 0.1658$_{\pm.0013}$ & 0.1642$_{\pm.0027}$ & 0.1654$_{\pm.0007}$ & \textbf{0.1522}$_{\pm.0016}$ \\
 & PatchTST & 0.1758$_{\pm.0010}$ & 0.1816$_{\pm.0048}$ & 0.1635$_{\pm.0018}$ & 0.1637$_{\pm.0010}$ & 0.1643$_{\pm.0027}$ & 0.1628$_{\pm.0029}$ & \textbf{0.1530}$_{\pm.0015}$ \\
 & TimeKAN & 0.1785$_{\pm.0002}$ & 0.1827$_{\pm.0029}$ & 0.1625$_{\pm.0005}$ & 0.1616$_{\pm.0010}$ & 0.1618$_{\pm.0009}$ & 0.1632$_{\pm.0006}$ & \textbf{0.1510}$_{\pm.0026}$ \\
 & CrossLinear & 0.2384$_{\pm.0015}$ & 0.1774$_{\pm.0024}$ & 0.1657$_{\pm.0023}$ & 0.1664$_{\pm.0025}$ & 0.1644$_{\pm.0024}$ & \textbf{0.1632}$_{\pm.0015}$ & 0.1759$_{\pm.0026}$ \\
 & Amplifier & 0.1799$_{\pm.0004}$ & 0.1875$_{\pm.0019}$ & 0.1656$_{\pm.0012}$ & 0.1652$_{\pm.0011}$ & 0.1654$_{\pm.0010}$ & 0.1655$_{\pm.0020}$ & \textbf{0.1568}$_{\pm.0009}$ \\
\addlinespace
Wastewater & TimeXer & 0.1772$_{\pm.0013}$ & 0.1681$_{\pm.0027}$ & 0.1434$_{\pm.0018}$ & 0.1449$_{\pm.0028}$ & 0.1436$_{\pm.0014}$ & \textbf{0.1417}$_{\pm.0028}$ & 0.1474$_{\pm.0052}$ \\
 & TiDE & 0.1991$_{\pm.0034}$ & 0.1809$_{\pm.0044}$ & 0.1549$_{\pm.0029}$ & 0.1570$_{\pm.0061}$ & \textbf{0.1534}$_{\pm.0019}$ & 0.1541$_{\pm.0029}$ & 0.1680$_{\pm.0046}$ \\
 & DUET & 0.1890$_{\pm.0025}$ & 0.1801$_{\pm.0152}$ & 0.1506$_{\pm.0031}$ & 0.1503$_{\pm.0021}$ & \textbf{0.1485}$_{\pm.0010}$ & 0.1549$_{\pm.0035}$ & 0.1657$_{\pm.0062}$ \\
 & PatchTST & 0.1875$_{\pm.0027}$ & 0.1699$_{\pm.0051}$ & 0.1451$_{\pm.0031}$ & 0.1447$_{\pm.0028}$ & \textbf{0.1440}$_{\pm.0012}$ & 0.1447$_{\pm.0016}$ & 0.1515$_{\pm.0060}$ \\
 & TimeKAN & 0.1942$_{\pm.0023}$ & 0.1574$_{\pm.0050}$ & \textbf{0.1430}$_{\pm.0020}$ & 0.1440$_{\pm.0011}$ & 0.1437$_{\pm.0014}$ & 0.1440$_{\pm.0029}$ & 0.1493$_{\pm.0049}$ \\
 & CrossLinear & 0.2005$_{\pm.0015}$ & \textbf{0.1520}$_{\pm.0026}$ & 0.1575$_{\pm.0025}$ & 0.1580$_{\pm.0004}$ & 0.1566$_{\pm.0034}$ & 0.1556$_{\pm.0029}$ & 0.1767$_{\pm.0031}$ \\
 & Amplifier & 0.2019$_{\pm.0005}$ & 0.1783$_{\pm.0018}$ & 0.1572$_{\pm.0016}$ & 0.1574$_{\pm.0026}$ & \textbf{0.1527}$_{\pm.0022}$ & 0.1553$_{\pm.0036}$ & 0.1732$_{\pm.0046}$ \\
\bottomrule
\end{tabular}

\end{table}

\begin{table}[H]
    \caption{Experiment 3, plan encoding with \textsc{Gate}: the complete grid.
    The \textsc{Instant} entries reuse the \textsc{Gate} results from
    Table~\ref{tab:cells-exp2}.}
    \label{tab:cells-exp3-gate}
    \centering
    \small
    \setlength{\tabcolsep}{4pt}
    \renewcommand{\arraystretch}{1.0}
    \begin{tabular}{llcccc}
\toprule
Dataset & Backbone & \textsc{Instant} & \textsc{Decay} & \textsc{Conv} & \textsc{Attn} \\
\midrule
Greenhouse & TimeXer & \textbf{0.0608}$_{\pm.0012}$ & 0.0610$_{\pm.0017}$ & 0.0611$_{\pm.0018}$ & 0.0608$_{\pm.0010}$ \\
 & TiDE & 0.0615$_{\pm.0006}$ & 0.0614$_{\pm.0007}$ & 0.0616$_{\pm.0009}$ & \textbf{0.0614}$_{\pm.0011}$ \\
 & DUET & 0.0649$_{\pm.0017}$ & 0.0657$_{\pm.0011}$ & 0.0656$_{\pm.0017}$ & \textbf{0.0646}$_{\pm.0009}$ \\
 & PatchTST & 0.0590$_{\pm.0014}$ & \textbf{0.0586}$_{\pm.0016}$ & 0.0596$_{\pm.0009}$ & 0.0589$_{\pm.0018}$ \\
 & TimeKAN & 0.0636$_{\pm.0017}$ & 0.0637$_{\pm.0016}$ & 0.0635$_{\pm.0018}$ & \textbf{0.0634}$_{\pm.0015}$ \\
 & CrossLinear & \textbf{0.0703}$_{\pm.0019}$ & 0.0706$_{\pm.0028}$ & 0.0705$_{\pm.0013}$ & 0.0713$_{\pm.0007}$ \\
 & Amplifier & \textbf{0.0625}$_{\pm.0008}$ & 0.0626$_{\pm.0007}$ & 0.0629$_{\pm.0008}$ & 0.0628$_{\pm.0014}$ \\
\addlinespace
VitalDB & TimeXer & \textbf{0.0308}$_{\pm.0002}$ & 0.0310$_{\pm.0002}$ & 0.0309$_{\pm.0002}$ & 0.0315$_{\pm.0003}$ \\
 & TiDE & 0.0310$_{\pm.0001}$ & \textbf{0.0307}$_{\pm.0004}$ & 0.0309$_{\pm.0003}$ & 0.0309$_{\pm.0002}$ \\
 & DUET & \textbf{0.0311}$_{\pm.0001}$ & 0.0311$_{\pm.0001}$ & 0.0312$_{\pm.0002}$ & 0.0318$_{\pm.0005}$ \\
 & PatchTST & 0.0307$_{\pm.0002}$ & \textbf{0.0306}$_{\pm.0001}$ & 0.0307$_{\pm.0002}$ & 0.0311$_{\pm.0003}$ \\
 & TimeKAN & \textbf{0.0304}$_{\pm.0001}$ & 0.0305$_{\pm.0002}$ & 0.0305$_{\pm.0002}$ & 0.0307$_{\pm.0002}$ \\
 & CrossLinear & 0.0329$_{\pm.0008}$ & 0.0332$_{\pm.0004}$ & 0.0334$_{\pm.0012}$ & \textbf{0.0325}$_{\pm.0009}$ \\
 & Amplifier & 0.0319$_{\pm.0004}$ & \textbf{0.0318}$_{\pm.0005}$ & 0.0318$_{\pm.0004}$ & 0.0318$_{\pm.0005}$ \\
\addlinespace
CGMacros & TimeXer & 0.0085$_{\pm.0002}$ & 0.0085$_{\pm.0002}$ & \textbf{0.0084}$_{\pm.0002}$ & 0.0092$_{\pm.0002}$ \\
 & TiDE & \textbf{0.0092}$_{\pm.0002}$ & 0.0093$_{\pm.0004}$ & 0.0092$_{\pm.0002}$ & 0.0097$_{\pm.0006}$ \\
 & DUET & 0.0088$_{\pm.0002}$ & 0.0088$_{\pm.0001}$ & \textbf{0.0087}$_{\pm.0001}$ & 0.0091$_{\pm.0002}$ \\
 & PatchTST & \textbf{0.0084}$_{\pm.0002}$ & 0.0084$_{\pm.0002}$ & 0.0084$_{\pm.0001}$ & 0.0090$_{\pm.0003}$ \\
 & TimeKAN & \textbf{0.0087}$_{\pm.0008}$ & 0.0090$_{\pm.0010}$ & 0.0088$_{\pm.0009}$ & 0.0097$_{\pm.0007}$ \\
 & CrossLinear & 0.0084$_{\pm.0000}$ & \textbf{0.0084}$_{\pm.0001}$ & 0.0084$_{\pm.0002}$ & 0.0086$_{\pm.0004}$ \\
 & Amplifier & \textbf{0.0085}$_{\pm.0000}$ & 0.0085$_{\pm.0001}$ & 0.0086$_{\pm.0002}$ & 0.0093$_{\pm.0003}$ \\
\addlinespace
Shanghai & TimeXer & \textbf{0.1174}$_{\pm.0030}$ & 0.1196$_{\pm.0025}$ & 0.1185$_{\pm.0031}$ & 0.1185$_{\pm.0008}$ \\
 & TiDE & \textbf{0.1142}$_{\pm.0031}$ & 0.1147$_{\pm.0029}$ & 0.1147$_{\pm.0032}$ & 0.1157$_{\pm.0012}$ \\
 & DUET & 0.1221$_{\pm.0008}$ & 0.1217$_{\pm.0018}$ & \textbf{0.1211}$_{\pm.0010}$ & 0.1235$_{\pm.0019}$ \\
 & PatchTST & 0.1169$_{\pm.0023}$ & 0.1177$_{\pm.0028}$ & 0.1167$_{\pm.0027}$ & \textbf{0.1165}$_{\pm.0021}$ \\
 & TimeKAN & 0.1141$_{\pm.0014}$ & 0.1139$_{\pm.0010}$ & \textbf{0.1138}$_{\pm.0012}$ & 0.1187$_{\pm.0027}$ \\
 & CrossLinear & 0.1155$_{\pm.0014}$ & \textbf{0.1153}$_{\pm.0015}$ & 0.1156$_{\pm.0013}$ & 0.1153$_{\pm.0013}$ \\
 & Amplifier & 0.1151$_{\pm.0012}$ & 0.1147$_{\pm.0009}$ & 0.1147$_{\pm.0021}$ & \textbf{0.1142}$_{\pm.0009}$ \\
\addlinespace
MIMIC-Cardio & TimeXer & \textbf{0.0535}$_{\pm.0001}$ & 0.0536$_{\pm.0001}$ & 0.0538$_{\pm.0002}$ & 0.0539$_{\pm.0001}$ \\
 & TiDE & 0.0543$_{\pm.0002}$ & 0.0543$_{\pm.0003}$ & \textbf{0.0541}$_{\pm.0004}$ & 0.0542$_{\pm.0003}$ \\
 & DUET & 0.0541$_{\pm.0001}$ & \textbf{0.0540}$_{\pm.0001}$ & 0.0547$_{\pm.0005}$ & 0.0546$_{\pm.0002}$ \\
 & PatchTST & 0.0539$_{\pm.0003}$ & \textbf{0.0536}$_{\pm.0002}$ & 0.0537$_{\pm.0001}$ & 0.0539$_{\pm.0002}$ \\
 & TimeKAN & 0.0539$_{\pm.0001}$ & \textbf{0.0537}$_{\pm.0002}$ & 0.0540$_{\pm.0000}$ & 0.0539$_{\pm.0001}$ \\
 & CrossLinear & 0.0553$_{\pm.0001}$ & 0.0552$_{\pm.0002}$ & \textbf{0.0551}$_{\pm.0002}$ & 0.0553$_{\pm.0004}$ \\
 & Amplifier & 0.0551$_{\pm.0002}$ & \textbf{0.0550}$_{\pm.0002}$ & 0.0550$_{\pm.0002}$ & 0.0551$_{\pm.0002}$ \\
\addlinespace
PleiaData & TimeXer & 0.0251$_{\pm.0068}$ & 0.0290$_{\pm.0013}$ & \textbf{0.0217}$_{\pm.0074}$ & 0.0217$_{\pm.0052}$ \\
 & TiDE & 0.0110$_{\pm.0009}$ & 0.0110$_{\pm.0009}$ & 0.0109$_{\pm.0008}$ & \textbf{0.0104}$_{\pm.0003}$ \\
 & DUET & 0.0230$_{\pm.0051}$ & 0.0241$_{\pm.0020}$ & \textbf{0.0206}$_{\pm.0020}$ & 0.0215$_{\pm.0035}$ \\
 & PatchTST & 0.0219$_{\pm.0025}$ & 0.0222$_{\pm.0013}$ & 0.0215$_{\pm.0021}$ & \textbf{0.0183}$_{\pm.0012}$ \\
 & TimeKAN & 0.0392$_{\pm.0019}$ & 0.0392$_{\pm.0012}$ & 0.0329$_{\pm.0018}$ & \textbf{0.0253}$_{\pm.0005}$ \\
 & CrossLinear & 0.0071$_{\pm.0002}$ & 0.0070$_{\pm.0000}$ & 0.0067$_{\pm.0000}$ & \textbf{0.0067}$_{\pm.0001}$ \\
 & Amplifier & 0.0159$_{\pm.0006}$ & 0.0156$_{\pm.0003}$ & 0.0156$_{\pm.0004}$ & \textbf{0.0145}$_{\pm.0003}$ \\
\addlinespace
PREDIST & TimeXer & 0.1646$_{\pm.0005}$ & 0.1584$_{\pm.0012}$ & \textbf{0.1579}$_{\pm.0004}$ & 0.1587$_{\pm.0021}$ \\
 & TiDE & 0.1636$_{\pm.0018}$ & 0.1577$_{\pm.0011}$ & \textbf{0.1568}$_{\pm.0011}$ & 0.1586$_{\pm.0047}$ \\
 & DUET & 0.1654$_{\pm.0007}$ & 0.1615$_{\pm.0027}$ & \textbf{0.1583}$_{\pm.0003}$ & 0.1592$_{\pm.0015}$ \\
 & PatchTST & 0.1628$_{\pm.0029}$ & 0.1581$_{\pm.0016}$ & 0.1562$_{\pm.0016}$ & \textbf{0.1560}$_{\pm.0013}$ \\
 & TimeKAN & 0.1632$_{\pm.0006}$ & 0.1583$_{\pm.0016}$ & \textbf{0.1548}$_{\pm.0010}$ & 0.1576$_{\pm.0024}$ \\
 & CrossLinear & 0.1632$_{\pm.0015}$ & 0.1590$_{\pm.0015}$ & \textbf{0.1563}$_{\pm.0017}$ & 0.1613$_{\pm.0025}$ \\
 & Amplifier & 0.1655$_{\pm.0020}$ & 0.1604$_{\pm.0015}$ & \textbf{0.1573}$_{\pm.0009}$ & 0.1604$_{\pm.0025}$ \\
\addlinespace
Wastewater & TimeXer & 0.1417$_{\pm.0028}$ & \textbf{0.1321}$_{\pm.0029}$ & 0.1333$_{\pm.0022}$ & 0.1438$_{\pm.0014}$ \\
 & TiDE & 0.1541$_{\pm.0029}$ & 0.1528$_{\pm.0038}$ & \textbf{0.1414}$_{\pm.0032}$ & 0.1515$_{\pm.0031}$ \\
 & DUET & 0.1549$_{\pm.0035}$ & 0.1424$_{\pm.0045}$ & \textbf{0.1401}$_{\pm.0033}$ & 0.1455$_{\pm.0018}$ \\
 & PatchTST & 0.1447$_{\pm.0016}$ & 0.1345$_{\pm.0024}$ & \textbf{0.1339}$_{\pm.0020}$ & 0.1414$_{\pm.0059}$ \\
 & TimeKAN & 0.1440$_{\pm.0029}$ & 0.1331$_{\pm.0008}$ & \textbf{0.1321}$_{\pm.0015}$ & 0.1412$_{\pm.0036}$ \\
 & CrossLinear & 0.1556$_{\pm.0029}$ & 0.1546$_{\pm.0027}$ & \textbf{0.1501}$_{\pm.0030}$ & 0.1566$_{\pm.0033}$ \\
 & Amplifier & 0.1553$_{\pm.0036}$ & 0.1495$_{\pm.0025}$ & \textbf{0.1416}$_{\pm.0022}$ & 0.1520$_{\pm.0021}$ \\
\bottomrule
\end{tabular}

\end{table}

\begin{table}[H]
    \caption{Experiment 3, plan encoding with \textsc{FiLM-0}: the complete grid.
    The \textsc{Instant} entries reuse the \textsc{FiLM-0} results from
    Table~\ref{tab:cells-exp2}; Table~\ref{tab:cells-exp3-gate} gives the same
    grid with \textsc{Gate}.}
    \label{tab:cells-exp3-film}
    \centering
    \small
    \setlength{\tabcolsep}{4pt}
    \renewcommand{\arraystretch}{1.0}
    \begin{tabular}{llcccc}
\toprule
Dataset & Backbone & \textsc{Instant} & \textsc{Decay} & \textsc{Conv} & \textsc{Attn} \\
\midrule
Greenhouse & TimeXer & 0.0617$_{\pm.0008}$ & \textbf{0.0613}$_{\pm.0007}$ & 0.0623$_{\pm.0013}$ & 0.0618$_{\pm.0011}$ \\
 & TiDE & 0.0629$_{\pm.0012}$ & 0.0629$_{\pm.0012}$ & 0.0631$_{\pm.0013}$ & \textbf{0.0627}$_{\pm.0015}$ \\
 & DUET & \textbf{0.0642}$_{\pm.0006}$ & 0.0642$_{\pm.0014}$ & 0.0649$_{\pm.0005}$ & 0.0655$_{\pm.0014}$ \\
 & PatchTST & 0.0605$_{\pm.0026}$ & \textbf{0.0596}$_{\pm.0009}$ & 0.0609$_{\pm.0024}$ & 0.0598$_{\pm.0010}$ \\
 & TimeKAN & 0.0645$_{\pm.0014}$ & 0.0654$_{\pm.0026}$ & 0.0648$_{\pm.0015}$ & \textbf{0.0637}$_{\pm.0013}$ \\
 & CrossLinear & 0.0721$_{\pm.0039}$ & 0.0724$_{\pm.0038}$ & 0.0722$_{\pm.0033}$ & \textbf{0.0720}$_{\pm.0029}$ \\
 & Amplifier & 0.0636$_{\pm.0010}$ & \textbf{0.0636}$_{\pm.0008}$ & 0.0643$_{\pm.0010}$ & 0.0641$_{\pm.0006}$ \\
\addlinespace
VitalDB & TimeXer & 0.0311$_{\pm.0004}$ & \textbf{0.0309}$_{\pm.0001}$ & 0.0309$_{\pm.0002}$ & 0.0316$_{\pm.0001}$ \\
 & TiDE & \textbf{0.0309}$_{\pm.0003}$ & 0.0309$_{\pm.0002}$ & 0.0309$_{\pm.0001}$ & 0.0309$_{\pm.0001}$ \\
 & DUET & \textbf{0.0312}$_{\pm.0001}$ & 0.0313$_{\pm.0001}$ & 0.0312$_{\pm.0001}$ & 0.0316$_{\pm.0004}$ \\
 & PatchTST & 0.0309$_{\pm.0003}$ & 0.0309$_{\pm.0001}$ & \textbf{0.0309}$_{\pm.0001}$ & 0.0315$_{\pm.0003}$ \\
 & TimeKAN & \textbf{0.0306}$_{\pm.0001}$ & 0.0306$_{\pm.0002}$ & 0.0307$_{\pm.0001}$ & 0.0307$_{\pm.0002}$ \\
 & CrossLinear & 0.0329$_{\pm.0006}$ & 0.0327$_{\pm.0007}$ & \textbf{0.0327}$_{\pm.0006}$ & 0.0328$_{\pm.0004}$ \\
 & Amplifier & 0.0323$_{\pm.0004}$ & \textbf{0.0322}$_{\pm.0004}$ & 0.0323$_{\pm.0004}$ & 0.0323$_{\pm.0004}$ \\
\addlinespace
CGMacros & TimeXer & 0.0086$_{\pm.0002}$ & \textbf{0.0085}$_{\pm.0003}$ & 0.0086$_{\pm.0004}$ & 0.0091$_{\pm.0002}$ \\
 & TiDE & \textbf{0.0092}$_{\pm.0002}$ & 0.0092$_{\pm.0002}$ & 0.0092$_{\pm.0001}$ & 0.0097$_{\pm.0002}$ \\
 & DUET & 0.0087$_{\pm.0003}$ & 0.0087$_{\pm.0003}$ & \textbf{0.0087}$_{\pm.0001}$ & 0.0090$_{\pm.0002}$ \\
 & PatchTST & 0.0084$_{\pm.0001}$ & 0.0084$_{\pm.0001}$ & \textbf{0.0084}$_{\pm.0002}$ & 0.0090$_{\pm.0002}$ \\
 & TimeKAN & 0.0094$_{\pm.0005}$ & \textbf{0.0091}$_{\pm.0006}$ & 0.0093$_{\pm.0005}$ & 0.0104$_{\pm.0002}$ \\
 & CrossLinear & \textbf{0.0084}$_{\pm.0001}$ & 0.0084$_{\pm.0001}$ & 0.0085$_{\pm.0001}$ & 0.0088$_{\pm.0002}$ \\
 & Amplifier & \textbf{0.0085}$_{\pm.0001}$ & 0.0085$_{\pm.0002}$ & 0.0086$_{\pm.0002}$ & 0.0092$_{\pm.0003}$ \\
\addlinespace
Shanghai & TimeXer & 0.1195$_{\pm.0015}$ & 0.1188$_{\pm.0020}$ & \textbf{0.1182}$_{\pm.0019}$ & 0.1186$_{\pm.0020}$ \\
 & TiDE & 0.1148$_{\pm.0007}$ & 0.1151$_{\pm.0006}$ & \textbf{0.1142}$_{\pm.0011}$ & 0.1154$_{\pm.0005}$ \\
 & DUET & 0.1214$_{\pm.0013}$ & 0.1215$_{\pm.0015}$ & \textbf{0.1205}$_{\pm.0010}$ & 0.1238$_{\pm.0018}$ \\
 & PatchTST & 0.1164$_{\pm.0010}$ & 0.1162$_{\pm.0012}$ & \textbf{0.1157}$_{\pm.0020}$ & 0.1162$_{\pm.0016}$ \\
 & TimeKAN & 0.1142$_{\pm.0026}$ & 0.1137$_{\pm.0025}$ & \textbf{0.1127}$_{\pm.0020}$ & 0.1161$_{\pm.0019}$ \\
 & CrossLinear & 0.1153$_{\pm.0009}$ & 0.1155$_{\pm.0014}$ & 0.1154$_{\pm.0015}$ & \textbf{0.1149}$_{\pm.0020}$ \\
 & Amplifier & 0.1169$_{\pm.0009}$ & 0.1166$_{\pm.0010}$ & \textbf{0.1161}$_{\pm.0010}$ & 0.1161$_{\pm.0008}$ \\
\addlinespace
MIMIC-Cardio & TimeXer & \textbf{0.0538}$_{\pm.0001}$ & 0.0539$_{\pm.0001}$ & 0.0540$_{\pm.0001}$ & 0.0540$_{\pm.0002}$ \\
 & TiDE & 0.0544$_{\pm.0003}$ & 0.0543$_{\pm.0003}$ & \textbf{0.0542}$_{\pm.0002}$ & 0.0544$_{\pm.0003}$ \\
 & DUET & \textbf{0.0542}$_{\pm.0001}$ & 0.0543$_{\pm.0002}$ & 0.0544$_{\pm.0002}$ & 0.0550$_{\pm.0003}$ \\
 & PatchTST & 0.0541$_{\pm.0001}$ & \textbf{0.0539}$_{\pm.0002}$ & 0.0540$_{\pm.0002}$ & 0.0544$_{\pm.0004}$ \\
 & TimeKAN & \textbf{0.0539}$_{\pm.0003}$ & 0.0541$_{\pm.0003}$ & 0.0541$_{\pm.0002}$ & 0.0540$_{\pm.0001}$ \\
 & CrossLinear & 0.0558$_{\pm.0007}$ & 0.0559$_{\pm.0007}$ & \textbf{0.0554}$_{\pm.0004}$ & 0.0557$_{\pm.0004}$ \\
 & Amplifier & 0.0551$_{\pm.0002}$ & 0.0551$_{\pm.0002}$ & 0.0551$_{\pm.0001}$ & \textbf{0.0550}$_{\pm.0004}$ \\
\addlinespace
PleiaData & TimeXer & 0.0251$_{\pm.0009}$ & 0.0261$_{\pm.0024}$ & 0.0240$_{\pm.0005}$ & \textbf{0.0239}$_{\pm.0010}$ \\
 & TiDE & 0.0103$_{\pm.0001}$ & 0.0104$_{\pm.0005}$ & 0.0099$_{\pm.0004}$ & \textbf{0.0099}$_{\pm.0002}$ \\
 & DUET & 0.0220$_{\pm.0028}$ & 0.0202$_{\pm.0029}$ & 0.0208$_{\pm.0034}$ & \textbf{0.0187}$_{\pm.0038}$ \\
 & PatchTST & 0.0203$_{\pm.0011}$ & 0.0224$_{\pm.0020}$ & \textbf{0.0196}$_{\pm.0016}$ & 0.0216$_{\pm.0029}$ \\
 & TimeKAN & 0.0411$_{\pm.0012}$ & 0.0403$_{\pm.0015}$ & 0.0350$_{\pm.0013}$ & \textbf{0.0265}$_{\pm.0017}$ \\
 & CrossLinear & \textbf{0.0070}$_{\pm.0002}$ & 0.0071$_{\pm.0003}$ & 0.0071$_{\pm.0002}$ & 0.0072$_{\pm.0001}$ \\
 & Amplifier & 0.0160$_{\pm.0004}$ & 0.0161$_{\pm.0007}$ & 0.0158$_{\pm.0002}$ & \textbf{0.0145}$_{\pm.0003}$ \\
\addlinespace
PREDIST & TimeXer & 0.1650$_{\pm.0030}$ & 0.1608$_{\pm.0014}$ & \textbf{0.1570}$_{\pm.0017}$ & 0.1588$_{\pm.0014}$ \\
 & TiDE & 0.1663$_{\pm.0011}$ & 0.1616$_{\pm.0010}$ & \textbf{0.1565}$_{\pm.0012}$ & 0.1577$_{\pm.0014}$ \\
 & DUET & 0.1642$_{\pm.0027}$ & 0.1623$_{\pm.0006}$ & \textbf{0.1582}$_{\pm.0008}$ & 0.1595$_{\pm.0023}$ \\
 & PatchTST & 0.1643$_{\pm.0027}$ & 0.1596$_{\pm.0014}$ & \textbf{0.1569}$_{\pm.0012}$ & 0.1576$_{\pm.0015}$ \\
 & TimeKAN & 0.1618$_{\pm.0009}$ & 0.1567$_{\pm.0016}$ & \textbf{0.1540}$_{\pm.0013}$ & 0.1585$_{\pm.0040}$ \\
 & CrossLinear & 0.1644$_{\pm.0024}$ & 0.1594$_{\pm.0013}$ & \textbf{0.1584}$_{\pm.0010}$ & 0.1639$_{\pm.0008}$ \\
 & Amplifier & 0.1654$_{\pm.0010}$ & 0.1614$_{\pm.0006}$ & \textbf{0.1592}$_{\pm.0016}$ & 0.1597$_{\pm.0022}$ \\
\addlinespace
Wastewater & TimeXer & 0.1436$_{\pm.0014}$ & \textbf{0.1327}$_{\pm.0008}$ & 0.1335$_{\pm.0024}$ & 0.1428$_{\pm.0041}$ \\
 & TiDE & 0.1534$_{\pm.0019}$ & 0.1470$_{\pm.0015}$ & \textbf{0.1426}$_{\pm.0037}$ & 0.1480$_{\pm.0039}$ \\
 & DUET & 0.1485$_{\pm.0010}$ & \textbf{0.1376}$_{\pm.0041}$ & 0.1399$_{\pm.0034}$ & 0.1505$_{\pm.0040}$ \\
 & PatchTST & 0.1440$_{\pm.0012}$ & 0.1338$_{\pm.0020}$ & \textbf{0.1332}$_{\pm.0010}$ & 0.1405$_{\pm.0046}$ \\
 & TimeKAN & 0.1437$_{\pm.0014}$ & 0.1334$_{\pm.0012}$ & \textbf{0.1333}$_{\pm.0008}$ & 0.1408$_{\pm.0037}$ \\
 & CrossLinear & 0.1566$_{\pm.0034}$ & 0.1550$_{\pm.0031}$ & \textbf{0.1480}$_{\pm.0024}$ & 0.1557$_{\pm.0033}$ \\
 & Amplifier & 0.1527$_{\pm.0022}$ & 0.1459$_{\pm.0012}$ & \textbf{0.1406}$_{\pm.0020}$ & 0.1486$_{\pm.0033}$ \\
\bottomrule
\end{tabular}

\end{table}

\par
\endgroup

\end{document}